\documentclass{article} %
\usepackage{iclr2027_conference,times}

\usepackage{amsmath,amsfonts,bm}

\def\eqref#1{equation~\ref{#1}}

\def\1{\bm{1}}

\DeclareMathAlphabet{\mathsfit}{\encodingdefault}{\sfdefault}{m}{sl}
\SetMathAlphabet{\mathsfit}{bold}{\encodingdefault}{\sfdefault}{bx}{n}

\usepackage{graphicx}
\usepackage{hyperref}
\usepackage{url}
\usepackage{booktabs}
\usepackage{makecell}
\usepackage{wrapfig}
\usepackage{placeins}

\title{The Shape of the DPO Hyperparameter Landscape: What to Measure, How It Moves with Scale, and When Not to Transfer It}

\author{%
\mdseries
\begin{tabular}{@{}l@{\hspace{4em}}l@{}}
\textbf{Hannan Mahadik} & \textbf{David Salinas} \\
ELLIS Institute Tübingen & ELLIS Institute Tübingen \& Prior Labs \\
Germany & Germany \\
\texttt{hannan.mahadik@tue.ellis.eu} & \texttt{david.salinas@tue.ellis.eu} \\[2ex]
\textbf{Jacek Golebiowski} & \textbf{Aaron Klein} \\
distil labs & ELLIS Institute Tübingen \\
Germany & Germany \\
\texttt{jacek@distillabs.ai} & \texttt{aaron.klein@tue.ellis.eu}
\end{tabular}}

\newcommand{\zall}{$Z_{\mathrm{All}}$}
\newcommand{\zdynamic}{$Z_{\mathrm{Dynamic}}$}
\newcommand{\zstatic}{$Z_{\mathrm{Static}}$}
\newcommand{\zmacro}{$Z_{\mathrm{Macro}}$}

\iclrfinalcopy %
\begin{document}

\maketitle

\begin{abstract}
Direct preference optimisation (DPO) is widely used for language model post-training, but its performance depends strongly on the choice of its hyperparameters, such as the learning rate and the reference-policy regularisation parameter $\beta$. To better understand the DPO hyperparameter landscape, we conduct extensive experiments using a dense hyperparameter grid across different model scales from the same family. This allows us to analyse how the optimal DPO hyperparameters evolve with model size and the number of update steps, with intermediate checkpoints providing an inexpensive signal to accelerate the hyperparameter optimisation process. Furthermore, by evaluating our models on both static and dynamic benchmarks, we identify which benchmarks provide the strongest signal for hyperparameter selection. Finally, we investigate how different state-of-the-art hyperparameter optimisation strategies perform in this setting. Our experiments show that the optimal learning rate varies with model size and that hyperparameters are not transferable across training datasets. We also find that validation loss is a poor proxy for hyperparameter optimisation, as hyperparameters that minimize validation loss only weakly correlate with optimal hyperparameters for downstream benchmarks. Therefore, DPO hyperparameters need to be tuned specifically for the target model, dataset, and evaluation objective. We also show that short candidate training runs can substantially reduce the cost of this selection. We release all our training configurations and evaluation results to support reproducible analysis of selection quality and training cost.
\end{abstract}

\section{Introduction}
\label{sec:introduction}

Large language models (LLMs) have rapidly advanced in both capability and scale, but pretrained models are not necessarily well-suited for interaction with users: they may fail to reliably follow instructions, answer questions, or respond appropriately to user intent. Post-training is therefore a critical step in transforming a pretrained model into a useful assistant ~\citep{ouyang2022training}. 
Common post-training recipes consist of supervised fine-tuning (SFT), which teaches a model to follow instructions using demonstration data, followed by preference optimisation that adapts model behaviour toward preferred responses. 

Direct preference optimisation (DPO)~\citep{rafailov2023direct} optimises directly from preference pairs relative to a fixed SFT reference policy, without a separate reward-model training stage. 
Its learning rate controls the size of the update, while $\beta$ controls the strength of the reference-adjusted preference signal in the DPO objective and therefore the objective's trade-off between fitting preference pairs and remaining close to the reference policy. Both hyperparameters may therefore have effects that depend on the reference checkpoint and preference data. 
Prior work has already shown sensitivity to $\beta$ and the characteristics of preference pairs ~\citep{wu2024beta}.
This raises the question: how reliably can a DPO configuration selected in one setting be applied to another?

\begin{figure}
\centering
\includegraphics[width=0.45\textwidth]{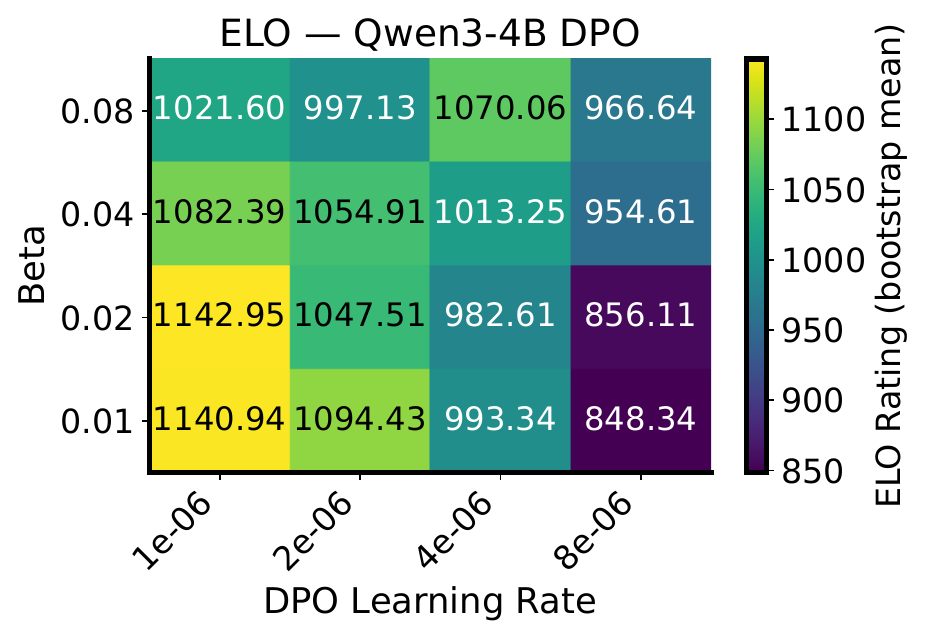}
\hfill
\includegraphics[width=0.45\textwidth]{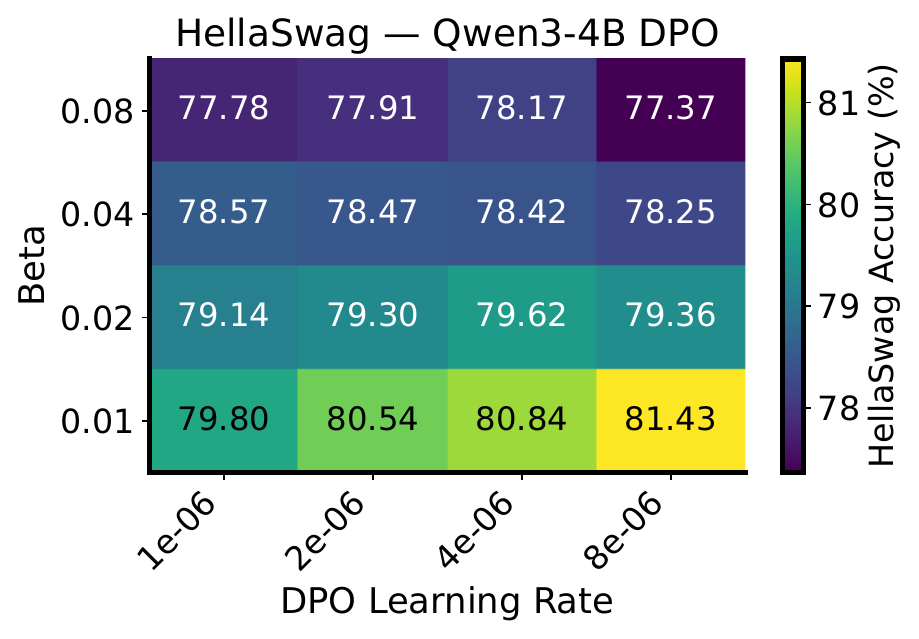}
\caption{\textbf{DPO performance is sensitive to the jointly tuned learning rate and \(\beta\).}
Each cell is one configuration in the displayed \(4\times4\) Qwen3-4B subgrid, scored by Elo (\textbf{left}) and HellaSwag (\textbf{right}); higher is better. Elo spans roughly 295 points, and the two evaluations select different configurations.}
\label{fig:teaser}
\end{figure}

Post-training recipes commonly report configurations that perform well for a particular model, dataset, and reference checkpoint, often using a fixed suite of static capability benchmarks to guide those choices~\citep{tunstall2023zephyr,ivison2023camels,lambert2024tulu,olmo2025olmo3,huggingfacetb2025smol}. These configurations are then generally reused across related models or experiments, implicitly assuming both that the configuration transfers and that the static evaluation signal used to select it remains informative. Whether either assumption holds across model scale has received substantially less systematic investigation. Getting this wrong is not a minor efficiency loss. In our primary sweep, the gap between the best and the worst configuration within a single model size reaches roughly 295 Elo points at a 4B parameter scale (Figure~\ref{fig:teaser}). Under the fitted Bradley--Terry model used to construct these scores, this corresponds to an estimated $84.5\%$ probability that the best configuration wins a pairwise comparison with the worst. By comparison, the DPO improvement over its SFT checkpoint is 176 Elo points, corresponding to a $73.3\%$ win probability. 
The worst configurations in the grid also land \emph{below} the SFT checkpoint. A poor configuration can therefore erase the entire benefit of the stage and leave a model worse than its SFT checkpoint. 

We study DPO selection across five Qwen3~\citep{qwen2025qwen3} sizes (0.6B--14B), three complementary Llama models~\citep{grattafiori2024llama3,meta2024llama32}, eleven downstream benchmarks, and two LLM judges. Fully logged training trajectories allow us to analyse configuration transfer and compare single- and multi-fidelity hyperparameter optimisation (HPO) methods without additional training. We release the resulting checkpoint-level benchmark, representing approximately 11,849 H100 GPU-hours of completed DPO training and evaluation (including the complementary experiments), so future work can reuse these results without having to repeat this compute. Appendix~\ref{app:compute-accounting} provides the breakdown.

Our contributions are as follows:
\begin{itemize}
    \item \textbf{Evaluation signals for DPO tuning.}
    We show that validation loss is an unreliable selection signal and that chat benchmarks produce more consistent configuration rankings than static benchmarks, with limited sensitivity to the choice of the judge model used.
    \item \textbf{Transfer across scale and training data.} We show that DPO configurations do not reliably transfer across model sizes or training preference datasets. Preferred learning rates exhibit an overall decrease with Qwen3 model scale, but configuration rankings can vary sharply even between neighbouring models.
    \item \textbf{Compute-efficient hyperparameter selection.} Partial model training runs can provide useful ranking information within roughly $20$--$30\%$ of training. We also show that under sufficient budgets, adaptive HPO methods can outperform random search.
    \item \textbf{An offline benchmark for DPO hyperparameter selection.} We release checkpoint-level Qwen3 and Llama trajectories with training costs and scores on eleven benchmarks, along with code to reconstruct the surrogate environments and HPO experiments. This enables reproducible research on selection signals, early-selection policies, transfer, and compute allocation without repeating these expensive DPO runs.
\end{itemize}

The remainder of the paper is organized as follows: Section~\ref{sec:related} situates our study within work on DPO, hyperparameter transfer, and evaluation reliability. Section~\ref{sec:setup} describes the models, data, training grids, evaluation suite, downstream aggregation objectives, and offline HPO simulation framework. Section~\ref{sec:results} presents the empirical results, followed by practical recommendations, limitations, and conclusions. The appendix provides complete experimental settings, supplementary analyses, and detailed HPO simulation results.

\section{Related Work}
\label{sec:related}

\paragraph{DPO and hyperparameter sensitivity.}
DPO optimises preferences relative to a fixed reference policy without training a separate reward model~\citep{rafailov2023direct}. Alternative objectives such as ORPO and SimPO change the treatment of the reference policy, preference margin, or regularisation~\citep{hong2024orpo,meng2024simpo}. We hold the DPO objective fixed to isolate hyperparameter transfer within one method and leave an extension of this analysis with alternate preference-optimisation objectives to future work. Prior work shows that DPO is sensitive to training data~\citep{ivison2024unpacking} and the joint choice of learning rate and $\beta$~\citep{wu2024beta,meng2024simpo}, while post-training recipes tune these values for their particular models and datasets~\citep{tunstall2023zephyr,ivison2023camels,lambert2024tulu}. We quantify how a jointly selected $(\mathrm{learning\ rate},\beta)$ configuration transfers across model scales and preference data, and whether the target model reveals a suitable configuration before training completes.

\paragraph{Hyperparameter transfer and scaling.}
Scaling laws characterize how training behaviour changes with model and data size~\citep{kaplan2020scaling,hoffmann2022training}, while $\mu$Transfer and related work study when hyperparameters tuned on smaller models predict larger-model behaviour~\citep{yang2022tensor,li2025steplaw}. DPO presents a different setting because it follows pretraining, depends on an SFT reference policy, and introduces preference-specific hyperparameters. We test whether an analogous learning-rate trend appears in DPO and assess whether configurations selected at one scale remain effective at another.

\paragraph{Evaluation signals and benchmark reliability.}
Post-training studies commonly use both fixed capability benchmarks and LLM-as-judge evaluations such as MT-Bench ~\citep{zheng2023judging}, AlpacaEval ~\citep{dubois2024length}, and Arena-Hard ~\citep{li2024arenahard}. LLM-as-judge evaluations have documented sensitivities including position bias ~\citep{wang2024large}, length bias ~\citep{dubois2024length}, and model self-preference ~\citep{panickssery2024llm}. Relatedly, FineTasks selects multilingual evaluation tasks using properties such as non-random performance, low variability, and stable model rankings across training checkpoints~\citep{kydlicek2024finetasks}. Our setting is closely related but addresses a distinct post-training question: 
\emph{whether a metric provides a reliable ranking signal for hyperparameter selection}. We examine whether validation loss, static capability benchmarks, and chat-based evaluations produce consistent rankings over DPO configurations, and whether rankings observed at intermediate checkpoints predict the final ordering of configurations. Evaluation is therefore treated not only as a reporting mechanism but as part of the HPO problem itself.

\section{Experimental Setup}
\label{sec:setup}

\paragraph{Models.}
We select Qwen3 as the primary model family because it provides competitive, text-only base checkpoints at multiple practical scales~\citep{qwen2025qwen3}. Our study uses the 0.6B, 1.7B, 4B, 8B, and 14B models, which span a $23\times$ parameter range within one family. This allows us to study how DPO hyperparameter selection changes with size while holding the model family and modality fixed. Complementary experiments on Llama models are reported in Appendix~\ref{app:llama-experiments}.
\paragraph{Data.}
We first perform SFT on Dolci-Instruct-SFT and then train with DPO on Dolci-Instruct-DPO~\citep{olmo2025olmo3}. To isolate the effects of training data, we also repeat the shared DPO grid on UltraFeedback Binarized~\citep{cui2023ultrafeedback} dataset for the same Qwen3 models. For this ablation, we retain the SFT checkpoint and evaluation suite while replacing the DPO training dataset. 

\paragraph{Training Protocol.}
\label{sec:training-protocol}
We use an SFT learning rate of $3\times10^{-5}$ for all primary runs, while comparing DPO configurations across model sizes. Appendix~\ref{app:sft-learning-rate} reports a five-point SFT learning rate sweep showing that this learning rate choice lies in a stable regime. Rankings vary across the five sizes and evaluation aggregates, and the highest learning rate tested consistently degrades the downstream performance. We leave joint optimisation of the SFT and DPO hyperparameters to future work.
DPO begins from the corresponding SFT checkpoint. Its search space contains a fully observed $3\times3$ core shared by every model: learning rates $ \in \{1\times10^{-6},\,2\times10^{-6},\,4\times10^{-6}\}$ and $\beta$ values $ \in \{0.01,\,0.02,\,0.04\}$. We extend this core where needed to cover each size's promising learning-rate range. The shared core supports like-for-like transfer and the size-specific extensions inform the local optimum. We save intermediate checkpoints and evaluate them alongside the final checkpoints. All other training settings are held fixed within each sweep. Per-size search spaces, checkpoint steps, training settings, and distributed training setups are reported in Appendix~\ref{app:search-spaces}. All post-training runs use \texttt{TRL} ~\citep{vonwerra2022trl}.

\paragraph{Evaluation.}

We evaluate eleven benchmarks in two groups. The chat group comprises Elo~\citep{chiang2024chatbotarenaopenplatform}, Arena-Hard~\citep{li2024arenahard}, MT-Bench~\citep{zheng2023judging}, and length-controlled AlpacaEval 2.0~\citep{dubois2024length}. These evaluations measure instruction following and response quality through LLM judgments. We score the chat responses separately with Qwen3-Next-80B-A3B-Instruct and Gemma 4 31B-it~\citep{qwen2025qwen3,gemmateam2026gemma4} to assess their sensitivity to judge choice~\citep{wang2024large,panickssery2024llm}. The main paper results use the Qwen model as the judge, unless specified otherwise. The static group comprises ARC-C~\citep{clark2018think}, GPQA~\citep{rein2023gpqa}, GSM8K~\citep{cobbe2021training}, HellaSwag~\citep{zellers2019hellaswag}, PIQA~\citep{bisk2020piqa}, TruthfulQA~\citep{lin2022truthfulqa}, and IFEval~\citep{zhou2023instruction}. We select these benchmarks to cover distinct capabilities commonly examined in language-model evaluation: scientific and mathematical reasoning, commonsense and physical reasoning, truthfulness, and instruction and format compliance. This diversity allows us to test whether a configuration that performs well on one fixed capability transfers to other capabilities, and to monitor capability-specific regressions during DPO. 
We also record validation loss as a candidate selection signal. The static benchmarks use \texttt{lm-evaluation-harness}~\citep{eval-harness}. The chat benchmarks use their corresponding benchmark implementations, including \texttt{JudgeArena} for MT-Bench and Elo ratings~\citep{lushtaku2026judgearena}, with candidate and judge models served through \texttt{vLLM}~\citep{kwon2023efficient}. Complete benchmark settings are reported in Appendix~\ref{app:benchmark-settings}.

\paragraph{Aggregation and Selection Criteria.}
\label{sec:aggregation}
For configuration $i$ and benchmark $b$, let $x_{i,b}$ denote the observed benchmark score for a specific configuration. Because the scales across benchmarks differ, we standardize scores as
$z_{i,b}=(x_{i,b}-\mu_b)/\sigma_b$, where $\mu_b$ and $\sigma_b$ are the mean and population standard deviation of benchmark $b$ across the relevant configuration pool. For an aggregate $G$ of benchmarks, we average these standardized scores as
$Z_{i,G}=\frac{1}{|G|}\sum_{b\in G} z_{i,b}$.
We use $G$ for the aggregation of chat benchmarks (\zdynamic{}), static benchmarks (\zstatic{}), or all benchmarks (\zall{}). Trajectory plots use final-grid statistics as a fixed reference, whereas checkpoint-selection analyses recompute them at each step. For configuration $i$ and aggregate $G$, we define normalised regret as 
$ \operatorname{Regret}_{i,G}=
\frac{\max_{j\in\mathcal{C}} Z_{j,G}-Z_{i,G}}
{\max_{j\in\mathcal{C}} Z_{j,G}-\min_{j\in\mathcal{C}} Z_{j,G}}$
where $\mathcal{C}$ is the set of jointly scored configurations used for the comparison: normally the full grid at one model size, and the shared $3\times3$ core at the target size for cross-scale transfer. Thus, regret is zero for the best configuration in $\mathcal{C}$ and one for the worst.

\paragraph{Offline HPO Simulation.}
Hyperparameter optimisation (HPO)~\citep{franceschi-ftml25} uses the outcomes and costs of earlier trials to decide which training configurations to evaluate next. Multi-fidelity HPO additionally uses partial training runs as cheaper, lower-fidelity signals when allocating further compute. Training progress is the fidelity variable. Each method optimises \zdynamic{} under matched budgets, and we evaluate its recommended configuration at true final fidelity.
At each Qwen3 size, we fit four surrogate blackboxes separately using a Gaussian process~\citep{seeger2004gaussian} and one-, three-, and five-nearest-neighbour regressors~\citep{hastie2009elements}. This offline-HPO setup enables repeated comparisons under matched budgets and random seeds, although its conclusions are conditional on the fitted interpolation models.
We compare three single-fidelity methods, random search~\citep{bergstra2012random}, TPE~\citep{bergstra2011algorithms}, and, lastly, CQR~\citep{salinas2023cqr}, which evaluates every proposed configuration to completion, with the multi-fidelity methods ASHA~\citep{li2020massively} and BOHB~\citep{falkner2018bohb}, which allocate compute using intermediate results from step 400 onward. Each setting uses two workers and 25 seeds under an $8\times$ wall-clock budget and a restrictive one-unit budget, corresponding to approximately two full-run equivalents of aggregate compute across both workers. At each cutoff, we evaluate the recommended configuration at final fidelity; before any trial completes, we compare candidates at their highest shared fidelity. The simulations use \texttt{syne-tune}~\citep{salinas2022syne}. Appendices~\ref{app:additional-related-work} and~\ref{app:hpo-settings} give additional related work and full HPO settings.

\section{Results}
\label{sec:results}

\subsection{Identifying The Metric To Tune On}
\label{sec:rq1}

\paragraph{Validation loss is an unreliable selection criterion.} 

\begin{figure}[h]
\centering
\includegraphics[width=0.9\textwidth]{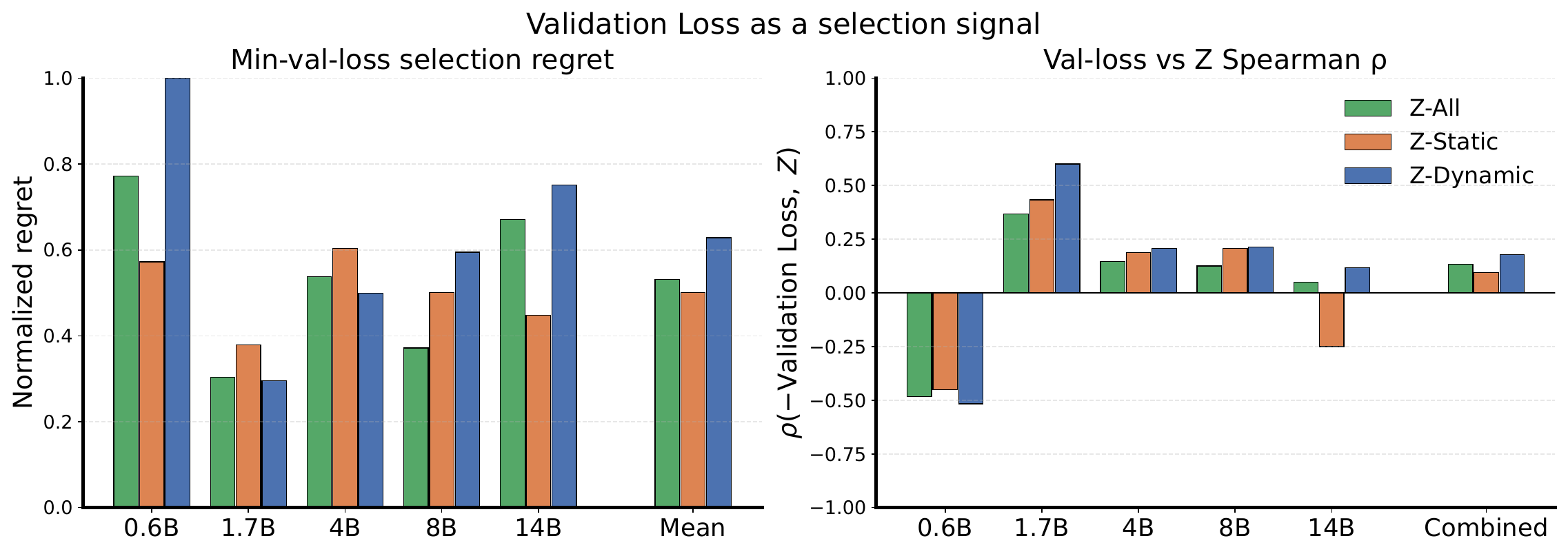}
\caption{\textbf{Validation loss is an unreliable DPO selection signal.}
\textbf{Left:} Normalised final-performance regret from selecting the configuration with lowest final validation loss at each Qwen3 size; zero denotes the best and one the worst configuration under the corresponding aggregate. \textbf{Right:} Spearman correlation between negative final validation loss and final \zstatic{}, \zdynamic{}, and \zall{} scores. Positive values indicate that lower loss is associated with better evaluation performance. The minimum-loss choice is never near-optimal, and correlations are weak and inconsistent across scales.}
\label{fig:validation-loss-selection}
\end{figure}

Validation loss is an insufficient proxy signal to select hyperparameters as it poorly correlates with downstream performance (see Figure~\ref{fig:validation-loss-selection}).
Selecting the configuration with minimum validation loss yields mean normalised regret of $0.50$, $0.62$, and $0.53$ under \zstatic{}, \zdynamic{}, and \zall{}, respectively. At 0.6B, it selects the worst configuration under \zdynamic{}. Pooled across sizes, rank correlations are only $0.10$, $0.18$, and $0.14$, and sometimes change sign across scales. Because DPO loss depends on $\beta$, we also rank learning rates separately within each fixed $\beta$. Agreement with \zall{} still varies across Qwen3 sizes ($\rho=-0.67$ to $0.67$; Appendix~\ref{app:within-beta-loss}). The underlying benchmark-level correlation matrices appear in Appendix~\ref{app:qwen3-benchmark-loss-correlations}.

\paragraph{Chat benchmarks produce more consistent configuration rankings.}
We measure the pairwise Spearman correlation between the configuration rankings produced by benchmarks within each evaluation group. Table~\ref{tab:qwen-benchmark-agreement} shows that mean agreement among static benchmarks is consistently weak across the five Qwen3 sizes, whereas mean agreement among chat benchmarks is consistently positive. Agreement also varies among benchmark pairs: the standard deviation ranges from $\sigma=0.10$ to $0.20$ for chat-benchmark pairs, compared with $\sigma=0.37$ to $0.64$ for static-benchmark pairs. Higher agreement does not establish that chat benchmarks measure ground-truth quality, but it indicates that hyperparameter selection based on the chat group is less sensitive to the particular benchmark chosen. Because the static suite intentionally spans heterogeneous capabilities, its lower agreement is partly expected and should not be interpreted as evidence that its individual benchmarks are invalid. Instead, the result shows that the DPO configuration preferred for one capability may not be preferred for another. The corresponding Llama and Qwen3 (UltraFeedback) analyses in Appendices~\ref{app:llama-experiments} and \ref{app:ultrafeedback} show the same pattern.

\begin{table}[h]
\centering
\small
\setlength{\tabcolsep}{5pt}
\caption{\textbf{Chat benchmarks produce more consistent DPO configuration rankings than static benchmarks on Qwen3.} Mean pairwise Spearman correlation and population standard deviation across within-group benchmark pairs, computed separately at each size over jointly scored configurations. Validation-loss columns are the mean $\rho$ of $(-\text{eval loss})$ vs.\ each benchmark in that group}
\label{tab:qwen-benchmark-agreement}
\begin{tabular}{lcccc}
\toprule
Qwen3 size & Chat ($\bar{\rho}\pm\sigma$) & Static ($\bar{\rho}\pm\sigma$) & Val.\ loss vs chat & Val.\ loss vs static \\
\midrule
0.6B  & $0.47\pm0.20$ & $ 0.03\pm0.54$ & $-0.19\pm0.10$ & $-0.18\pm0.36$ \\
1.7B  & $0.84\pm0.10$ & $ 0.27\pm0.55$ & $ 0.50\pm0.17$ & $ 0.38\pm0.37$ \\
4B    & $0.78\pm0.14$ & $ 0.18\pm0.37$ & $ 0.37\pm0.06$ & $ 0.13\pm0.08$ \\
8B    & $0.57\pm0.17$ & $ 0.01\pm0.64$ & $ 0.16\pm0.15$ & $ 0.01\pm0.34$ \\
14B   & $0.63\pm0.18$ & $-0.10\pm0.58$ & $ 0.47\pm0.20$ & $-0.08\pm0.40$ \\
\bottomrule
\end{tabular}
\end{table}

\begin{figure}[h]
    \centering
    \includegraphics[width=0.95\textwidth]{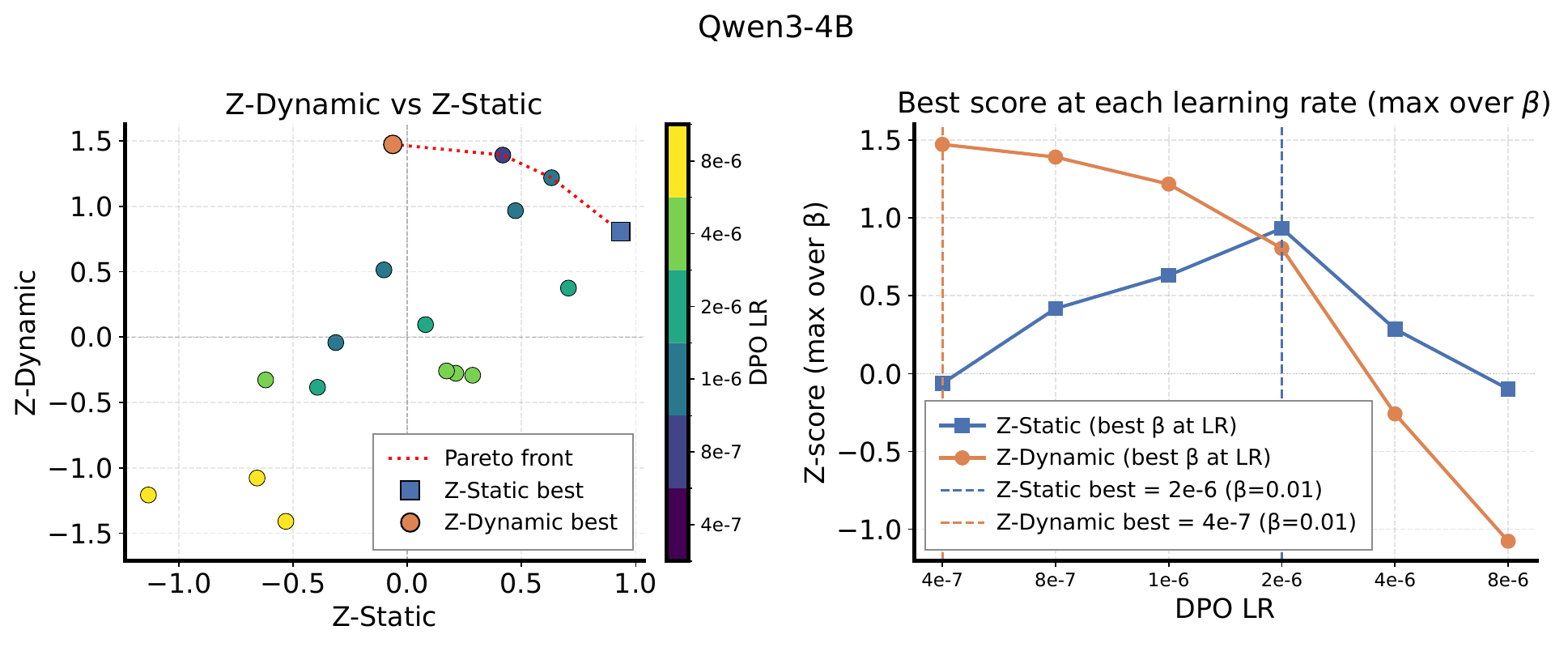}
    \caption{\textbf{Dynamic and static evaluations favour different Qwen3-4B DPO learning rates.}
    \textbf{Left:} Final \zdynamic{} and \zstatic{} scores. The red curve marks the Pareto frontier, and squares and circles identify configurations that maximize the \zstatic{} and \zdynamic{} scores. \textbf{Right:} Best score at each learning rate after maximizing over \(\beta\).}
    \label{fig:dynamic-static-tradeoff}
\end{figure}

\paragraph{Static and chat aggregates favour different configurations.}
At Qwen3-4B, maximizing \zstatic{} selects $(2\times10^{-6}, 0.01)$, whereas maximizing \zdynamic{} selects $(4\times10^{-7}, 0.01)$. Figure~\ref{fig:dynamic-static-tradeoff} shows that these configurations occupy different parts of the Pareto frontier. Thus, the appropriate learning rate depends on whether the intended target is static capability, chat quality, or an explicit aggregate of the two. Appendix~\ref{app:dynamic-static-tradeoffs} provides the corresponding Pareto comparisons for the other four Qwen3 sizes.

\paragraph{Independent judges largely agree on chat-performance rankings.}
To test sensitivity to judge choice, we compare the \zdynamic{} configuration rankings produced independently by Qwen3-Next-80B-A3B-Instruct and Gemma 4 31B-it over the nine-configuration learning-rate--$\beta$ core. Agreement is high for Qwen3-1.7B and larger models, with Spearman correlations of $\rho=0.90$, $0.98$, $0.90$, and $0.95$ at 1.7B, 4B, 8B, and 14B, respectively, but falls to $\rho=-0.27$ at 0.6B. This may reflect insufficiently distinguishable responses at the smallest scale. Within our setting, reliable chat-based model selection therefore emerges only above 0.6B, cautioning against using very small models as proxies for hyperparameter selection.
Because the candidate responses and evaluation procedures are held fixed while the judge model changes, this agreement makes it unlikely that the observed coherence is driven solely by idiosyncrasies of a single judge. It does not exclude biases shared across judges or effects of common prompts, aggregation procedures, and comparators. Complete aggregate and benchmark-level correlations appear in Appendix~\ref{app:judge-agreement}.

\subsection{The Hyperparameter Landscape Across Scale}
\label{sec:rq2}

\paragraph{The optimal Qwen3 configuration changes with scale.}

\begin{figure}
    \centering
    \includegraphics[width=0.9\textwidth]{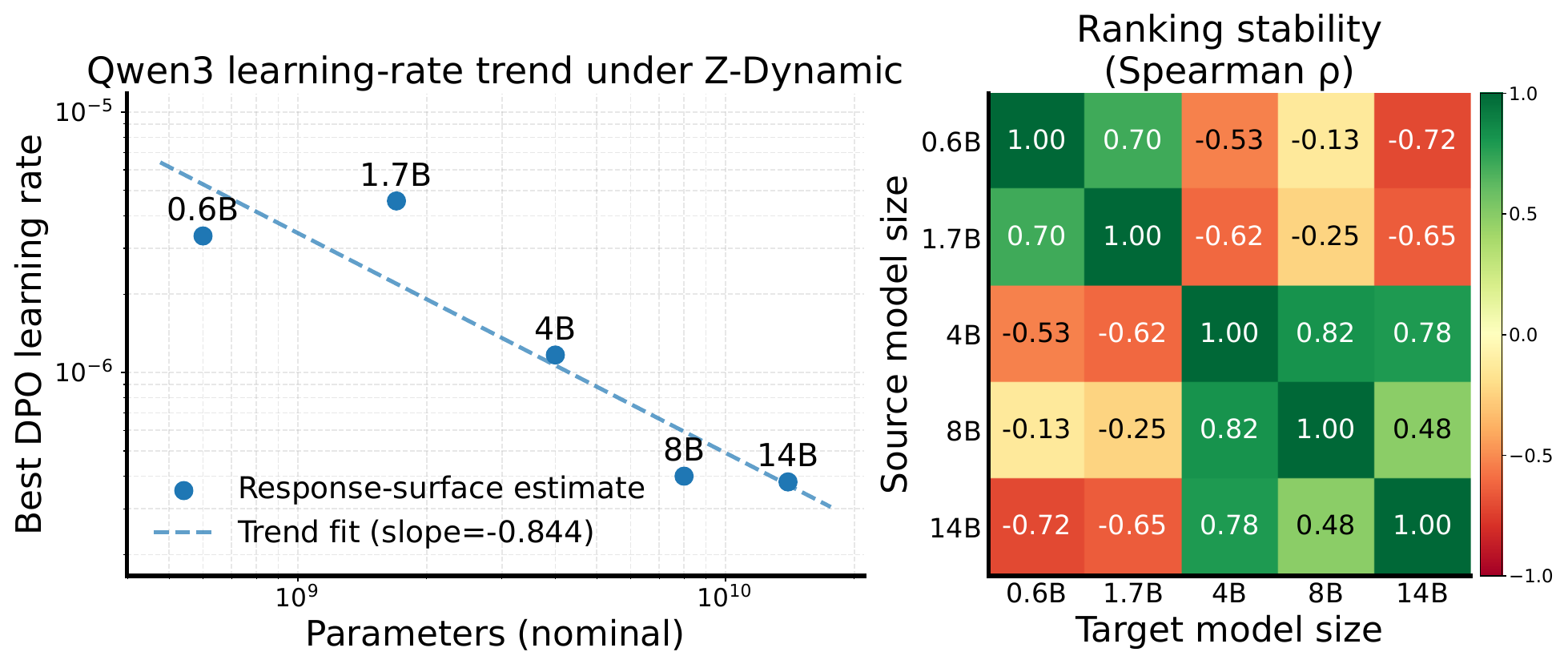}
    \caption{\textbf{Learning-rate trend and configuration-ranking stability under \zdynamic{} across Qwen3 sizes.}
    \textbf{Left:} Per-size response-surface estimates of the preferred learning rate and a descriptive fit across all five sizes. \textbf{Right:} Spearman correlations between rankings of the nine configurations in the shared ($3\times3$) grid. Positive values indicate similar configuration rankings and hence more plausible transfer, whereas negative values indicate ranking reversals. Rankings agree within the 0.6B--1.7B and 4B--14B groups but reverse across them.}
    \label{fig:scale-trend-and-ranking}
\end{figure}

Under \zdynamic{}, Figure~\ref{fig:scale-trend-and-ranking} shows the learning-rate estimates from two-dimensional quadratic response surfaces over $(\log_{10}\mathrm{LR},\log_{10}\beta)$ at each Qwen3 size. The estimates rise from 0.6B to 1.7B and then decline through 14B, showing an overall downward trend. We use the fitted line only as a descriptive summary of these five Qwen3 models, not as a monotonic scaling law or a predictor for an unseen size. Appendix~\ref{app:objective-specific-scaling} compares the corresponding estimates under \zall{}, \zstatic{}, and validation loss.

\paragraph{The selected configuration changes with the preference dataset.}
On Dolci-Instruct-DPO, the best observed \zall{} configuration uses $\beta=0.01$ at every Qwen3 size. We repeat the shared $3\times3$ grid with UltraFeedback at all Qwen3 sizes, retaining the corresponding Dolci-trained SFT checkpoint and evaluation suite. The \zall{} winners agree at 0.6B, differ only in $\beta$ at 1.7B, and differ in both coordinates at 4B--14B.
Dolci-to-UltraFeedback transfer regret rises from $0.00$ at 0.6B to $0.25$ at 1.7B and $0.67$ at 4B, reaching $1.00$ at both 8B and 14B, showing that the difference at larger models is consequential. Since $\beta=0.01$ is the lower Dolci boundary and $4\times10^{-6}$ is the upper UltraFeedback boundary, these are preferences within the tested grid rather than located optima. This comparison shows that dataset sensitivity varies across the tested scales; Appendix~\ref{app:ultrafeedback} gives the complete comparison.

\subsection{Cheaper Tuning and Transfer}
\label{sec:rq3}

\paragraph{Hyperparameter transfer is unreliable across scales and model families.}
The right panel of Figure~\ref{fig:scale-trend-and-ranking} shows pairwise Spearman correlations between the \zdynamic{} rankings of the nine configurations shared by every Qwen3 grid. The matrix forms two groups: rankings agree between 0.6B and 1.7B ($\rho=0.70$) and among the 4B--14B models ($\rho=0.48$--$0.82$), but all correlations across these groups are negative ($\rho=-0.72$ to $-0.13$). Transfer also depends on the metric used. Under \zall{} (Figure~\ref{fig:zall-vs-zmacro-transfer}), the small-model agreement strengthens ($\rho=0.93$) and cross-block correlations remain negative ($\rho=-0.78$ to $-0.25$), but agreement among the larger models largely disappears ($\rho=0.67$ between 4B and 8B, $0.23$ between 4B and 14B, and $-0.02$ between 8B and 14B).
Moreover, positive rank agreement does not guarantee low transfer regret. Model-size distance alone is therefore insufficient to determine whether a configuration will transfer.
Additional aggregate-specific, transfer, and cross-family analyses appear in Appendix~\ref{app:cross-scale-ranking-agreement}--\ref{app:cross-family-transfer}.

\paragraph{Partial runs provide an informative but scale-dependent selection signal.}
At each Qwen3 size, we compare the intermediate and final \zall{} rankings over configurations evaluated at both fidelities. After 600 steps ($\approx 30\%$ of training), their Spearman correlations are $0.85$, $0.90$, $0.85$, $0.76$, and $0.63$ from 0.6B to 14B. At 4B, selecting from the step-600 scores yields zero final \zall{} regret, while directly transferring the configuration selected at 1.7B incurs maximal regret ($1.00$). A short target-model run can therefore provide a more useful selection signal than transfer from a neighbouring scale. Under UltraFeedback, dynamic-ranking correlations at step 144, roughly $30\%$ of training, are $0.85$, $0.72$, $0.75$, $0.67$, and $0.65$ from 0.6B to 14B. Because agreement can fluctuate across checkpoints, we use early evaluations to prioritize promising configurations rather than as a universal stopping rule. The complete Qwen3 trajectories on Dolci-Instruct-DPO, UltraFeedback replications, and Llama trajectories are reported in Appendices~\ref{app:full-qwen3-results}, \ref{app:ultrafeedback}, and \ref{app:llama-experiments}.

\begin{figure}[h]
    \centering
    \includegraphics[width=0.95\linewidth]{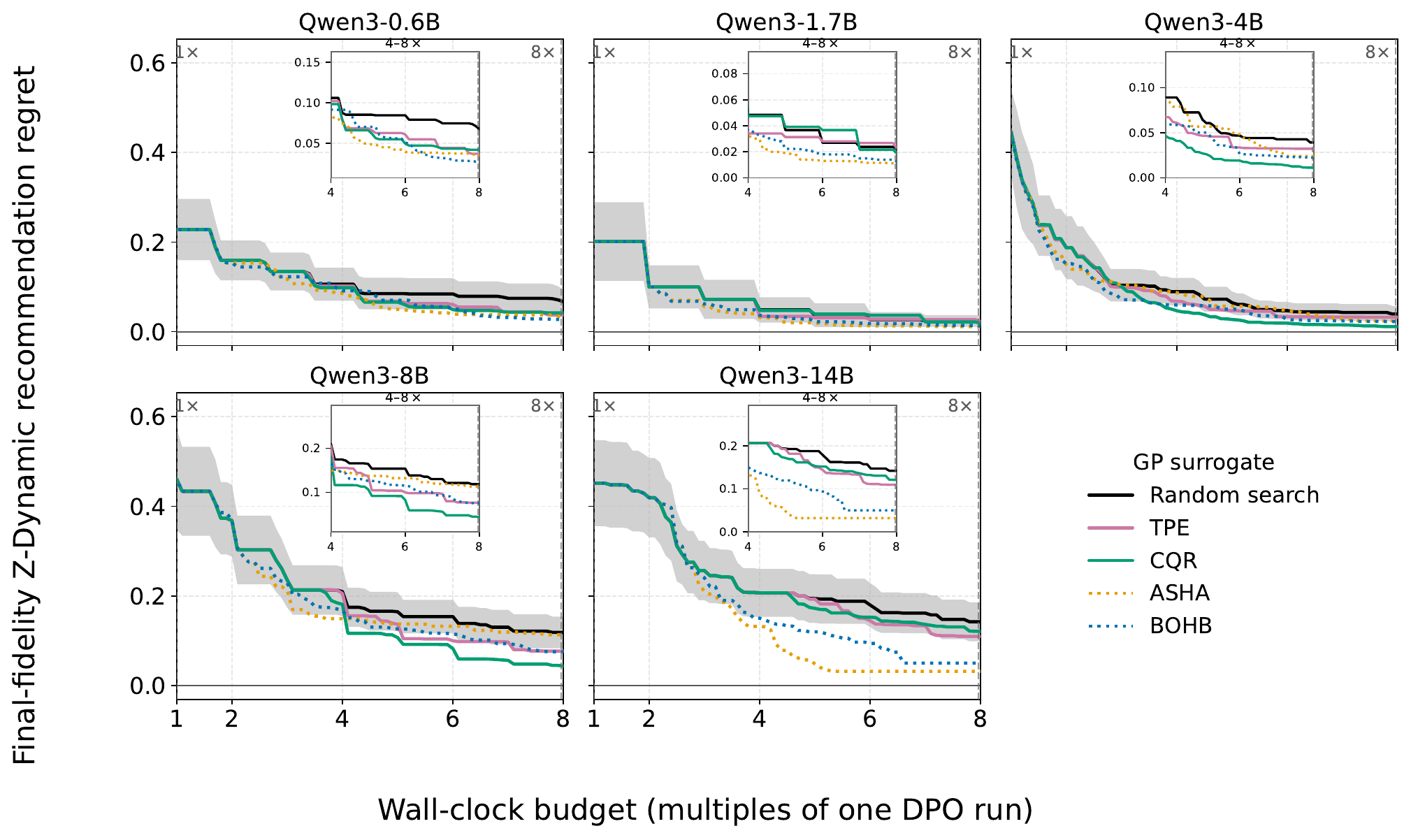}
    \caption{\textbf{HPO recommendation regret over time under the GP surrogate.}
    Each panel shows final-fidelity \zdynamic{} recommendation regret for one Qwen3 size as a function of normalised wall-clock budget; lower is better. Lines show the mean and the shaded region shows the 95\% $t$-interval for random search across 25 seeds. Results for other surrogates appear in Appendix~\ref{app:cross-metric-hpo}.}
    \label{fig:hpo-regret-over-time}
\end{figure}

\paragraph{Adaptive search methods outperform random search when given sufficient budget.}
We compare five search strategies over continuous learning-rate and $\beta$ domains using four surrogate blackboxes fitted separately to the Qwen3 trajectories: a Gaussian process (GP) and one-, three-, and five-nearest-neighbour (1-NN, 3-NN, and 5-NN) regressors. Each method optimises \zdynamic{}, and we evaluate its recommendation at final fidelity. Under the $8\times$ budget, CQR is the only method with a conclusive improvement over random search under all four surrogates (Table~\ref{tab:hpo-generous-gains}). BOHB has the largest GP gain, whereas CQR has the largest gain under each NN surrogate. TPE and BOHB improve conclusively under the GP, 3-NN, and 5-NN surrogates, while ASHA does so under the GP and 3-NN surrogates.
Figure~\ref{fig:hpo-regret-over-time} shows how GP regret decreases with budget at each model size. The gains are uneven across sizes: separation from random search is largest at 8B and 14B and nearly absent at 1.7B. Thus, the relative advantage of an HPO method depends on the response surface and the model size, although CQR is the most consistent method in these simulations. Under the restrictive budget, no method differs detectably from random search because the budget generally expires before adaptive proposals begin.
Complete results and regret curves appear in Appendix~\ref{app:cross-metric-hpo}.

\begin{table}
\centering
\footnotesize
\setlength{\tabcolsep}{5pt}
\caption{\textbf{Final \zdynamic{} improvement over random search under the $8\times$ wall-clock budget.} Differences are paired by seed after averaging the five Qwen3 sizes within each of 25 seeds. Bold marks the largest mean improvement for each surrogate; $\dagger$ denotes a paired 95\% interval excluding zero.}
\label{tab:hpo-generous-gains}
\begin{tabular}{lrrrr}
\toprule
Surrogate & TPE & CQR & ASHA & BOHB \\
\midrule
Gaussian process & $+0.05^{\dagger}$ & $+0.08^{\dagger}$ & $+0.08^{\dagger}$ & $\mathbf{+0.10}^{\dagger}$ \\
$1$-NN & $+0.04$ & $\mathbf{+0.09}^{\dagger}$ & $+0.01$ & $+0.03$ \\
$3$-NN & $+0.06^{\dagger}$ & $\mathbf{+0.11}^{\dagger}$ & $+0.04^{\dagger}$ & $+0.05^{\dagger}$ \\
$5$-NN & $+0.09^{\dagger}$ & $\mathbf{+0.16}^{\dagger}$ & $+0.02$ & $+0.06^{\dagger}$ \\
\bottomrule
\end{tabular}
\end{table}

\section{Practical Recommendations}
\label{sec:recommendations}

\paragraph{Select configurations using downstream evaluations.}
Do not use validation loss alone to choose a DPO configuration: its minimum can be far from the best downstream configuration. Nor should a single downstream benchmark determine the choice, since benchmark rankings often disagree. When the goal is conversational response quality, select using multiple chat benchmarks rather than a single metric, and check individual static benchmarks for regressions. Where two judges yield different configuration rankings, treat the selected configuration as evaluation-dependent and include targeted human assessment before making consequential deployment decisions.

\paragraph{Retune for the target setting.} 
Treat a configuration from another model or preference dataset as a starting point for a small target-specific search, not as a default. Search learning rate and $\beta$ jointly: model-size distance does not reliably predict whether configurations retain their ranking, and changing the preference dataset can alter the selected configuration, with substantial transfer regret at larger model sizes. If the selected value lies at the edge of the tested range, extend the range before interpreting it as a well-localized optimum.

\paragraph{Use partial target-model runs before committing full training.}
Configuration rankings become informative within 20--30\% of training  across the model families and preference-data settings studied, making early checkpoints useful for narrowing the candidate set, although they do not guarantee the final winner at every scale. When using adaptive HPO, allocate enough compute for the method to move beyond its initialization phase. Under our larger budget, CQR provides the most consistent improvement, while TPE, ASHA, and BOHB also outperform random search. Because gains vary across model sizes, validate the allocation strategy on the target setting. Our compute budgets include training but exclude evaluations.

\section{Limitations}
\label{sec:limitations}

\paragraph{Limited cross-family and preference-data coverage.}
The five-size scaling analysis is confined to Qwen3. The three Llama checkpoints span two releases rather than a homogeneous scaling series, so they provide qualitative tests of benchmark agreement and transfer but cannot establish whether the Qwen3 learning-rate trend generalizes across families. We test preference-data sensitivity only on one alternative dataset within one model family, by replacing Dolci-Instruct-DPO with UltraFeedback Binarized while retaining the SFT checkpoint and evaluation suite. One dataset pair cannot establish how common the resulting reversal in preferred learning rate is. The selected Dolci-Instruct learning rates and $\beta$ values should not be treated as defaults for other datasets.

\paragraph{Finite grids, offline search, and training variance.}
We select among logged configurations whose coverage differs across sizes; shared-core transfer need not equal full-grid transfer, and boundary selections need not locate unconstrained optima. The Qwen3-only HPO experiments interpolate logged trajectories through four fitted response surfaces rather than adding DPO runs. Each DPO configuration uses one training seed, and benchmark bootstrapping and 25 HPO seeds measure evaluation-composition and search variation, but not training-seed variance.

\paragraph{Evaluation sensitivity.}
At Qwen3-0.6B, the two judges provide conflicting rankings, possibly because small response differences are especially judge-sensitive at this scale. Arena-Hard and MT-Bench also compare configurations with matched SFT checkpoints, which may introduce comparator-specific judge preferences. This limits conclusions from individual metrics or judges.

\section{Conclusion}
\label{sec:conclusions}

Our findings show that DPO hyperparameter selection depends on the target model, preference data, and evaluation objective. Validation loss can select configurations far from the best-performing downstream configuration, static benchmarks disagree about configuration rankings, and neither parameter-count distance nor a configuration selected on another preference dataset reliably predicts transfer quality. Although the selected Qwen3 learning rate exhibits an overall decline with scale, its small-scale rise and the boundary selection of $\beta$ argue against a universal configuration or simple transfer rule. Partial target-model runs provide useful ranking information within $20$--$30\%$ of training across the model families and training data settings studied, offering a lower-cost way to narrow the candidate set before full training.
Continuous-surrogate HPO experiments show that informed search can improve final configuration selection when the $8\times$ budget permits adaptive proposals.
The released trajectories provide an offline benchmark for studying this trade-off between selection quality and training cost.

\subsection*{AI use statement}

We used generative AI tools to help polish the writing and presentation of this paper. During experimentation, generative AI was used to help prepare Slurm submission scripts and YAML configuration files; all generated files were manually inspected and verified before use. AI was not used to generate the remaining training and HPO simulation code, which was drawn from their respective mentioned libraries. For evaluation and analysis, generative AI was used to help write plotting code that produced figures from the experimental data. We manually checked the resulting plots and their underlying values for correctness. The authors take full responsibility for the paper and for all artifacts produced with AI assistance.

\subsection*{Reproducibility statement}

All training and evaluation settings are documented in Section~\ref{sec:setup} and the Appendix, including the per-size search spaces, optimisation settings, evaluation protocols, and judge configurations. We release a checkpoint-level offline benchmark: \url{https://github.com/HannanJaved/dpo-hpo-blackbox}. It contains checkpoint-level evaluation data for all Qwen3 (including the UltraFeedback Binarized replications) and Llama DPO experiments, including training fidelity, elapsed training cost, and scores on the eleven downstream benchmarks. Producing each entry requires a full DPO training run and a costly downstream evaluation suite. The completed DPO runs consumed approximately 8,454 H100 GPU-hours and evaluations (including intermediate checkpoints and two judges) consumed 3,395 H100 GPU-hours; Appendix~\ref{app:compute-accounting} provides the breakdown and accounting scope of the 11,849 total H100 GPU-hours used. The release enables future work to compare selection signals, early-selection policies, transfer rules, and HPO methods without repeating that compute. The blackbox tables additionally provide \zstatic{}, \zdynamic{}, \zall{}, and \zmacro{} aggregates. The artifact includes manifests, the HPO protocol, and code to reconstruct the Qwen3 blackboxes and surrogate environments. It also provides complete Qwen3 HPO traces across five model sizes, four surrogates, two budgets, three objectives, and five optimisers. The training checkpoints and cluster logs are not included in the artifact due to storage limitations.

\section*{Acknowledgments}
This research was partially funded by the European Commission under the grant No. 10119523 (OpenEuroLLM).
The authors gratefully acknowledge the computing time made available to them on the high-performance computer at the NHR Center of TU Dresden. This center is jointly supported by the Federal Ministry of Research, Technology and Space of Germany and the state governments participating in the NHR. 

\begin{center}
 \includegraphics[width=0.3\textwidth]{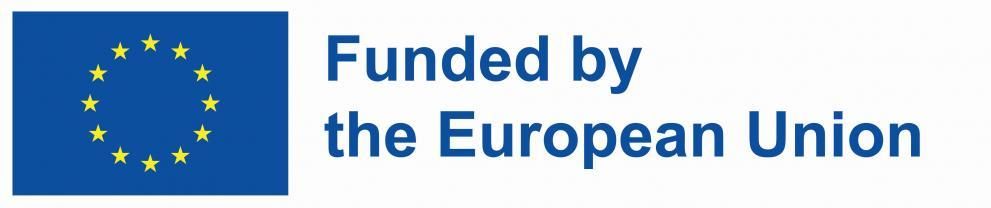}
\end{center}

\bibliography{iclr2027_conference}
\bibliographystyle{iclr2027_conference}

\appendix

\section{Additional Related Work}
\label{app:additional-related-work}

\paragraph{Hyperparameter optimisation and multi-fidelity search.}
Hyperparameter optimisation (HPO) seeks a training configuration that performs well while minimizing the cost of evaluating candidate configurations. In single-fidelity HPO, each candidate is trained to completion before the next informed choice is taken. Bayesian optimisation (BO) instead fits a model to completed trials and uses it to propose new configurations~\citep{franceschi-ftml25}. Multi-fidelity HPO additionally treats partial training runs as cheaper, lower-fidelity evaluations and allocates further training only to candidates that seem promising under the chosen evaluation objective. This is the case in Hyperband, ASHA, and BOHB~\citep{li2018hyperband,li2020massively,falkner2018bohb}. We use logged DPO trajectories to compare random search~\citep{bergstra2012random}, TPE~\citep{bergstra2011algorithms}, CQR~\citep{salinas2023cqr}, ASHA~\citep{li2020massively}, and BOHB~\citep{falkner2018bohb} under matched training-time budgets, treating training progress as fidelity and downstream performance as the objective.

\paragraph{HPO benchmark suites.}
Benchmark suites such as HPOBench and YAHPO Gym support controlled, reproducible comparisons of HPO methods on expensive multi-fidelity problems~\citep{eggensperger2021hpobench,pfisterer2022yahpo}. HPOBench packages reproducible benchmark problems, whereas YAHPO Gym uses fitted surrogates to make repeated optimiser comparisons inexpensive. Our released benchmark applies this paradigm to DPO. It provides checkpoint-level trajectories, measured training costs, and downstream evaluation outcomes from expensive post-training runs. It therefore enables future work to compare evaluation signals, early-selection policies, transfer rules, and HPO allocation strategies without repeating the underlying DPO training.

\section{Dataset Details}
\label{app:dataset-details}

\paragraph{Dataset versions and splits.}
Our primary post-training recipe uses the \texttt{train} splits of \texttt{allenai/Dolci-Instruct-SFT} for supervised fine-tuning and \texttt{allenai/Dolci-Instruct-DPO} for preference optimisation~\citep{olmo2025olmo3}. The same datasets and preprocessing are used for all Qwen3 and Llama experiments. We split Dolci-Instruct-DPO with \texttt{test\_size=0.005} and seed 42, yielding 258,622 training pairs and a fixed held-out validation set of 1,300 pairs. The Qwen3 preference-data ablations instead use the \texttt{train\_prefs} split of \texttt{HuggingFaceH4/ultrafeedback\_binarized}~\citep{cui2023ultrafeedback}, while retaining the same Dolci-trained SFT checkpoint. Of its 61,135 preference pairs, 60,000 are used for training and 1,135 for validation. Table~\ref{tab:dataset-summary} summarizes the resulting example counts.

\begin{table}[h]
\centering
\small
\setlength{\tabcolsep}{8pt}
\caption{\textbf{Datasets used for post-training.} Hub-split sizes and the numbers retained for training and validation after preprocessing. Repository names and configurations are given in the text.}
\label{tab:dataset-summary}
\begin{tabular}{lrrr}
\toprule
Dataset & Hub split & Training & Validation \\
\midrule
Dolci-Instruct-SFT & 2,152,112 & 1,944,341 & --- \\
Dolci-Instruct-DPO & 259,922 & 258,622 & 1,300 \\
UltraFeedback & 61,135 & 60,000 & 1,135 \\
\bottomrule
\end{tabular}
\end{table}

\paragraph{SFT formatting and filtering.}
SFT examples are represented as \texttt{messages} and formatted with a custom OLMo-style ChatML template using \texttt{<|im\_start|>} and \texttt{<|im\_end|>} delimiters. The template includes a default system prompt supporting function calling and is saved with each SFT checkpoint as \texttt{chat\_template.jinja}. We train with assistant-only loss, masking user and system tokens, and do not use sequence packing. Before training, we discard examples with a missing or invalid \texttt{messages} field or with \texttt{content=None} in any turn, reducing Dolci-Instruct-SFT from 2,152,112 to 1,944,341 examples. We apply no further deduplication or subsampling. The maximum SFT sequence length is 8,192 tokens for all Qwen3 and all Llama models; longer conversations are truncated by \texttt{SFTTrainer}.

\paragraph{DPO formatting and truncation.}
The DPO runs load the tokenizer and chat template stored with their corresponding SFT checkpoint. For Dolci-Instruct-DPO, the \texttt{chosen} and \texttt{rejected} conversations are retained while turn-level metadata is removed, leaving only \texttt{role} and \texttt{content}; any existing \texttt{prompt} or \texttt{messages} columns are removed so that TRL reconstructs the common prompt from the two conversations. The same processing is applied to UltraFeedback after retaining only its \texttt{chosen} and \texttt{rejected} columns. Dolci-Instruct-DPO uses a maximum sequence length of 4,096 tokens and a maximum prompt length of 512 tokens. UltraFeedback uses 2,048 and 1,024 tokens, respectively. Both use TRL's default \texttt{keep\_end} truncation mode, which retains the end of an overlength prompt and truncates the response as needed to satisfy the total sequence-length limit. Beyond constructing the validation splits, we apply no additional filtering, deduplication, or subsampling to either DPO dataset.

\section{Additional Details On The Experimental Setup}
\subsection{Training Configurations}
\label{app:training-configs}

\paragraph{Supervised fine-tuning.}
We sweep SFT learning rates and use $3\times10^{-5}$ for the SFT checkpoint that initializes every primary DPO run. SFT is run for two epochs, corresponding to approximately 30,382 optimiser steps, with a global batch size of 128. The per-device batch size is one. Most runs use gradient accumulation over 16 steps on eight GPUs; Qwen3-14B uses eight accumulation steps on 16 GPUs, preserving the same global batch size. We optimise with \texttt{adamw\_torch}, Adam coefficients $(\beta_1,\beta_2)=(0.9,0.999)$, $\epsilon=10^{-8}$, no weight decay, and maximum gradient norm 1.0. The learning rate follows a linear schedule with a 0.03 warmup ratio. Training uses bfloat16 precision and seed 42. Checkpoints are saved every 3,000 steps for Qwen3 and Llama models.

\paragraph{Direct preference optimisation.}
The primary DPO grids vary learning rate and $\beta$ while holding the remaining settings fixed. Each configuration is trained for one epoch, with the logged trajectories ending at approximately 2,031 optimiser steps. We use a per-device batch size of one on eight GPUs and 16 gradient-accumulation steps, giving a global batch size of 128. The optimiser is \texttt{adamw\_torch} with the same Adam coefficients, numerical stability constant, zero weight decay, and maximum gradient norm as SFT. We use a linear learning-rate schedule with a 0.1 warmup ratio, bfloat16 precision, seed 8, the standard sigmoid DPO loss, and zero label smoothing. Checkpoints are saved every 200 steps, and the final checkpoint is also evaluated. Downstream evaluations are run separately on these saved checkpoints.

The Qwen3 UltraFeedback experiments use the same DPO recipe, except for the sequence-length settings reported in Appendix~\ref{app:dataset-details} and 32 gradient-accumulation steps. With four GPUs, this gives a global batch size of 128. Its one-epoch runs contain approximately 469 optimiser steps; for the trajectory analysis, we evaluate checkpoints every 48 steps and the final checkpoint.

\begin{table}
\centering
\small
\setlength{\tabcolsep}{7pt}
\caption{\textbf{Shared SFT and DPO optimisation settings.} Learning rate and $\beta$ are the variables swept during DPO; their per-model search spaces are given in Table~\ref{tab:dpo-search-spaces}. UltraFeedback uses the primary DPO values except where stated in the text.}
\label{tab:training-settings}
\begin{tabular}{lcc}
\toprule
Setting & SFT & Primary DPO \\
\midrule
Epochs & 2 & 1 \\
Learning rate & $3\times10^{-5}$ & Swept \\
$\beta$ & --- & Swept \\
Per-device batch size & 1 & 1 \\
Global batch size & 128 & 128 \\
Optimiser & \texttt{adamw\_torch} & \texttt{adamw\_torch} \\
Adam $(\beta_1,\beta_2)$; $\epsilon$ & $(0.9,0.999)$; $10^{-8}$ & $(0.9,0.999)$; $10^{-8}$ \\
Scheduler; warmup ratio & linear; 0.03 & linear; 0.10 \\
Weight decay; maximum gradient norm & 0; 1.0 & 0; 1.0 \\
Precision & bfloat16 & bfloat16 \\
Seed & 42 & 8 \\
Checkpoint interval & 3,000 steps & 200 steps \\
\bottomrule
\end{tabular}
\end{table}

\paragraph{Hardware and distributed training.}
All training runs use NVIDIA H100 GPUs and are executed through Slurm with four GPUs per node. Table~\ref{tab:distributed-training} reports the number of GPUs and distributed strategy used for each model group. The primary Dolci-Instruct DPO runs use eight GPUs, the UltraFeedback ablation uses four, and the largest SFT run, Qwen3-14B, uses 16 GPUs.

\subsubsection{Compute Accounting}
\label{app:compute-accounting}

\begin{table}[h]
\centering
\small
\setlength{\tabcolsep}{3.5pt}
\caption{\textbf{H100 GPU-hours for the reported DPO experiments.}
Training and evaluation columns include completed jobs only. 'Dolci' denotes Dolci-Instruct-DPO and 'UF' denotes UltraFeedback Binarized.}
\label{tab:training-compute}
\begin{tabular}{lcrrr}
\toprule
Series & Size & Training & Evaluation & Total \\
\midrule
Qwen3/Dolci & 0.6B & 234   & 410 & 644 \\
& 1.7B & 315   & 356 & 671 \\
& 4B   & 1,787 & 723 & 2,510 \\
& 8B   & 1,578 & 500 & 2,078 \\
& 14B  & 2,809 & 614 & 3,423 \\
\cmidrule(lr){2-5}
& Subtotal & 6,723 & 2,603 & 9,326 \\
\addlinespace
Llama 3/Dolci & 1B & 153 & 49  & 202 \\
& 3B & 300 & 138 & 438 \\
& 8B & 771 & 151 & 922 \\
\cmidrule(lr){2-5}
& Subtotal & 1,224 & 338 & 1,562 \\
\addlinespace
Qwen3/UF & 0.6B & 52 & 66  & 118 \\
& 1.7B & 49 & 68  & 117 \\
& 4B   & 72 & 112 & 184 \\
& 8B & 169 & 86 & 255 \\
& 14B & 165 & 122 & 287 \\
\cmidrule(lr){2-5}
& Subtotal & 507 & 454 & 961 \\
\midrule
\multicolumn{2}{r}{Reported-workload total}
& 8,454 & 3,395 & 11,849 \\
\bottomrule
\end{tabular}
\end{table}

We estimate compute as the number of allocated NVIDIA H100 GPUs multiplied by job wall-clock time. Table~\ref{tab:training-compute} reports training and benchmark-evaluation compute for the model and preference-data combinations analysed in this paper. Each distinct evaluation is counted once; when an evaluation was resubmitted, we retain only the latest completed job. Evaluation compute includes LM Evaluation Harness tasks, response generation, AlpacaEval, LLM-judge scoring, and Elo estimation at final and intermediate checkpoints. Distinct evaluations account for approximately 29\% of the reported training-and-evaluation workload. Jobs that were cancelled/crashed mid-run, SFT runs, and offline HPO simulation are excluded.

\begin{table}
\centering
\small
\caption{\textbf{Distributed training configuration.} Each node contributes four GPUs. DDP denotes distributed data parallel training~\citep{li2020pytorch}; ZeRO stages refer to DeepSpeed parameter and optimiser sharding~\citep{rasley2020deepspeed}.}
\label{tab:distributed-training}
\begin{tabular}{lcccc}
\toprule
Models & SFT GPUs & SFT strategy & DPO GPUs & DPO strategy \\
\midrule
Qwen3-0.6B and Qwen3-1.7B & 8 & DDP & 8 & DDP \\
Qwen3-4B & 8 & DDP & 8 & ZeRO-3 \\
Qwen3-8B & 8 & ZeRO-2 & 8 & ZeRO-3 \\
Qwen3-14B & 16 & ZeRO-3 & 8 & ZeRO-3 \\
Llama-3.2-1B and Llama-3.2-3B & 8 & DDP & 8 & DDP \\
Llama-3.1-8B & 8 & ZeRO-3 & 8 & ZeRO-3 \\
Qwen3-0.6B (UltraFeedback) & --- & --- & 4 & DDP \\
Qwen3-1.7B (UltraFeedback) & --- & --- & 4 & DDP \\
Qwen3-4B (UltraFeedback) & --- & --- & 4 & DDP \\
Qwen3-8B (UltraFeedback) & --- & --- & 4 & ZeRO-3 \\
Qwen3-14B (UltraFeedback) & --- & --- & 4 & ZeRO-3 \\
\bottomrule
\end{tabular}
\end{table}

\subsection{SFT Learning-Rate Sweep}
\label{app:sft-learning-rate}
\begin{figure}
\centering
\includegraphics[width=0.6\linewidth]{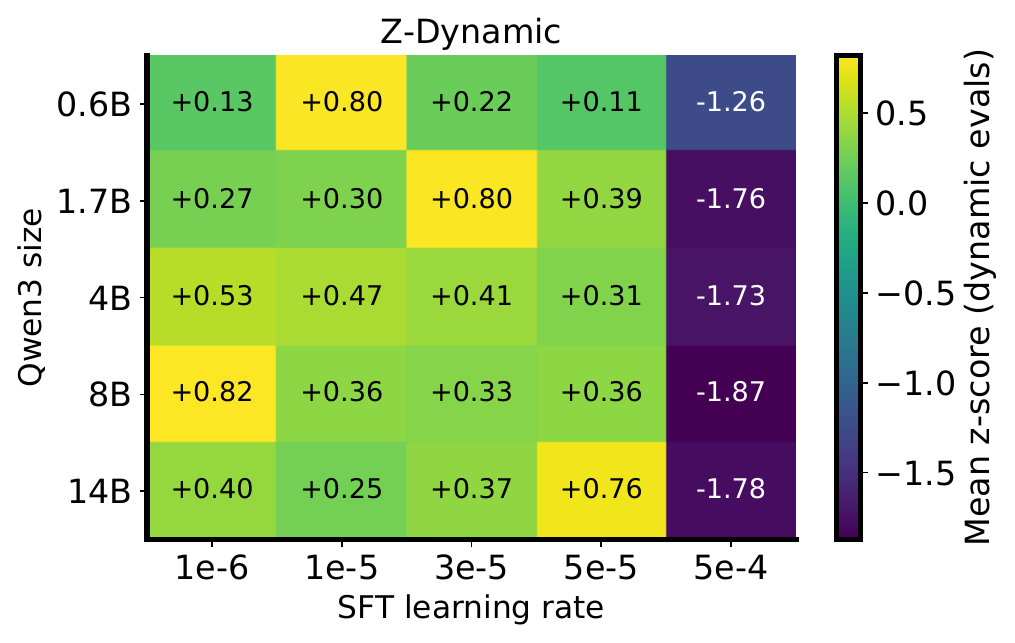}
\caption{\textbf{SFT learning-rate sensitivity on dynamic evaluations.} Mean \zdynamic{} score after SFT, standardized separately within each Qwen3 size across the five SFT learning rates. Rankings vary by model size, but $5\times10^{-4}$ is consistently poor.}
\label{fig:sft-lr-sweep-zdynamic}
\end{figure}

We evaluate SFT learning rates $\in \{10^{-6}, 10^{-5}, 3\times10^{-5}, 5\times10^{-5}, 5\times10^{-4}\}$ for each Qwen3 model size. Scores are standardized separately within each model size and evaluation aggregate. No single moderate learning rate value dominates across all sizes or across our aggregates (\zall{}, \zstatic{}, \zdynamic{}), with the relative rankings varying with both scale and objective. We therefore hold the SFT learning rate fixed at $3\times10^{-5}$ in the primary DPO experiments, rather than selecting a separate SFT optimum per size, which would
confound cross-size DPO comparisons. The sweep serves as a sensitivity check that this choice lies in the
non-degenerate plateau rather than as a claim that $3\times10^{-5}$ is
uniquely optimal.
By contrast, $5\times10^{-4}$ consistently yields substantially worse downstream scores, especially on \zdynamic{} (Figure~\ref{fig:sft-lr-sweep-zdynamic}).

\subsection{Search Spaces per Model Scale}
\label{app:search-spaces}

Every primary grid contains the nine configurations in the shared core
\begin{equation*}
\{1,2,4\}\!\times\!10^{-6}\ \text{(learning rate)}
\quad\times\quad
\{0.01,0.02,0.04\}\ \text{($\beta$)}.
\end{equation*}
This common core enables like-for-like transfer comparisons across Qwen3 sizes and between Qwen3 and Llama. We extend the grid at individual Qwen3 sizes to cover the locally promising learning-rate range and, at 4B, an additional $\beta$ value. Table~\ref{tab:dpo-search-spaces} gives the exact configurations. Cross-scale and cross-family transfer are evaluated on the shared core, whereas full-grid optima and response-surface analyses use every available configuration at the corresponding size.

\begin{table}
\centering
\small
\setlength{\tabcolsep}{6pt}
\caption{\textbf{DPO search spaces.} ``Core'' denotes the shared $3\times3$ Cartesian product defined above. Added configurations extend this support for a particular model or dataset.}
\label{tab:dpo-search-spaces}
\begin{tabular}{lll}
\toprule
Model & DPO search space & \# Configurations \\
\midrule
Qwen3-0.6B & Core $+\,(8\times10^{-6},0.01)$ & 10 \\
Qwen3-1.7B & Core $+\,(8\times10^{-6},0.01)$ & 10 \\
Qwen3-4B & Core + size-specific extensions & 18 \\
Qwen3-8B & $\{0.4,0.8,1,2,4\}\times10^{-6}\times\{0.01,0.02,0.04\}$ & 15 \\
Qwen3-14B & Core + size-specific extensions & 17 \\
Llama-3.2-1B & Core & 9 \\
Llama-3.2-3B & Core & 9 \\
Llama-3.1-8B & Core & 9 \\
Qwen3-0.6B (UltraFeedback) & Core & 9 \\
Qwen3-1.7B (UltraFeedback) & Core & 9 \\
Qwen3-4B (UltraFeedback) & Core & 9 \\
Qwen3-8B (UltraFeedback) & Core & 9 \\
Qwen3-14B (UltraFeedback) & Core & 9 \\
\bottomrule
\end{tabular}
\end{table}

Beyond the shared core, Qwen3-4B adds $\beta=0.08$ at learning
rates $1$, $2$, and $4\times10^{-6}$; all four tested $\beta$ values
at $8\times10^{-6}$; and $(4\times10^{-7},0.01)$ and
$(8\times10^{-7},0.01)$. Qwen3-14B adds
$(4\times10^{-7},\beta)$ and $(8\times10^{-7},\beta)$ for
$\beta\in\{0.01,0.02,0.04\}$, plus $(1\times10^{-7},0.01)$ and
$(8\times10^{-8},0.01)$.

For the Dolci-Instruct grids, we evaluate checkpoints every 200 steps and at the final training step. For UltraFeedback, we evaluate checkpoints every 48 steps and at the final step. These intervals correspond to evaluations at approximately $10\%$ intervals of training progress for both datasets. These trajectories support the early-selection and offline HPO analyses. The validation loss directed Bayesian-optimisation sweeps use continuous learning-rate and $\beta$ domains and are described separately in Appendix~\ref{app:validation-loss-bo}.

\subsection{Evaluation Benchmark Settings} 
\label{app:benchmark-settings}

\subsubsection{Static Benchmarks}
We evaluate static tasks with \texttt{lm-evaluation-harness}. Model weights are loaded in bfloat16 precision, and log-likelihood tasks use scaled dot-product attention. Table~\ref{tab:static-evaluation-settings} reports the splits, sample counts, few-shot settings, and metrics; inference details are given immediately below it. Except for IFEval, these tasks are evaluated as raw-text completions without applying the model's chat template, matching their standard harness protocols and facilitating comparison with previously reported results.

\begin{table}
\centering
\small
\setlength{\tabcolsep}{4pt}
\caption{\textbf{Static evaluation settings.} Shots are the few-shot examples supplied by the evaluation harness.}
\label{tab:static-evaluation-settings}
\begin{tabular}{lrrl}
\toprule
Benchmark & Split ($n$) & Shots & Metric \\
\midrule
ARC-Challenge & test (1,172) & 25 & Normalised accuracy \\
GPQA-Diamond & train (198) & 0 & Normalised accuracy \\
GSM8K & test (1,319) & 5 & Strict exact match \\
HellaSwag & validation (10,042) & 10 & Normalised accuracy \\
PIQA & validation (1,838) & 0 & Normalised accuracy \\
TruthfulQA-MC2 & validation (817) & 0 & Accuracy \\
IFEval & train (541) & 0 & Mean strict/loose accuracy \\
\bottomrule
\end{tabular}
\end{table}

ARC-C, GPQA-Diamond, HellaSwag, PIQA, and TruthfulQA-MC2 are scored by comparing the likelihood of the answer choices; normalised accuracy is used where shown in Table~\ref{tab:static-evaluation-settings}. GSM8K uses greedy generation with a limit of 1,024 new tokens and its standard stop strings. IFEval also uses greedy generation, with a limit of 1,280 new tokens, and averages prompt- and instruction-level accuracy under both strict and loose matching. GPQA and IFEval use the only splits supplied by their source datasets. TruthfulQA's nominal zero-shot setting includes the fixed six-example QA preamble embedded in its standard prompt.

\paragraph{IFEval chat-template convention.}
The default IFEval scores apply the chat template stored with each model. These scores are used in the heatmaps, aggregate trajectories, benchmark correlations, and phase-gain analyses. 

\subsubsection{Chat-based evaluations}
We evaluate every chat benchmark independently with two judges: Qwen3-Next-80B-A3B-Instruct-FP8 and Gemma 4 31B-it. Both judges evaluate the same generated candidate and comparator responses using identical prompts, decoding settings, response-order rules, tie handling, and aggregation procedures; only the judge model changes. This controlled comparison isolates sensitivity to judge choice. Table~\ref{tab:chat-evaluation-protocols} summarizes the comparator and reported score for each benchmark; response-order and aggregation details follow the table. 

\begin{table}
\centering
\small
\setlength{\tabcolsep}{5pt}
\caption{\textbf{Chat-evaluation protocols.} The SFT comparator is the checkpoint used to initialize the corresponding DPO run.}
\label{tab:chat-evaluation-protocols}
\begin{tabular}{lll}
\toprule
Evaluation & Comparator & Reported score \\
\midrule
Arena-Hard v2.0 & Matched SFT checkpoint & Bootstrapped win rate \\
MT-Bench & Matched SFT checkpoint & Pairwise win rate \\
AlpacaEval 2.0 & GPT-4 reference; 805 prompts & Length-controlled win probability \\
Elo & LMArena pool; 500 prompts/opponent & Bradley--Terry rating \\
\bottomrule
\end{tabular}
\end{table}

Arena-Hard judges every comparison in both response orders; ties contribute 0.5, and strong preferences receive threefold weight in the bootstrap. MT-Bench uses a fixed order with the candidate first and the SFT comparator second; ties again contribute 0.5. Its reference-assisted prompts are used for relevant mathematics, reasoning, and coding questions. AlpacaEval randomizes response order deterministically for each instruction and computes a continuous preference from the two possible one-token judgments. Elo evaluates each comparison in both orders and fits a Bradley--Terry model to the resulting pairwise scores; exact ties and unparseable judgments contribute 0.5 to each response.

Table~\ref{tab:chat-decoding-settings} gives the decoding settings for candidate-response generation and both judges. Arena-Hard and AlpacaEval generate candidate responses greedily. MT-Bench follows the FastChat category-specific temperatures. Elo is the only pipeline that samples both candidate and judge outputs: submitted jobs inherit the JudgeArena \texttt{ChatVLLM} defaults of temperature 0.6 and $\texttt{top\_p}=0.95$. Consequently, individual Elo evaluations are stochastic across repeated executions, unlike the other three pipelines under the submitted settings.

\begin{table}
\centering
\small
\setlength{\tabcolsep}{7pt}
\caption{\textbf{Decoding settings for chat evaluation.} Limits are candidate/judge output tokens.}
\label{tab:chat-decoding-settings}
\begin{tabular}{lccc}
\toprule
Evaluation & Candidate temperature & Judge temperature & Token limit \\
\midrule
Arena-Hard v2.0 & 0 & 0 & 4,096 / 4,096 \\
MT-Bench & Category-specific & 0 & 4,096 / 2,048 \\
AlpacaEval 2.0 & 0 & 1 (one token) & 2,048 / 1 \\
Elo & 0.6 ($p=0.95$) & 0.6 ($p=0.95$) & 32,768 / 32,768 \\
\bottomrule
\end{tabular}
\end{table}

For MT-Bench, candidate-generation temperatures are kept at their default values: 0.7 for writing and role-playing, 0.1 for STEM and humanities, and 0 for extraction, mathematics, coding, reasoning, and Arena-Hard-200 categories. We use the reference prompt templates supplied by Arena-Hard v2.0, MT-Bench, and AlpacaEval 2.0.

\subsubsection{Held-out DPO loss}
\label{app:validation-loss-evaluation}
For the primary Dolci-Instruct grids, we construct the fixed 1,300-pair validation set described in Appendix~\ref{app:dataset-details}. We do not evaluate on this set during DPO training. After training, we evaluate the latest saved checkpoint, or the final model when no intermediate checkpoint is available, once on the held-out pairs. The resulting loss is the validation loss value used in Section~\ref{sec:rq1}. The validation loss directed Bayesian-optimisation experiments use the same held-out set but evaluate loss every 100 training steps and save checkpoints every 400 steps.

\subsection{Hyperparameter-optimisation Simulation Settings}
\label{app:hpo-settings}

\paragraph{Compared optimisers.}
We compare three single-fidelity methods, which train each proposed configuration to completion, with two methods that use intermediate training results to allocate further compute:
\begin{itemize}
    \item \textbf{Random search} samples learning-rate and $\beta$ configurations independently of previous outcomes~\citep{bergstra2012random}.
    
    \item \textbf{TPE} models regions associated with good and poor completed trials using separate density estimators, then proposes configurations favoured by their density ratio~\citep{bergstra2011algorithms}.
    
    \item \textbf{CQR} fits conformalized quantile-regression models to predict performance and uncertainty from completed configurations, then uses these estimates to propose subsequent configurations~\citep{salinas2023cqr}.
    
    \item \textbf{ASHA} uses asynchronous successive halving: candidates with strong partial-training results receive more training, while weaker candidates are stopped at intermediate resource levels~\citep{li2020massively}.
    
    \item \textbf{BOHB} combines model-based configuration proposals with Hyperband-style allocation of training resources across candidates~\citep{falkner2018bohb}.
\end{itemize}

\paragraph{Simulation protocol and compute budgets.}
Each simulation searches a continuous domain in $(\log_{10}\mathrm{LR},\log_{10}\beta)$ bounded by the minimum and maximum values measured at that model size. Training progress is the fidelity variable, with checkpoints at steps 200, 400, $\ldots$, 2,000 and the final step 2,031; the surrogate's cumulative training time is the trial cost. Two trials run in parallel, and each optimiser--surrogate--size setting is repeated with 25 random seeds. We define one budget unit as the measured wall-clock duration of one complete reference DPO run. The larger budget is eight units, which allows CQR's five initial random trials to finish and leaves time for adaptive proposals. The restrictive budget is one unit, corresponding to approximately two full-run equivalents of total compute across both workers.

We compare random search, the tree-structured Parzen estimator (TPE), conformalized quantile regression (CQR), asynchronous successive halving (ASHA), and Bayesian optimisation with Hyperband (BOHB). For TPE, we set \texttt{top\_n\_percent}=33 and \texttt{num\_min\_data\_points}=3; all other TPE settings use the implementation defaults. Random search, TPE, and CQR evaluate configurations to completion, whereas ASHA and BOHB may terminate candidates at intermediate checkpoints. All methods receive the same time limit, workers, and continuous domain. \zdynamic{} is the primary maximization objective; \zall{} is retained as a cross-objective evaluation.

\paragraph{Successive-halving schedules.}
ASHA and BOHB use a single successive-halving bracket with a grace period of 400 steps and reduction factor two, yielding comparisons at steps 400, 800, and 1,600 before final fidelity. Approximately half of the active configurations are retained at each comparison. Delaying the first decision until step 400 avoids pruning at step 200, where the 8B and 14B \zdynamic{} rankings are poorly aligned with their final orderings.

\paragraph{Continuous surrogate environments.}
The logged grids answer only at observed configurations, so we construct continuous surrogate versions of every per-size benchmark. Learning rate and $\beta$ are transformed to base-10 logarithms, and no query extrapolates outside the measured range. Training step remains discrete, and one fitted model predicts the complete trajectory at the recorded fidelities.

We fit four alternatives separately and never pool their predictions: one-nearest-neighbour regression, which is piecewise constant between measured configurations; distance-weighted three- and five-nearest-neighbour regression; and a Gaussian process with a constant component, Mat\'ern-$5/2$ kernel, and white-noise term initialized to $10^{-2}$. GP targets are normalised, kernel optimisation uses three restarts, and the fitting seed is zero. These environments test sensitivity to different interpolation assumptions and do not constitute additional observed DPO runs.

\paragraph{Recommendation and scoring.}
At a budget cutoff, we recommend the trial with the highest search score at final step 2,031 whenever at least one trial has completed. If no trial has completed, we compare only trials evaluated at the highest fidelity shared within that run. We then evaluate the recommended $(\mathrm{LR},\beta)$ at true final fidelity on the same surrogate. This prevents an optimiser from benefiting merely because it reports more intermediate checkpoints. For pooled comparisons, we first average the five model-size outcomes within each seed and then form a 95\% $t$-interval across the 25 seed-level averages. Differences from random search are paired by seed.

\section{Additional Experimental Results}
\subsection{Qwen3 Experiments}
\label{app:full-qwen3-results}

\subsubsection{Benchmark-Level Agreement And Validation Loss}
\label{app:qwen3-benchmark-loss-correlations}

\begin{figure}[h]
\centering
\includegraphics[width=\textwidth]{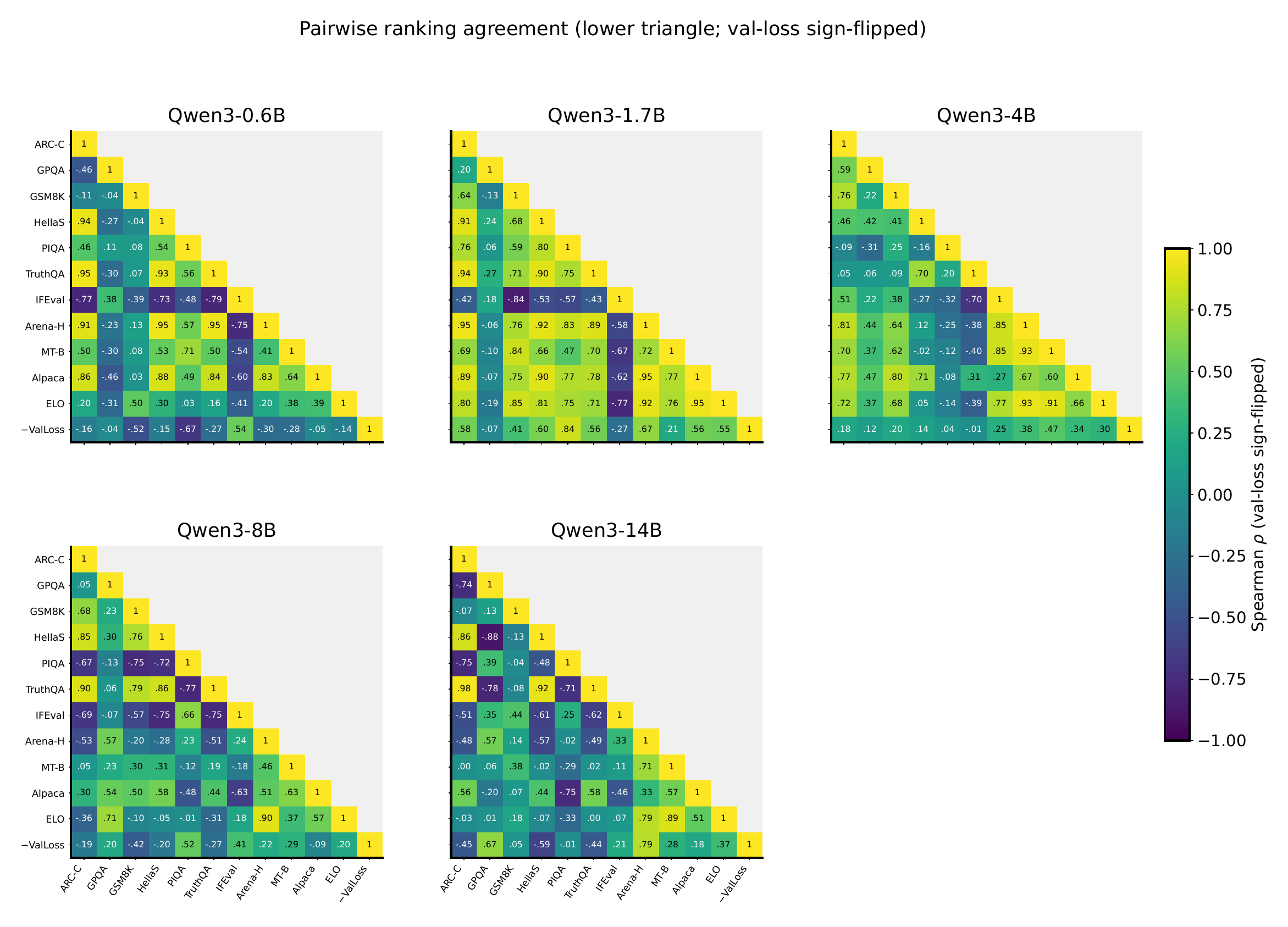}
\caption{\textbf{Benchmark-level configuration-ranking correlations and negative validation loss across Qwen3 sizes.}
Each cell gives the Spearman correlation between two final configuration rankings within one model-size grid. The final row and column use negative validation loss, so positive values indicate that lower validation loss is associated with stronger downstream performance. Benchmark agreement and alignment with validation loss both vary substantially by metric and model size.}
\label{fig:qwen3-benchmark-loss-correlations}
\end{figure}

Figure~\ref{fig:qwen3-benchmark-loss-correlations} gives the complete benchmark-level rank-correlation matrices underlying the aggregate analyses in Section~\ref{sec:rq1}. Each matrix shows that the relationship between negative validation loss and downstream rankings varies across both benchmarks and model sizes. It can align positively with some evaluations while being weakly or negatively associated with others, explaining why validation loss does not provide a reliable aggregate selection signal.

\subsubsection{Validation signals at fixed \texorpdfstring{$\beta$}{beta}}
\label{app:within-beta-loss}

DPO loss changes with $\beta$, so its values need not be directly comparable across the full grid. As a sensitivity check, we rank learning rates by lower final validation loss and higher final \zall{} separately within each $\beta$, then pool those within-$\beta$ ranks to compute Spearman correlation. From Qwen3-0.6B through 14B, the resulting correlations are $-0.67$, $0.33$, $0.05$, $0.30$, and $0.67$. The corresponding full-grid correlations are $-0.48$, $0.37$, $0.15$, $0.12$, and $0.05$. Fixed-$\beta$ comparisons therefore do not produce consistently strong agreement with downstream performance, although the relationship varies substantially by size.

We also compare TRL training reward accuracy with \zall{}. Unlike the magnitude of the DPO loss, the chosen-versus-rejected decision is unchanged by multiplying a fixed policy's reward margin by positive $\beta$. Its within-$\beta$ correlations are $0.95$, $1.00$, $-0.23$, $-0.40$, and $0.11$ across the same sizes. Reward accuracy is informative at the two smallest Qwen3 sizes but does not consistently rank configurations at larger sizes. Each $\beta$ slice contains only a few learning rates, so these correlations describe the tested configurations rather than precise population relationships.

\begin{figure}
\centering
\includegraphics[width=0.75\linewidth]{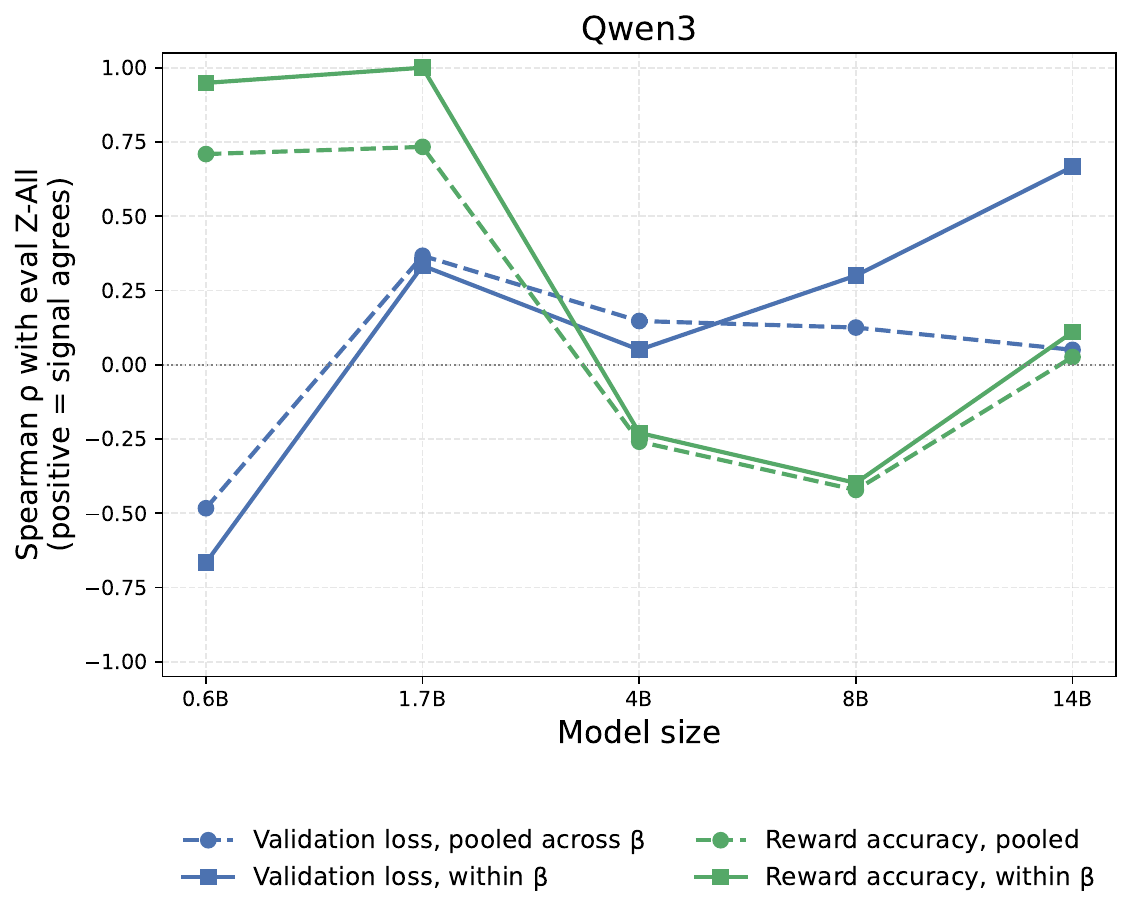}
\caption{\textbf{Agreement of final DPO signals with \zall{} across Qwen3 sizes.} Dashed lines rank configurations across the full grid; solid lines rank learning rates separately within each $\beta$ before pooling the ranks. Blue denotes negative validation loss, so positive correlation means lower loss agrees with higher \zall{}; green denotes reward accuracy.}
\label{fig:within-beta-loss}
\end{figure}

\subsubsection{Objective-Specific Learning-Rate Trends}
\label{app:objective-specific-scaling}

\begin{figure}[h]
\centering
\includegraphics[width=0.7\linewidth]{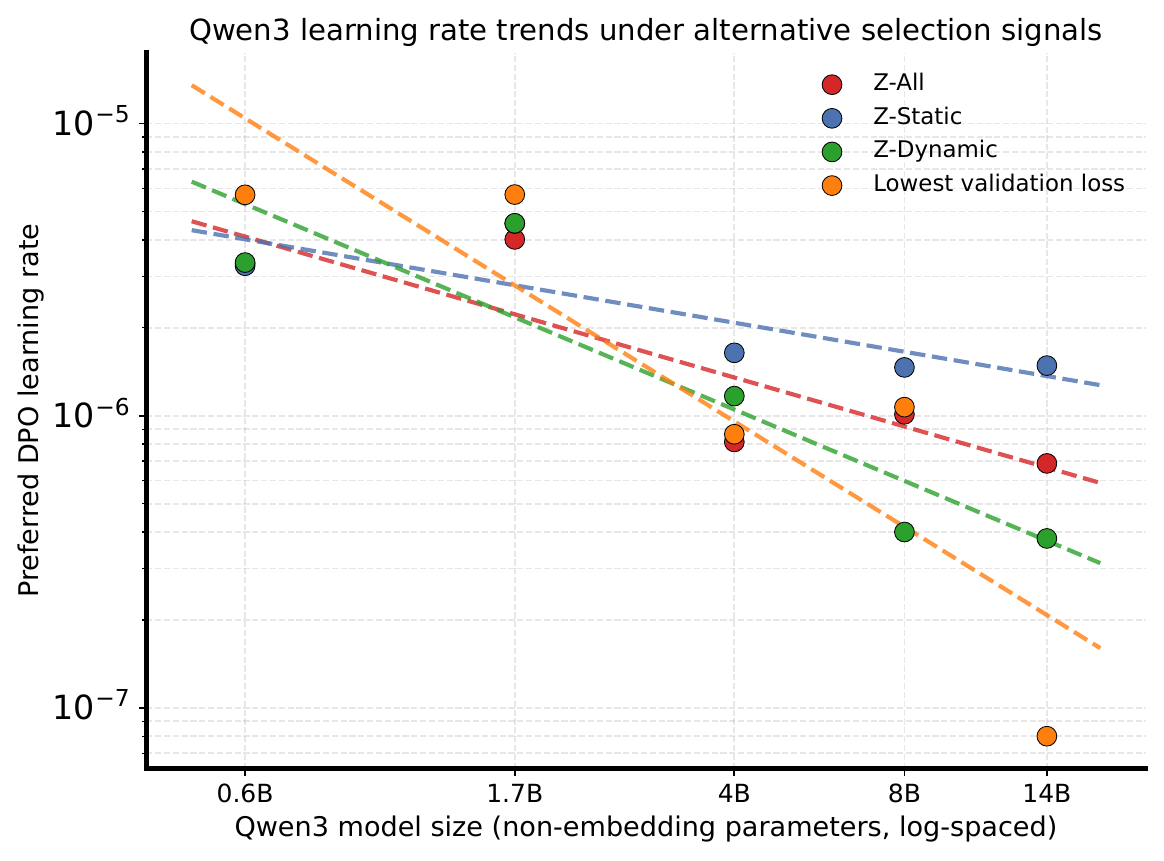}
\caption{\textbf{Preferred Qwen3 DPO learning rates differ by selection signal.}
Each point is the learning-rate component of a two-dimensional quadratic response-surface optimum over $(\log_{10}\mathrm{LR},\log_{10}\beta)$ at one Qwen3 size. The three downstream aggregates are maximized, whereas validation loss is minimized. The estimates describe preferences within the measured grids and are not extrapolated scaling laws.}
\label{fig:zscaling_val_trend}
\end{figure}

The learning-rate trend in Section~\ref{sec:rq2} uses \zdynamic{}, but the descriptive relationship depends on the evaluation objective. Figure~\ref{fig:zscaling_val_trend} compares the corresponding learning-rate estimates for \zall{}, \zstatic{}, \zdynamic{}, and validation loss. At each Qwen3 size, we fit a two-dimensional quadratic response surface over $(\log_{10}\mathrm{LR},\log_{10}\beta)$ and extract its preferred learning rate.

The estimates differ across selection signals, particularly at larger sizes: \zstatic{} favours higher learning rates than \zdynamic{}, while validation loss follows a different pattern again. These response-surface estimates summarize preferences within the measured grids; they are neither unconstrained optima nor predictions for an unseen model size.

\begin{figure*}[t]
\centering
\includegraphics[width=\linewidth]{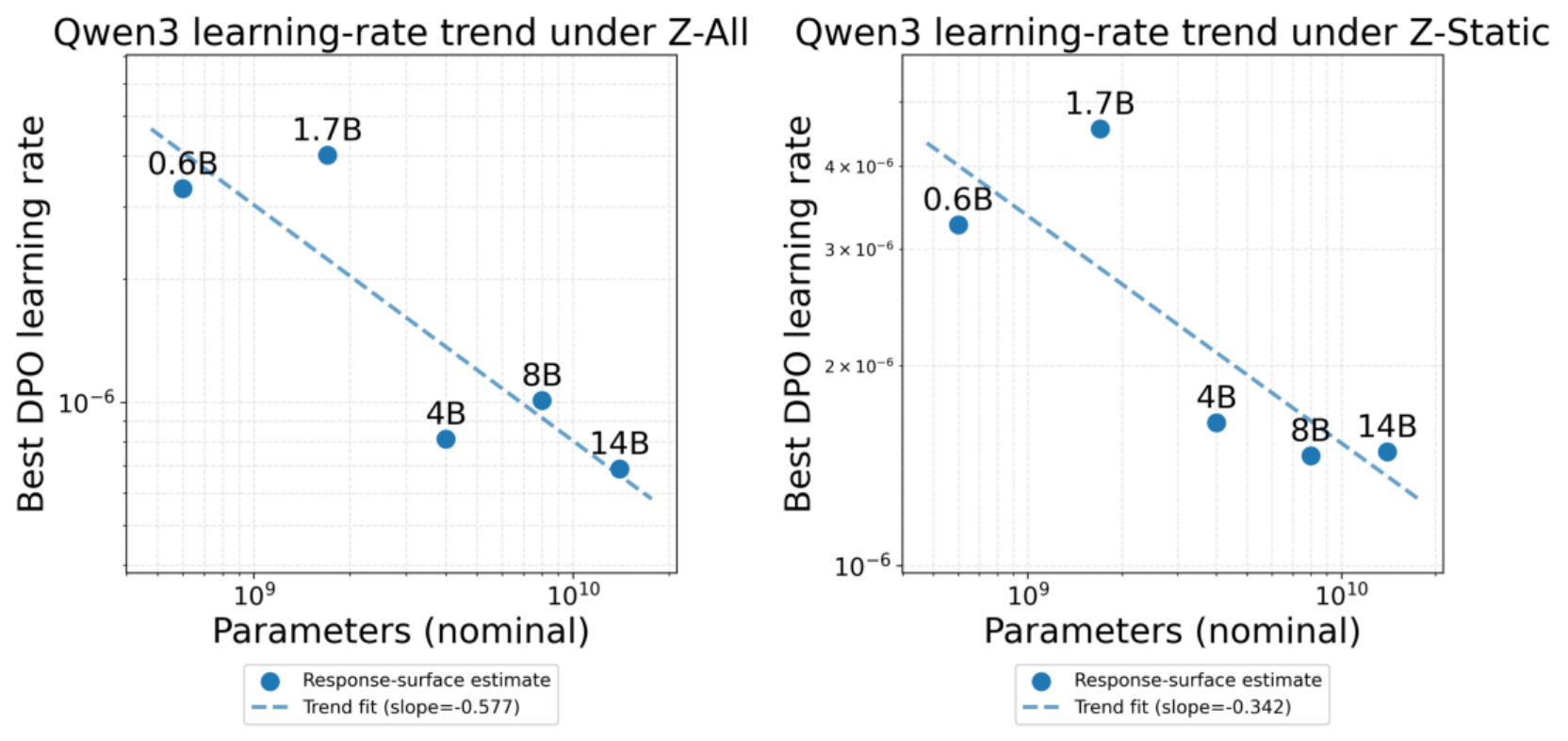}
\caption{\textbf{Objective-specific Qwen3 learning-rate trends.}
\textbf{Left:} \zall{}. \textbf{Right:} \zstatic{}. Each point is the learning-rate estimate from a two-dimensional response surface at one Qwen3 size; dashed lines are descriptive log-linear fits across the five estimates. These within-family fits are not held-out predictions.}
\label{fig:objective-specific-lr-scaling}
\end{figure*}

\subsubsection{Final aggregate optima}
Table~\ref{tab:qwen3-objective-optima} reports the best observed final configuration under each evaluation aggregate. The \zall{} selections reproduce the non-monotonic pattern discussed in Section~\ref{sec:rq2}: the learning rate first rises from 0.6B to 1.7B and then decreases through 14B. The objective-specific columns show that this pattern should not be separated from the evaluation criterion. At 8B, for example, \zdynamic{} selects a learning rate of $4\times10^{-7}$ whereas \zstatic{} selects $4\times10^{-6}$. The selected $\beta$ also differs across objectives at 0.6B, 1.7B, and 14B.

\begin{table}[h]
\centering
\small
\caption{\textbf{Best observed final Qwen3 configuration under each aggregate.} Each entry is $(\mathrm{learning\ rate},\beta)$ and is selected from the full grid available at that model size. $\downarrow$ and $\uparrow$ mark the lowest and highest tested learning rates, respectively. Wherever selected, $\beta=0.01$ is the lower tested bound; the 0.6B \zstatic{} winner uses the upper bound $\beta=0.04$. Boundary selections do not locate unconstrained optima.}
\label{tab:qwen3-objective-optima}
\begin{tabular}{lcccc}
\toprule
Model & \zall{} & \zmacro & \zdynamic{} & \zstatic{} \\
\midrule
Qwen3-0.6B & $(4\times10^{-6},0.01)$ & $(4\times10^{-6},0.01)$ & $(8\times10^{-6},0.01)^{\uparrow}$ & $(4\times10^{-6},0.04)$ \\
Qwen3-1.7B & $(8\times10^{-6},0.01)^{\uparrow}$ & $(8\times10^{-6},0.01)^{\uparrow}$ & $(4\times10^{-6},0.02)$ & $(8\times10^{-6},0.01)^{\uparrow}$ \\
Qwen3-4B   & $(2\times10^{-6},0.01)$ & $(1\times10^{-6},0.01)$ & $(4\times10^{-7},0.01)^{\downarrow}$ & $(2\times10^{-6},0.01)$ \\
Qwen3-8B   & $(1\times10^{-6},0.01)$ & $(4\times10^{-7},0.01)^{\downarrow}$ & $(4\times10^{-7},0.01)^{\downarrow}$ & $(4\times10^{-6},0.01)^{\uparrow}$ \\
Qwen3-14B  & $(4\times10^{-7},0.01)$ & $(4\times10^{-7},0.01)$ & $(4\times10^{-7},0.01)$ & $(4\times10^{-6},0.02)^{\uparrow}$ \\
\bottomrule
\end{tabular}
\end{table}

\subsubsection{Dynamic--static trade-offs across Qwen3 sizes}
\label{app:dynamic-static-tradeoffs}

\begin{figure}[h]
    \centering
    \includegraphics[width=0.90\textwidth]{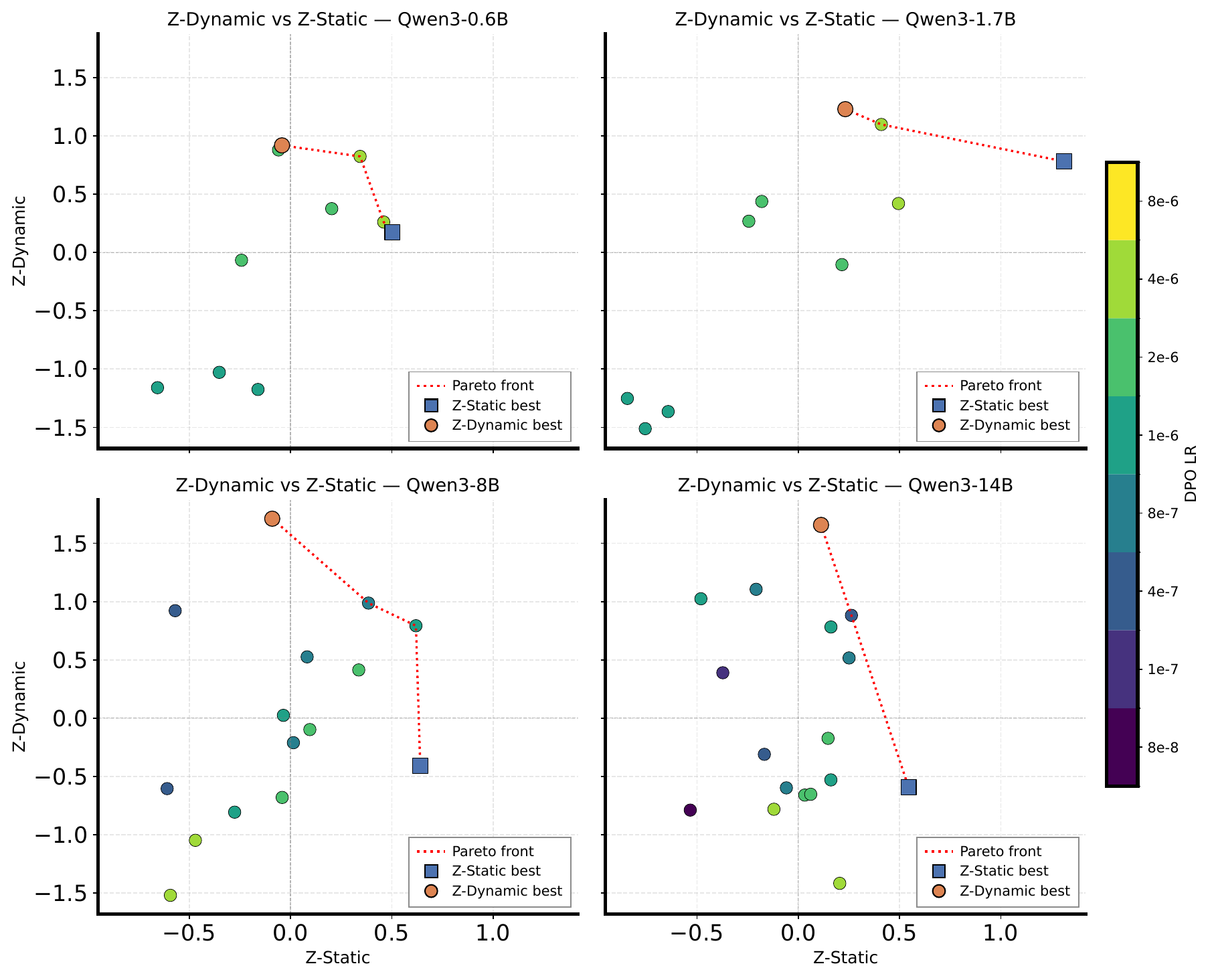}
    \caption{\textbf{Dynamic--static trade-offs at the remaining Qwen3 sizes.}
    Each point is one final DPO configuration, with \zdynamic{} and \zstatic{} standardized within its model-size grid. Red segments mark the per-size Pareto frontier; squares and circles identify the configurations maximizing \zstatic{} and \zdynamic{}, respectively. Point colour denotes DPO learning rate on a common logarithmic scale.}
    \label{fig:dynamic-static-other-sizes}
\end{figure}

Figure~\ref{fig:dynamic-static-other-sizes} extends the Qwen3-4B comparison in Figure~\ref{fig:dynamic-static-tradeoff} to the remaining model sizes. At each size, the configurations maximizing \zdynamic{} and \zstatic{} occupy different locations on, or near, the Pareto frontier. Thus, the objective-specific learning-rate choices in Table~\ref{tab:qwen3-objective-optima} reflect a dynamic--static trade-off across the Qwen3 family rather than only the 4B grid.

\subsubsection{\texorpdfstring{\zall{}}{Z-All} response surfaces}
\label{app:qwen3-zall-surfaces}

\begin{figure}
\centering

\includegraphics[width=\linewidth]{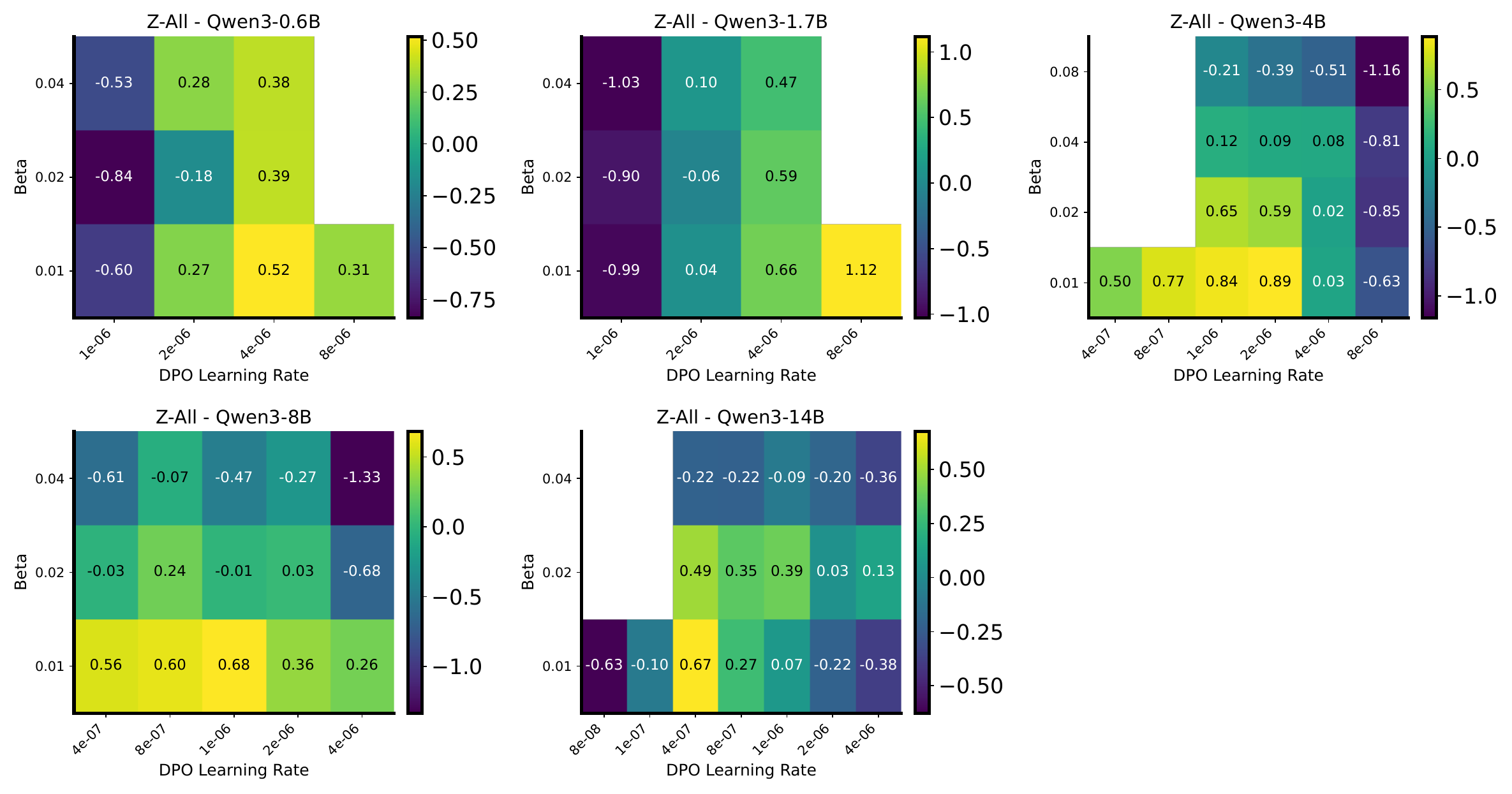}
\caption{\textbf{Final \zall{} response surfaces across Qwen3 sizes.}
Each cell gives the final all-benchmark aggregate score of one evaluated DPO configuration; lighter cells indicate stronger performance. The evaluated grids differ by size, so blank cells denote unobserved configurations and colour scales are normalised separately within each panel. The surfaces show a shift toward lower preferred learning rates at larger scales, but also substantial size-specific variation in the relative ordering of configurations.}
\label{fig:qwen3-zall-surfaces}
\end{figure}

Figure~\ref{fig:qwen3-zall-surfaces} shows the final \zall{} response surface over the configurations evaluated at each Qwen3 size. The preferred region shifts toward lower learning rates at larger sizes, while the irregular and size-specific surface shapes reinforce that a configuration selected at one scale need not rank well at another. Blank cells denote configurations that were not evaluated at that size.

\subsubsection{Equal-group aggregate robustness}
\label{app:equal-group-robustness}

\zall{} weights each benchmark equally, which assigns seven of eleven total weights to the static group. We therefore define
\[
Z_{\mathrm{Macro}}=\tfrac{1}{2}\left(Z_{\mathrm{Dynamic}}+Z_{\mathrm{Static}}\right),
\]
giving chat and static benchmarks equal total weight. The \zall{} and $Z_{\mathrm{Macro}}$ configuration rankings remain highly correlated at every Qwen3 size ($\rho=0.903$--$0.994$). Their selected configurations match at 0.6B, 1.7B, and 14B, while the preferred learning rate differs at 4B and 8B. Thus, equal group weighting can alter the precise selected configuration, but it preserves the main conclusions about validation loss and cross-scale ranking instability.

\begin{figure}[h]
    \centering
    \includegraphics[width=\textwidth]{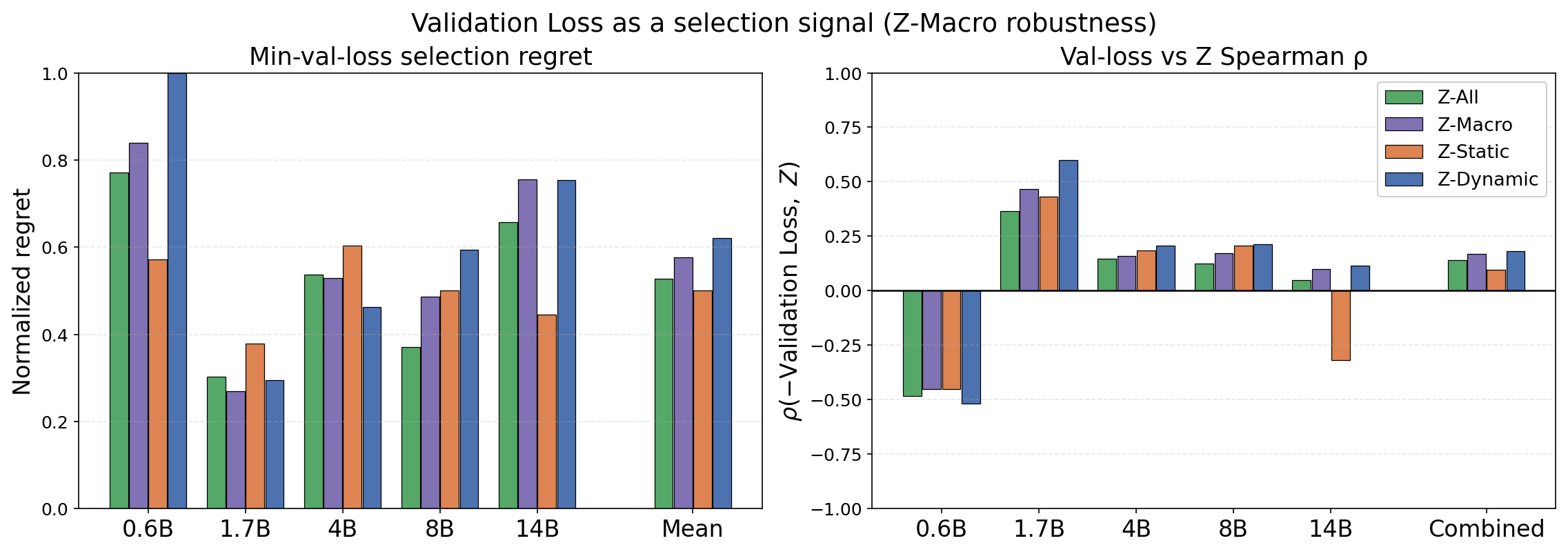}
    \caption{\textbf{Validation loss remains an unreliable selection signal under equal group weighting.}
    The $Z_{\mathrm{Macro}}$ aggregate gives chat and static benchmarks equal total weight. Selecting the minimum-validation-loss configuration remains far from the best $Z_{\mathrm{Macro}}$ configuration at every Qwen3 size, and validation-loss rankings agree only weakly with $Z_{\mathrm{Macro}}$.}
    \label{fig:zmacro-validation-loss}
\end{figure}

\begin{figure}
    \centering
    \includegraphics[width=\textwidth]{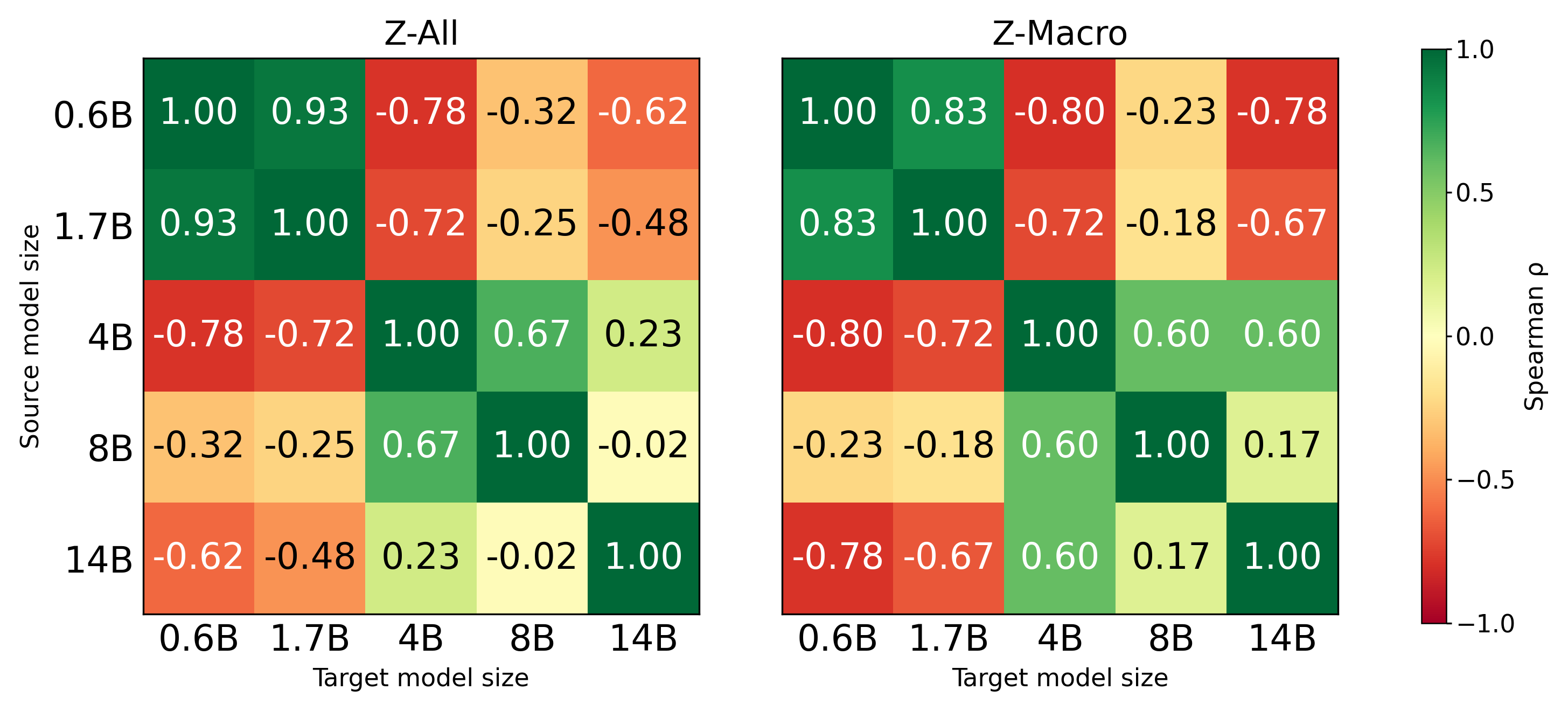}
    \caption{\textbf{Cross-scale ranking instability persists under equal group weighting.}
    Each cell gives Spearman's $\rho$ between rankings of the nine configurations in the shared $3\times3$ grid. \zall{} weights all eleven benchmarks equally, whereas $Z_{\mathrm{Macro}}$ assigns equal total weight to chat and static benchmarks. Neither aggregate shows a consistent relationship between model-size distance and ranking agreement.}
    \label{fig:zall-vs-zmacro-transfer}
\end{figure}

Minimum-validation-loss selection remains unreliable under $Z_{\mathrm{Macro}}$: its mean normalised regret is $0.57$, and the combined Spearman correlation between negative validation loss and $Z_{\mathrm{Macro}}$ is only $0.16$. Cross-scale agreement also remains non-monotonic in model-size distance (Figure~\ref{fig:zall-vs-zmacro-transfer}).

\subsubsection{Stability of the selected configuration}
\label{app:argmax-stability}

\begin{figure}[t]
\centering
\includegraphics[width=0.82\linewidth]{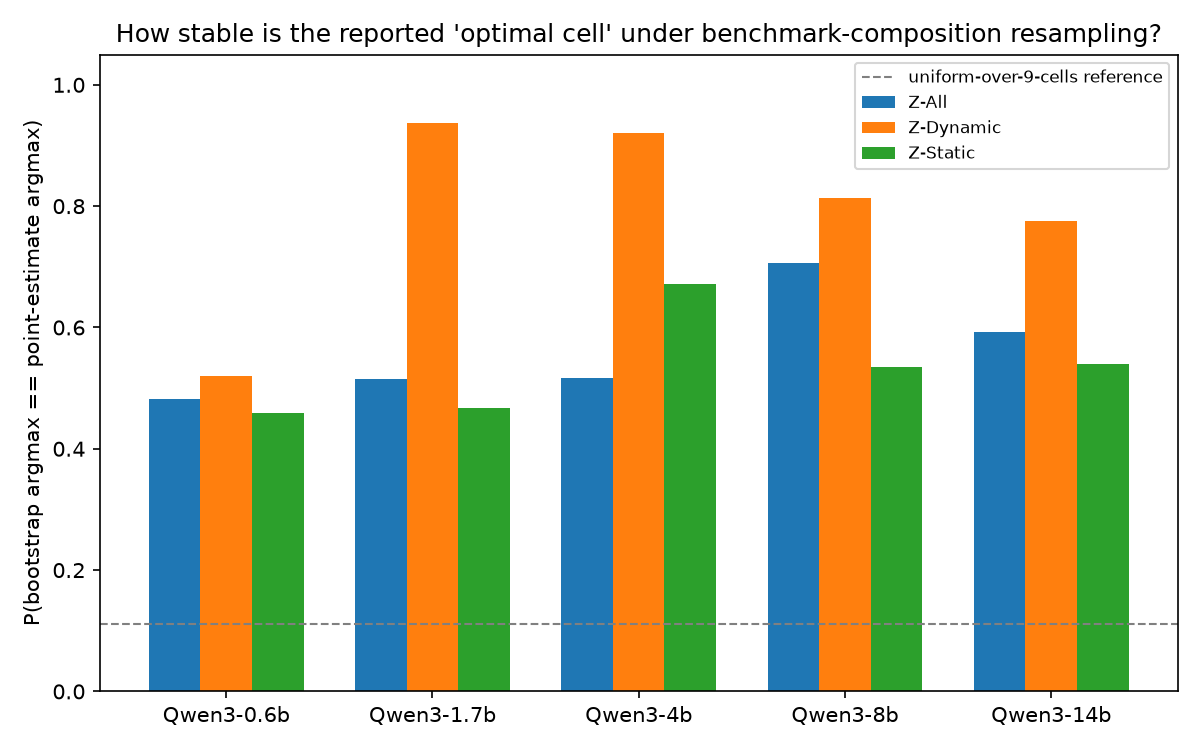}
\caption{\textbf{Winner recovery under benchmark bootstrap resampling.}
Bars show how often resampling the benchmarks within an aggregate selects the same configuration as the original aggregate on the shared nine-configuration core. The dotted line gives the $1/9$ uniform reference. \zdynamic{} has the most stable winner at every Qwen3 size.}
\label{fig:argmax-stability}
\end{figure}

We ask whether the selected configuration depends heavily on the particular benchmarks included in an aggregate. For each aggregate, we repeatedly resample its constituent benchmarks with replacement, recompute configuration scores on the shared $3\times3$ core, and record whether the winner from the original aggregate remains selected. Figure~\ref{fig:argmax-stability} reports this recovery probability. All three aggregates recover their original winner more often than the $1/9$ uniform reference, but stability differs by objective. \zdynamic{} recovers its winner in $52.0$--$93.7\%$ of resamples and is more stable than \zstatic{} at every Qwen3 size. \zall{} recovers its winner in $48.3$--$70.6\%$ of resamples. This analysis measures sensitivity to benchmark composition only and not the uncertainty within an individual benchmark.

\subsubsection{Benchmark-pair agreement}
\begin{figure}
\centering
\includegraphics[width=\textwidth]{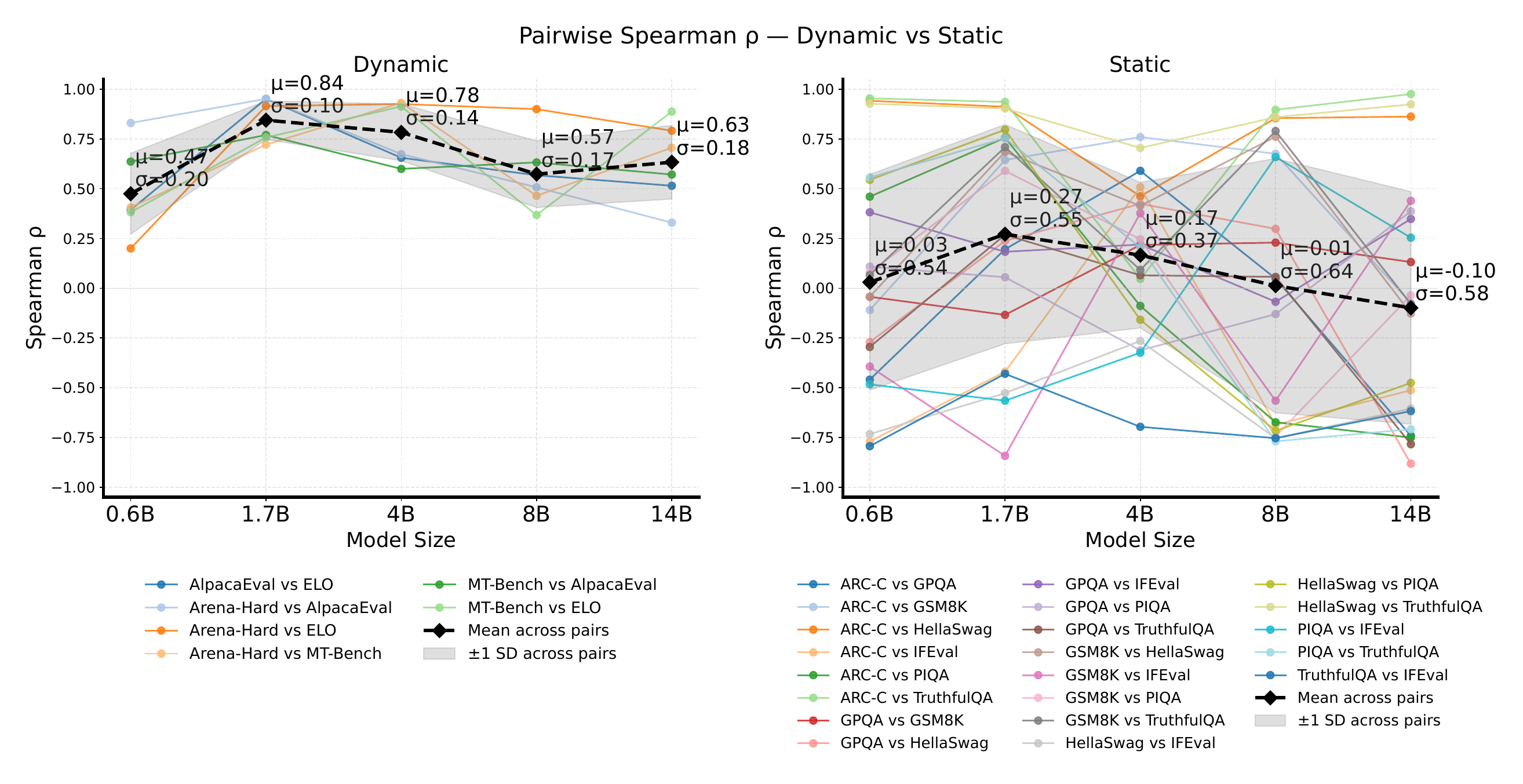}
\caption{\textbf{Pairwise agreement between Qwen3 configuration rankings.} Each coloured line gives the Spearman correlation between one pair of benchmarks at each model size. Black diamonds denote the mean across the six chat pairs (left) or 21 static pairs (right), and the shaded regions show one population standard deviation across pairs. ``Dynamic'' in the figure denotes the four chat-based evaluations.}
\label{fig:qwen3-benchmark-pair-agreement}
\end{figure}

Figure~\ref{fig:qwen3-benchmark-pair-agreement} expands the group summaries in Table~\ref{tab:qwen-benchmark-agreement} into their constituent benchmark pairs. Every chat-benchmark pair is positively correlated at every model size, although the strength of agreement varies. Static-benchmark pairs span strong positive and negative correlations at the same size. Their low mean agreement therefore reflects conflicting configuration rankings across capabilities rather than uniformly uninformative benchmarks.

\subsubsection{Ranking stability during training}
\begin{figure}[t]
\centering
\begin{minipage}{0.49\textwidth}
    \centering
    \includegraphics[width=\linewidth]{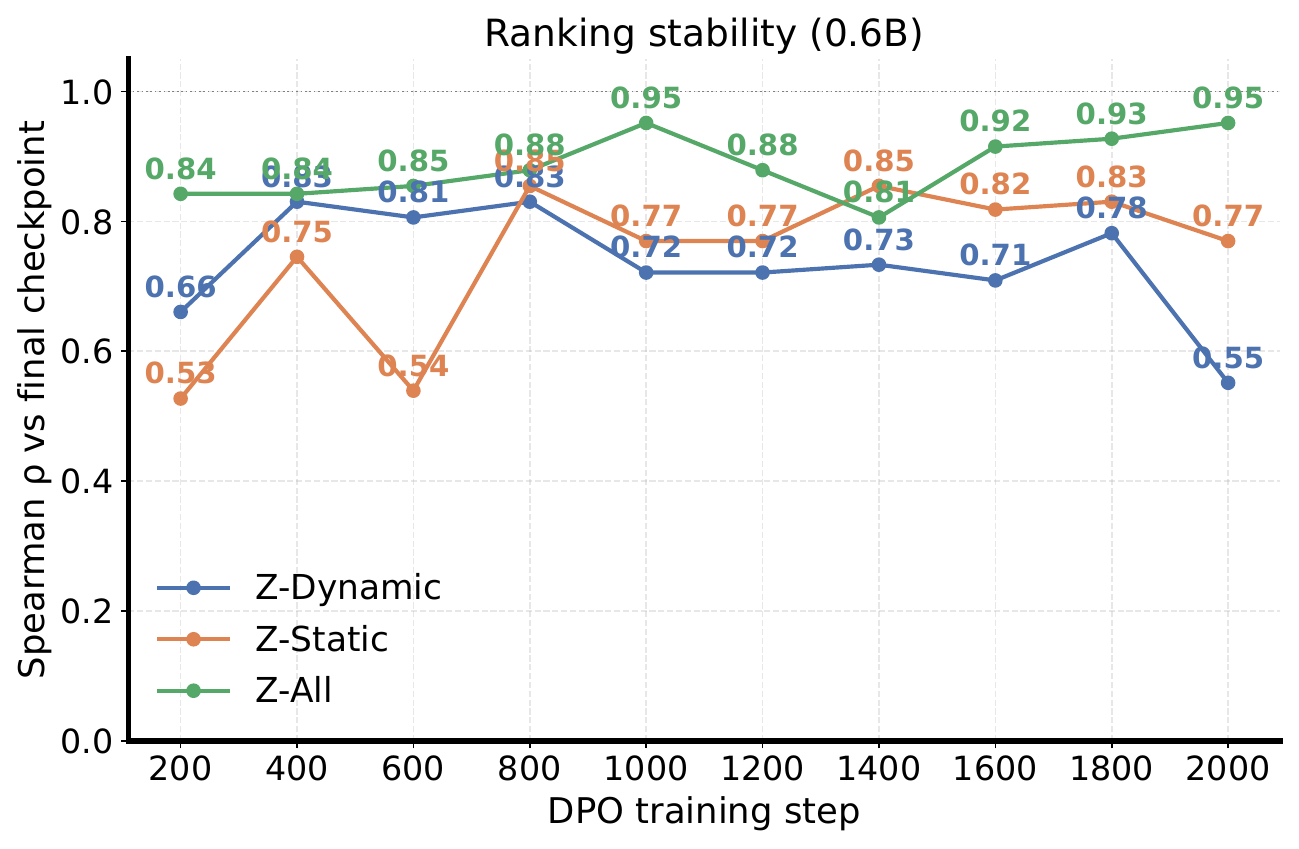}
\end{minipage}
\hfill
\begin{minipage}{0.49\textwidth}
    \centering
    \includegraphics[width=\linewidth]{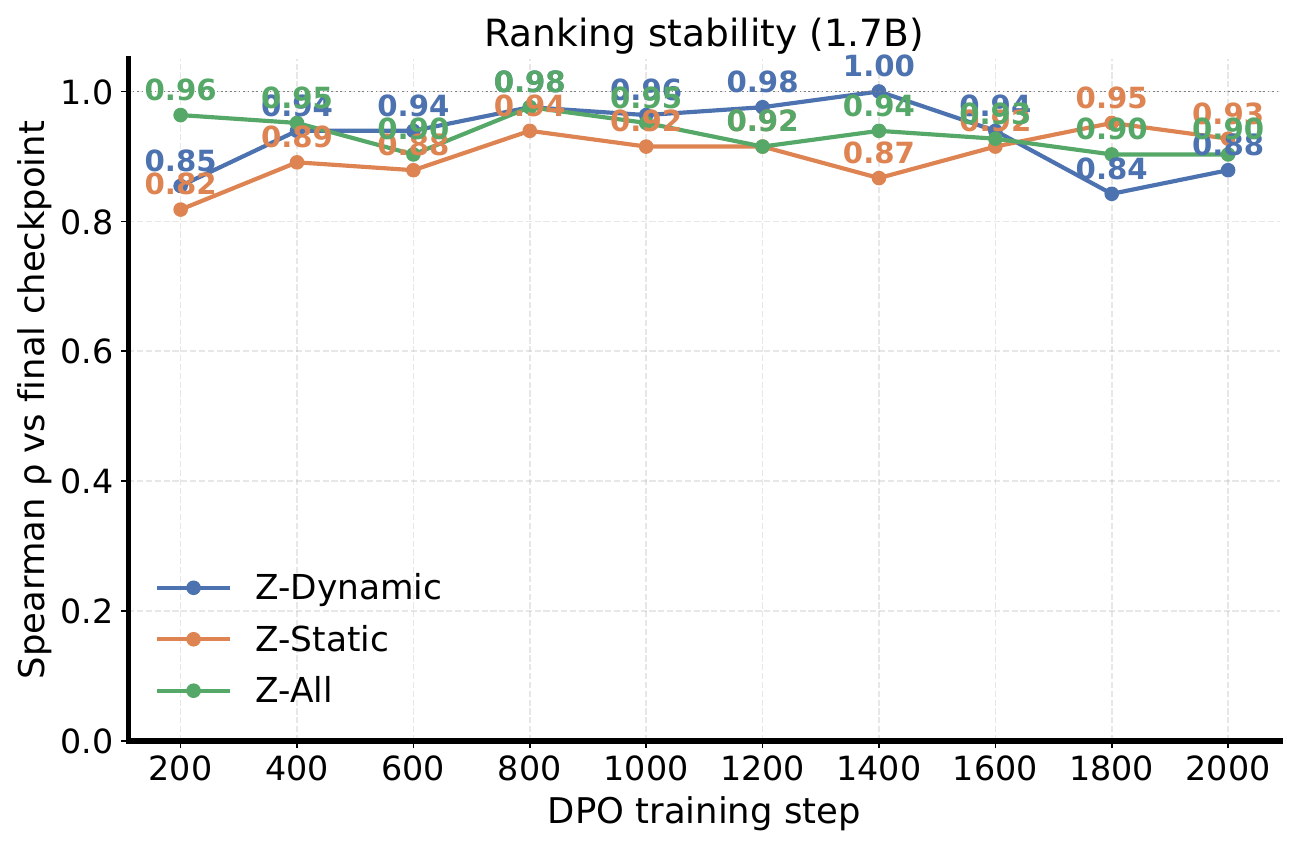}
\end{minipage}

\begin{minipage}{0.49\textwidth}
    \centering
    \includegraphics[width=\linewidth]{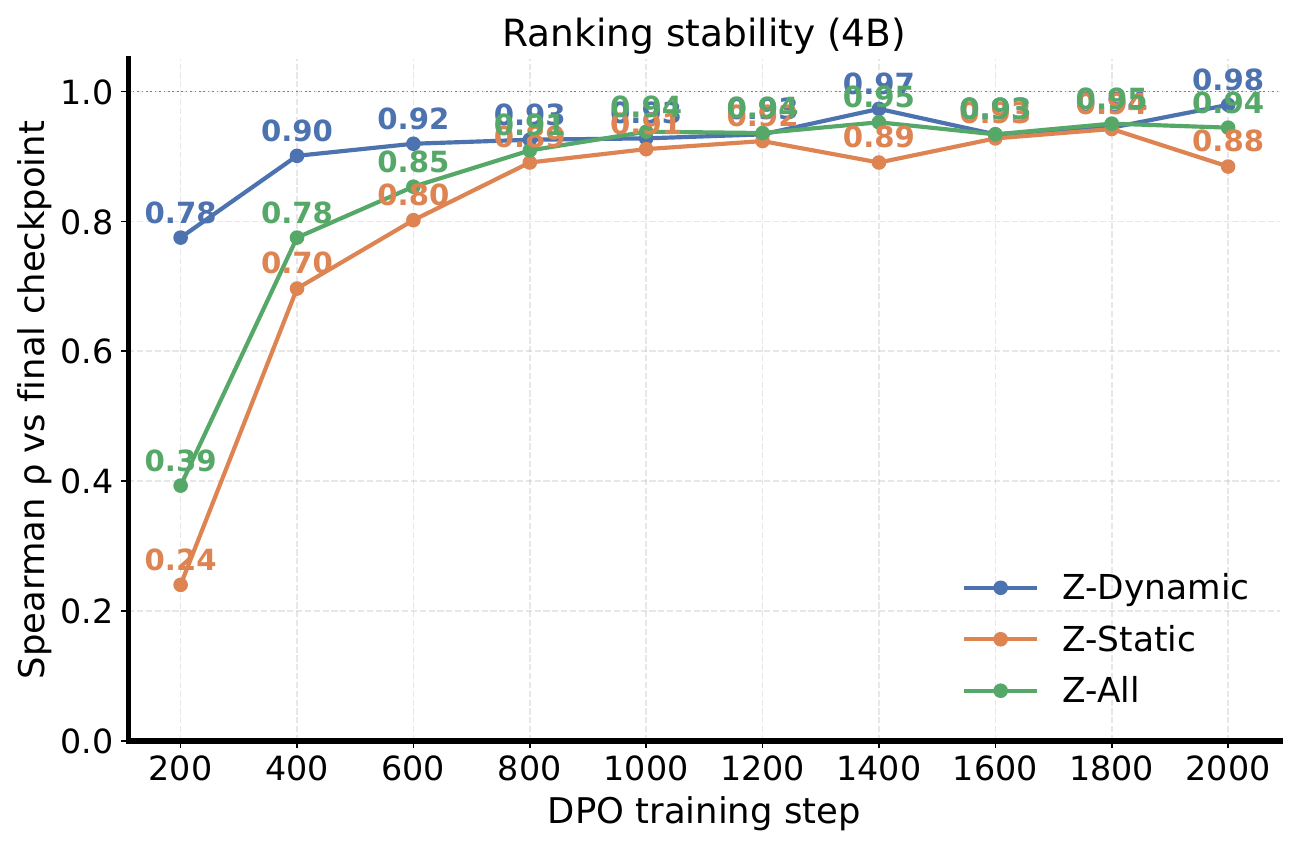}
\end{minipage}
\hfill
\begin{minipage}{0.49\textwidth}
    \centering
    \includegraphics[width=\linewidth]{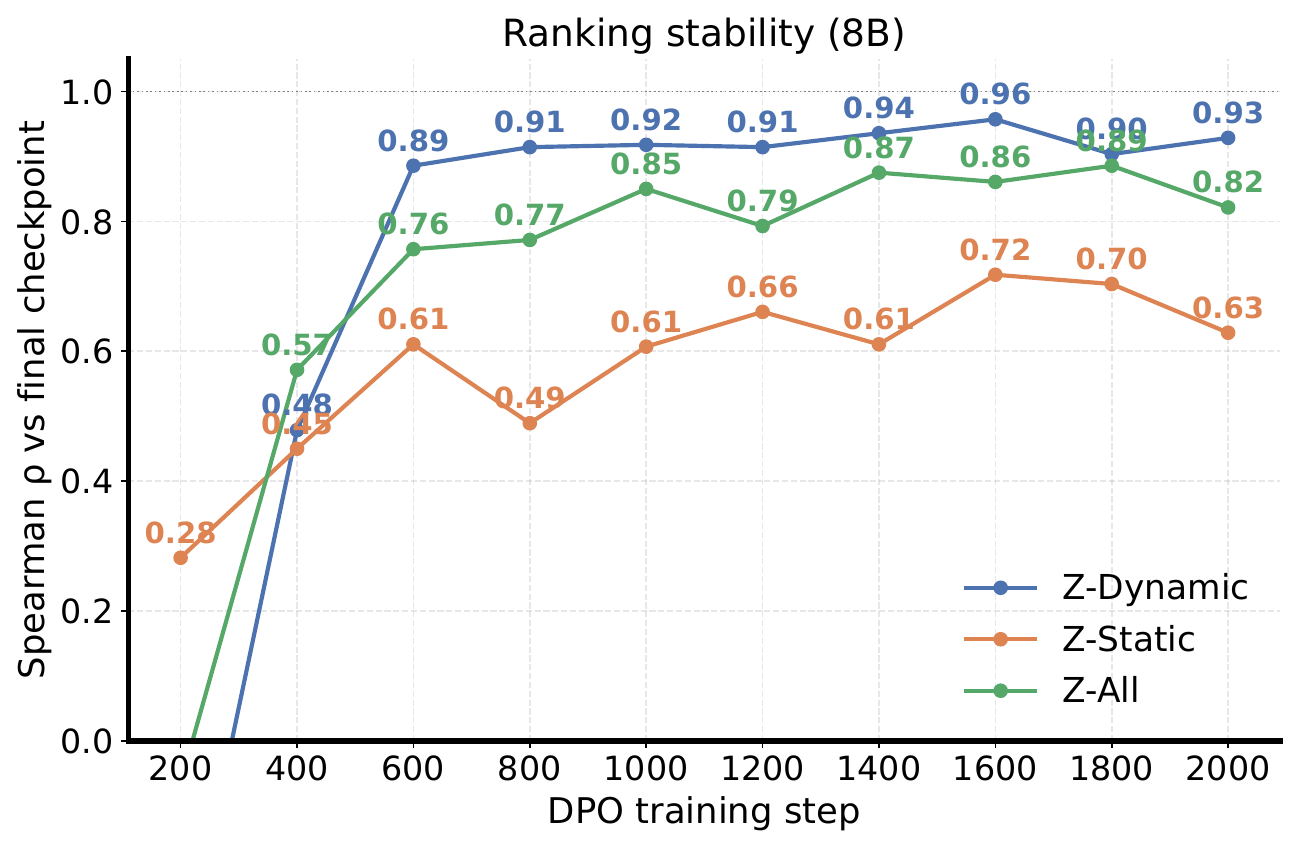}
\end{minipage}

\begin{minipage}{0.49\textwidth}
    \centering
    \includegraphics[width=\linewidth]{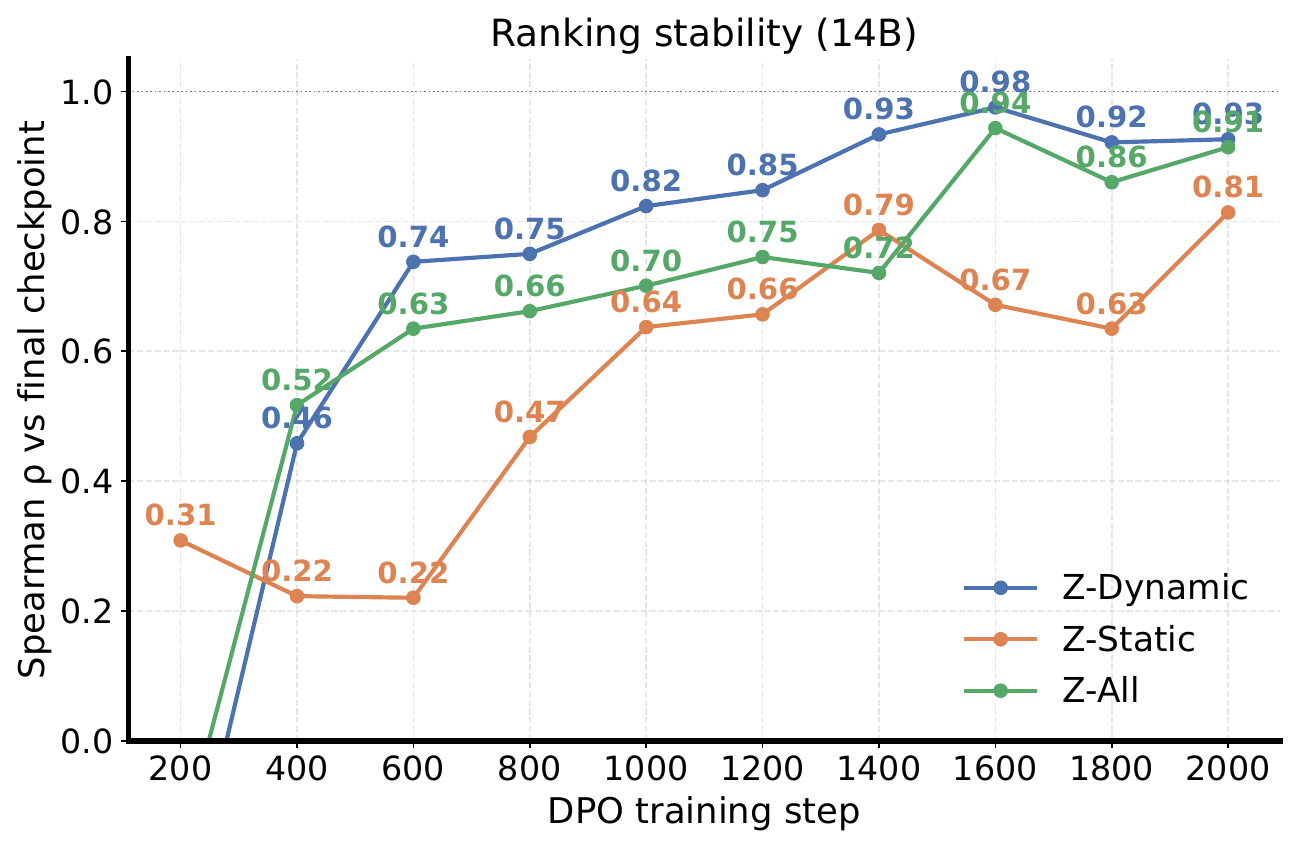}
\end{minipage}
\caption{\textbf{Intermediate Qwen3 rankings increasingly resemble the final ranking, but convergence is scale- and objective-dependent.} Each panel compares the ranking of all available $(\mathrm{learning\ rate},\beta)$ configurations at an intermediate checkpoint with their ranking after the full DPO run. Lines report Spearman correlation for \zall{}, \zdynamic{}, and \zstatic{}.}
\label{fig:qwen3-ranking-stability}
\end{figure}

Figure~\ref{fig:qwen3-ranking-stability} reports the complete ranking-stability trajectories underlying the step-600 values in Section~\ref{sec:rq3}. For this longitudinal comparison, each benchmark is standardized using the mean and population standard deviation of its final scores across the corresponding model-size grid; the same transformation is then applied to every intermediate checkpoint. This keeps the benchmark scale fixed over training. At each step, we compute the Spearman correlation between the intermediate and final configuration rankings using configurations observed at both points.

The trajectories confirm that useful ranking information appears early but at different rates across scales and objectives. At step 600, \zall{} correlations are $0.85$, $0.90$, $0.85$, $0.76$, and $0.63$ from 0.6B through 14B. The 1.7B ranking is already highly stable at the first checkpoint, while 14B improves more gradually. Several trajectories fluctuate after becoming informative, which is why we use early scores to allocate additional training rather than prescribe one universal stopping checkpoint.

\subsubsection{Early-selection regret}
\begin{figure}[h]
\centering
\includegraphics[width=0.72\linewidth]{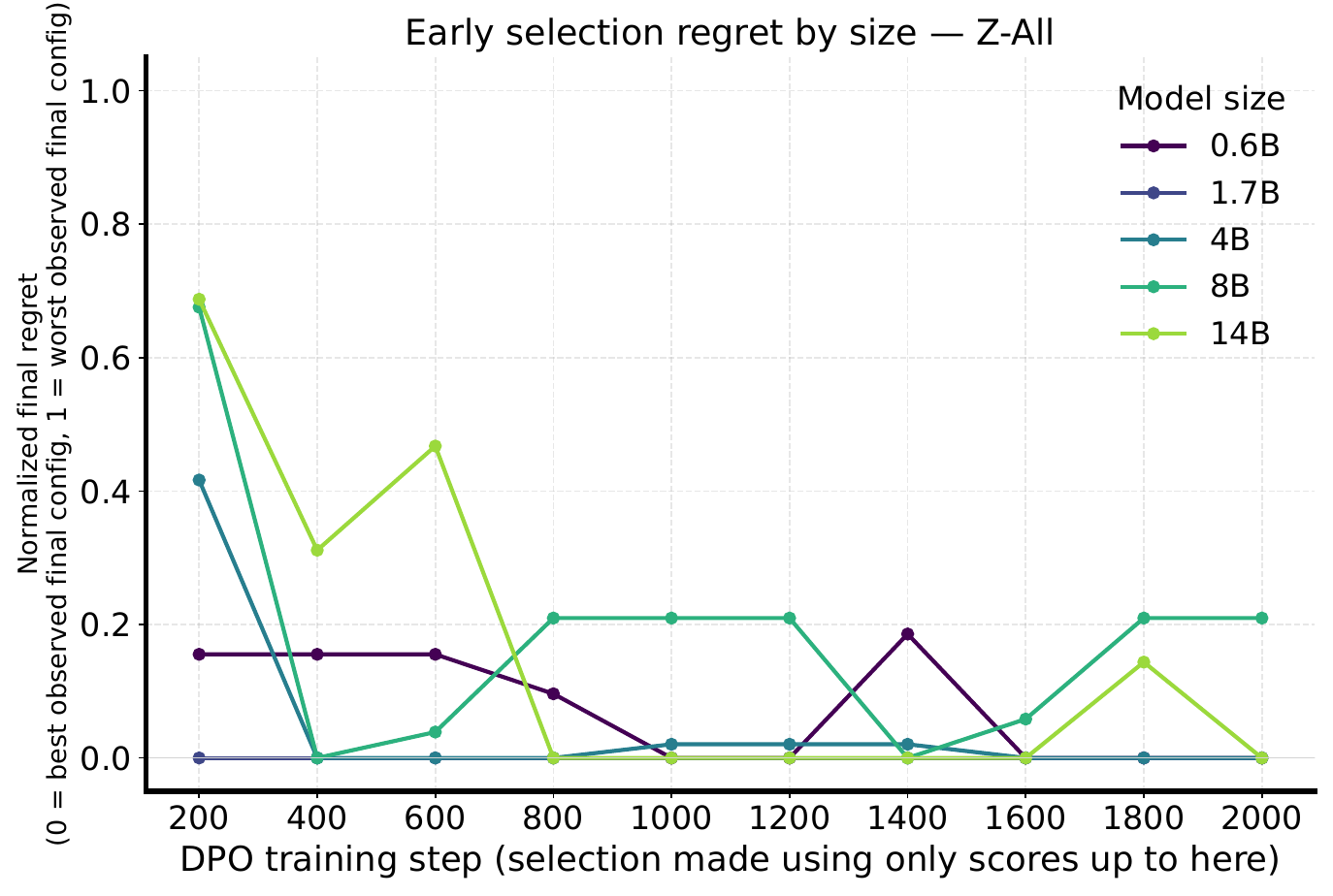}
\caption{\textbf{Early target-model selection can reduce tuning cost, but its reliability varies by scale.} Final normalised \zall{} regret when the configuration is selected using only the scores at each intermediate checkpoint. Regret is zero for the best final configuration and one for the worst.}
\label{fig:qwen3-early-selection}
\end{figure}

Rank correlation measures the full ordering, whereas configuration selection depends mainly on the top of that ordering. Figure~\ref{fig:qwen3-early-selection} therefore evaluates the final regret of the configuration that would be selected at each checkpoint. For this cross-sectional calculation, benchmark means and standard deviations are recomputed separately at every checkpoint over the configurations available at that exact step. We then select the highest-scoring \zall{} configuration and measure its final \zall{} regret relative to the best and worst final configurations in that model-size grid.

Low or zero regret can be reached well before exhaustive training, but the best selection checkpoint is not monotonic or shared across sizes. This variability motivates multi-fidelity allocation rather than a fixed early-stopping rule.

\subsubsection{UltraFeedback preference-data ablation}
\label{app:ultrafeedback}

We repeat the shared nine-configuration DPO grid on UltraFeedback at all five Qwen3 model sizes. These experiments retain the corresponding Dolci-trained SFT checkpoint, DPO objective, and evaluation suite, while using UltraFeedback and its dataset-specific sequence-length settings. Table~\ref{tab:ultrafeedback-transfer} compares the \zall{} winners on the common grid and measures the regret from transferring each dataset's selected configuration to the other dataset.

\begin{table}[t]
\centering
\small
\caption{\textbf{Preference-data sensitivity across Qwen3 sizes.} \zall{} winners are selected from the shared $3\times3$ grid. Transfer regret evaluates each dataset's winner on the other dataset and normalizes using the target dataset's best and worst shared-grid configurations. All Dolci winners use the lower tested $\beta$ boundary, while all UltraFeedback winners use the upper tested learning-rate boundary; the latter also select the upper $\beta$ boundary at 1.7B--14B.}
\label{tab:ultrafeedback-transfer}
\begin{tabular}{lcccc}
\toprule
Size & Dolci winner & UltraFeedback winner &
Dolci $\rightarrow$ UF & UF $\rightarrow$ Dolci \\
\midrule
0.6B & $(4\times10^{-6},0.01)$ & $(4\times10^{-6},0.01)$ & $0.00$ & $0.00$ \\
1.7B & $(4\times10^{-6},0.01)$ & $(4\times10^{-6},0.04)$ & $0.25$ & $0.11$ \\
4B & $(2\times10^{-6},0.01)$ & $(4\times10^{-6},0.04)$ & $0.67$ & $0.92$ \\
8B & $(1\times10^{-6},0.01)$ & $(4\times10^{-6},0.04)$ & $1.00$ & $1.00$ \\
14B & $(1\times10^{-6},0.01)$ & $(4\times10^{-6},0.04)$ & $1.00$ & $1.00$ \\
\bottomrule
\end{tabular}
\end{table}

The shared-grid winners agree at 0.6B, differ only in $\beta$ at 1.7B, and differ in both learning rate and $\beta$ at 4B--14B. Cross-dataset regret follows the same pattern: it is zero at 0.6B, limited at 1.7B, and substantial at 4B--14B. This five-size comparison strengthens the evidence for preference-data sensitivity, but does not establish a general scaling relationship. Moreover, every UltraFeedback \zall{} winner uses the upper learning-rate boundary, $4\times10^{-6}$; the 1.7B--14B winners also use the upper tested $\beta$. These results therefore identify preferences within the shared grid.

At 0.6B, \zstatic{} varies over a much narrower range than \zdynamic{}, making its particular argmax less informative. The stronger and more consistent UltraFeedback result is that \zdynamic{} and \zall{} both favour the upper tested learning rates at all sizes.

\begin{figure}
\centering
\begin{minipage}{0.49\textwidth}
    \centering
    \includegraphics[width=\linewidth]{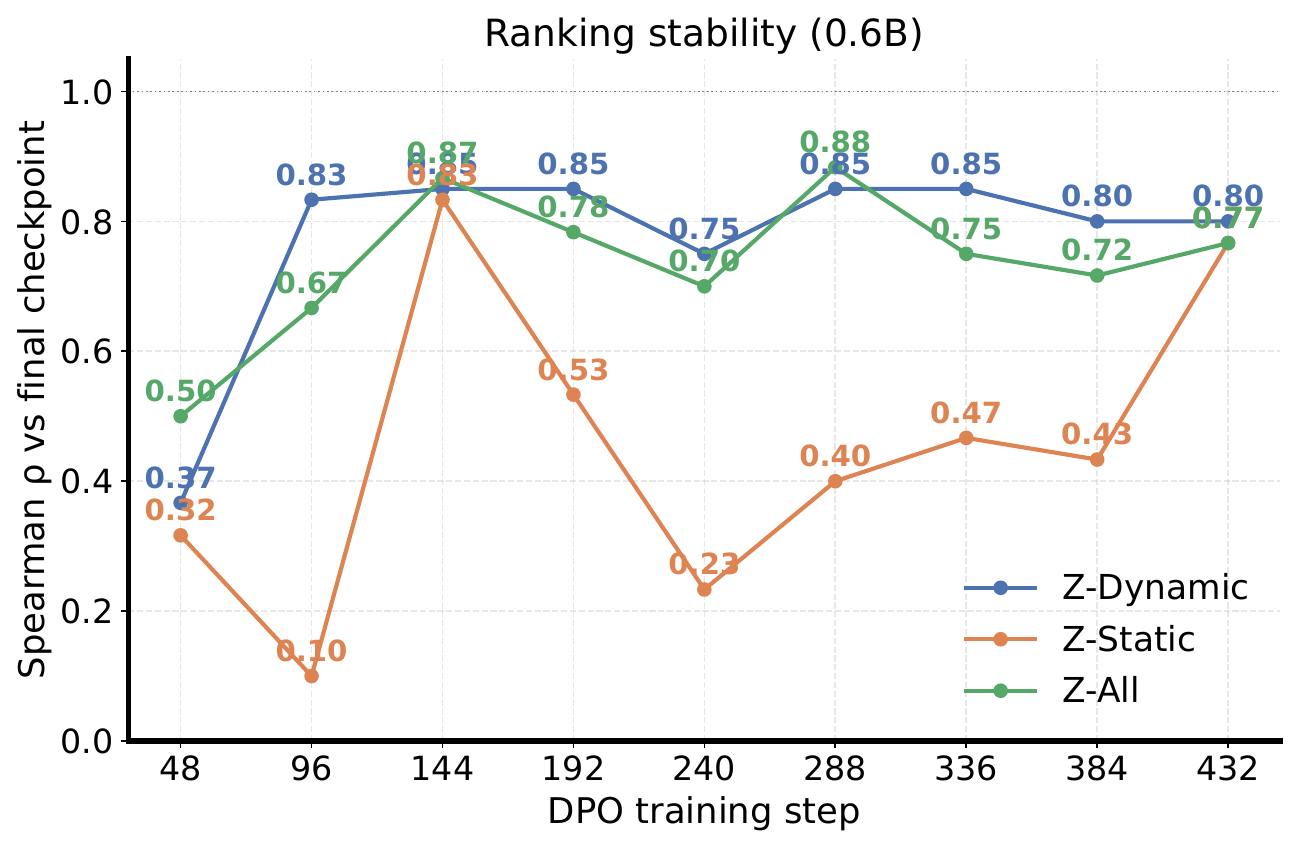}
\end{minipage}
\hfill
\begin{minipage}{0.49\textwidth}
    \centering
    \includegraphics[width=\linewidth]{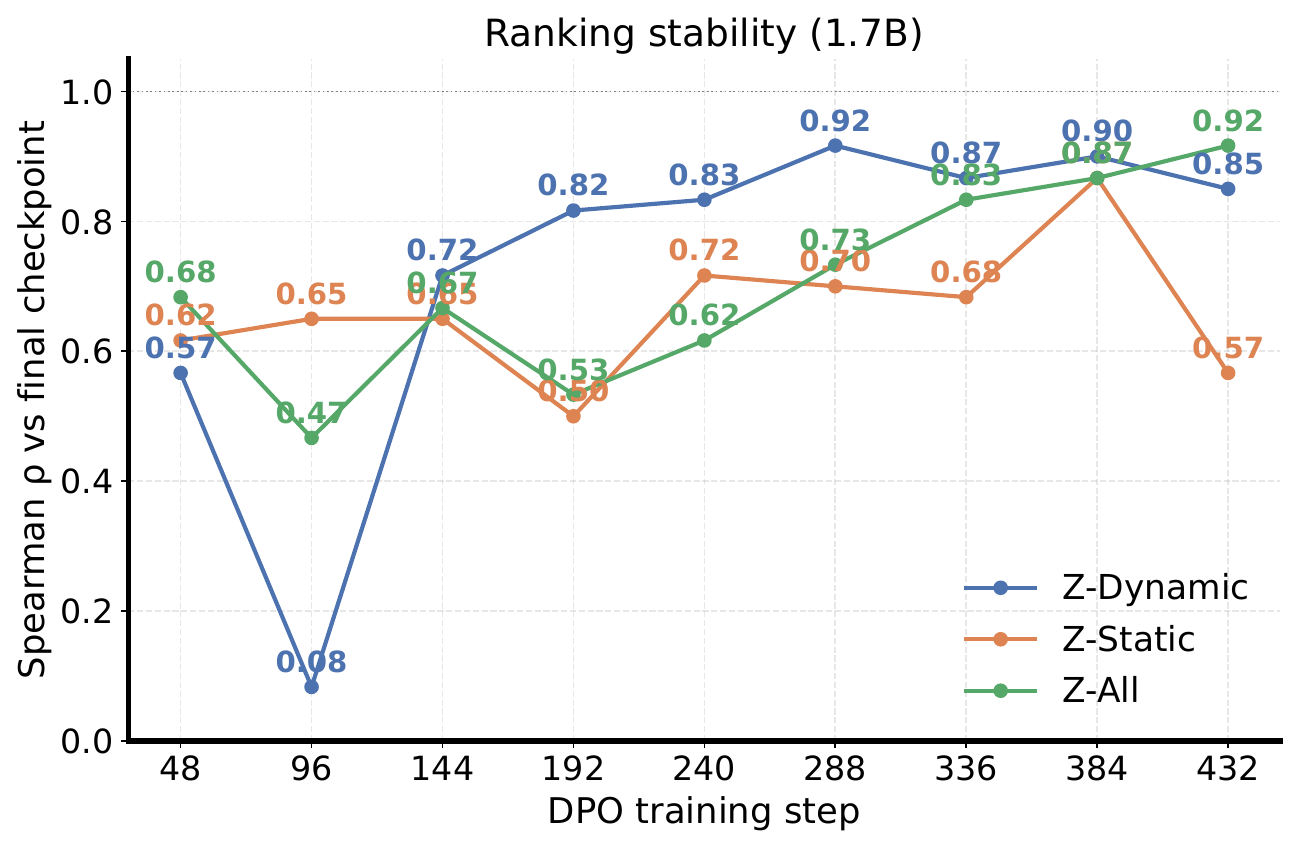}
\end{minipage}

\begin{minipage}{0.49\textwidth}
    \centering
    \includegraphics[width=\linewidth]{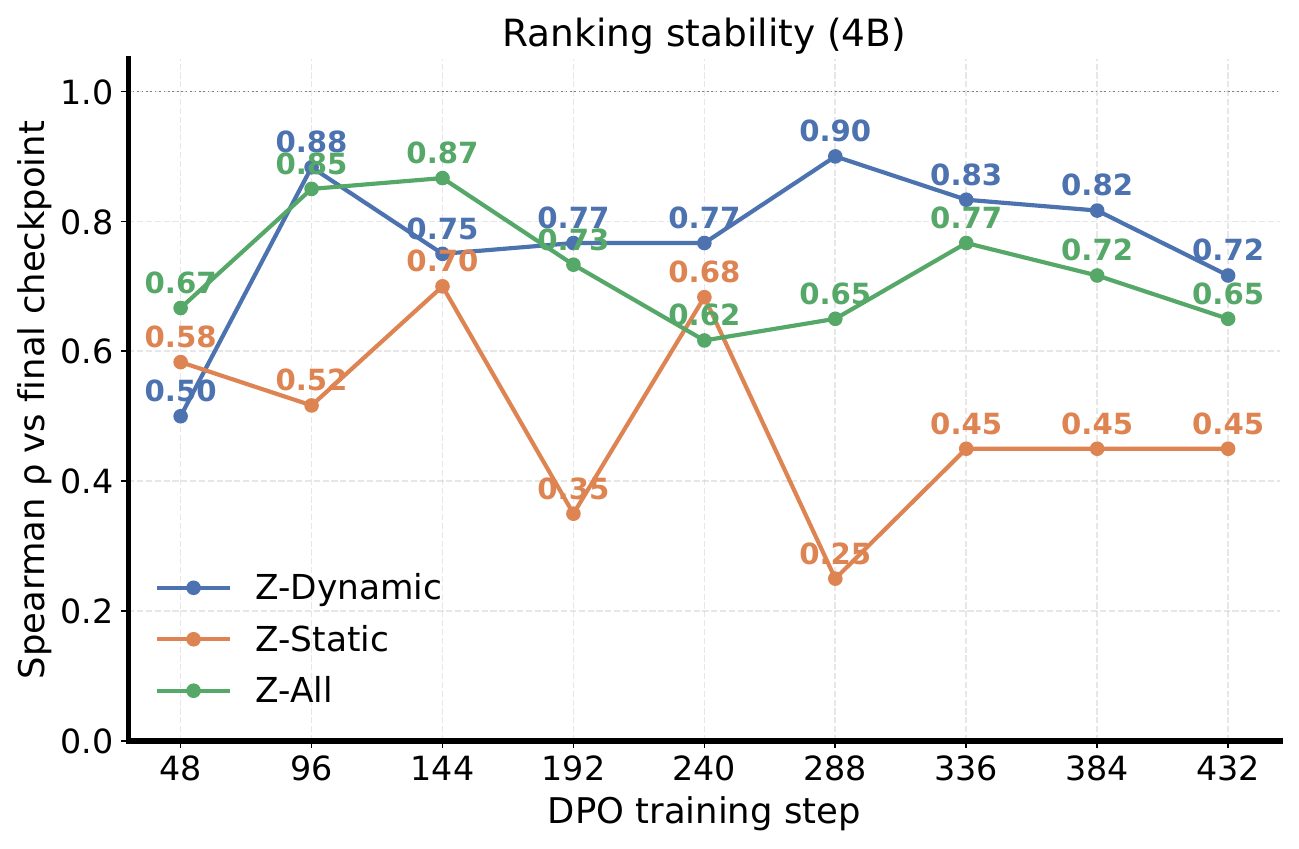}
\end{minipage}
\hfill
\begin{minipage}{0.49\textwidth}
    \centering
    \includegraphics[width=\linewidth]{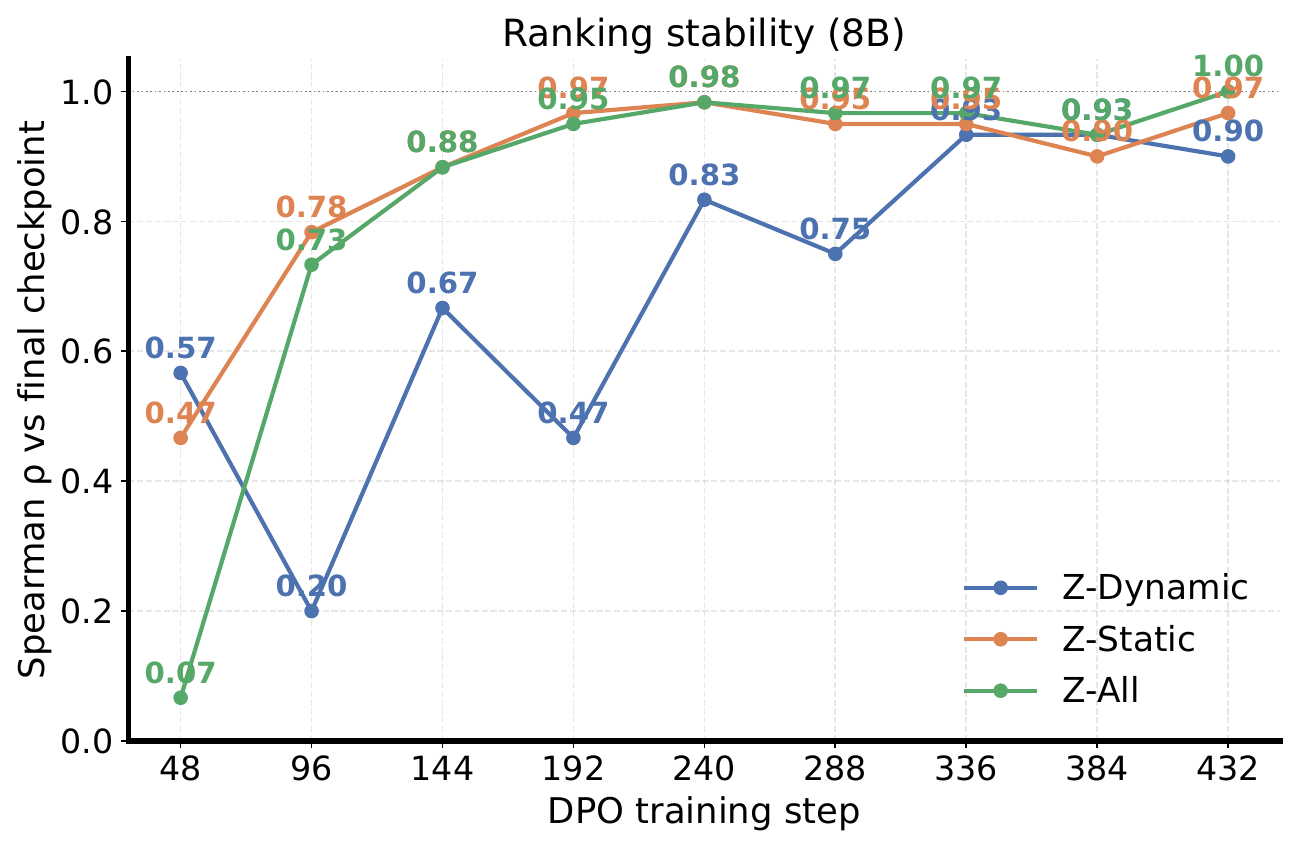}
\end{minipage}
\hfill
\begin{minipage}{0.49\textwidth}
    \centering
    \includegraphics[width=\linewidth]{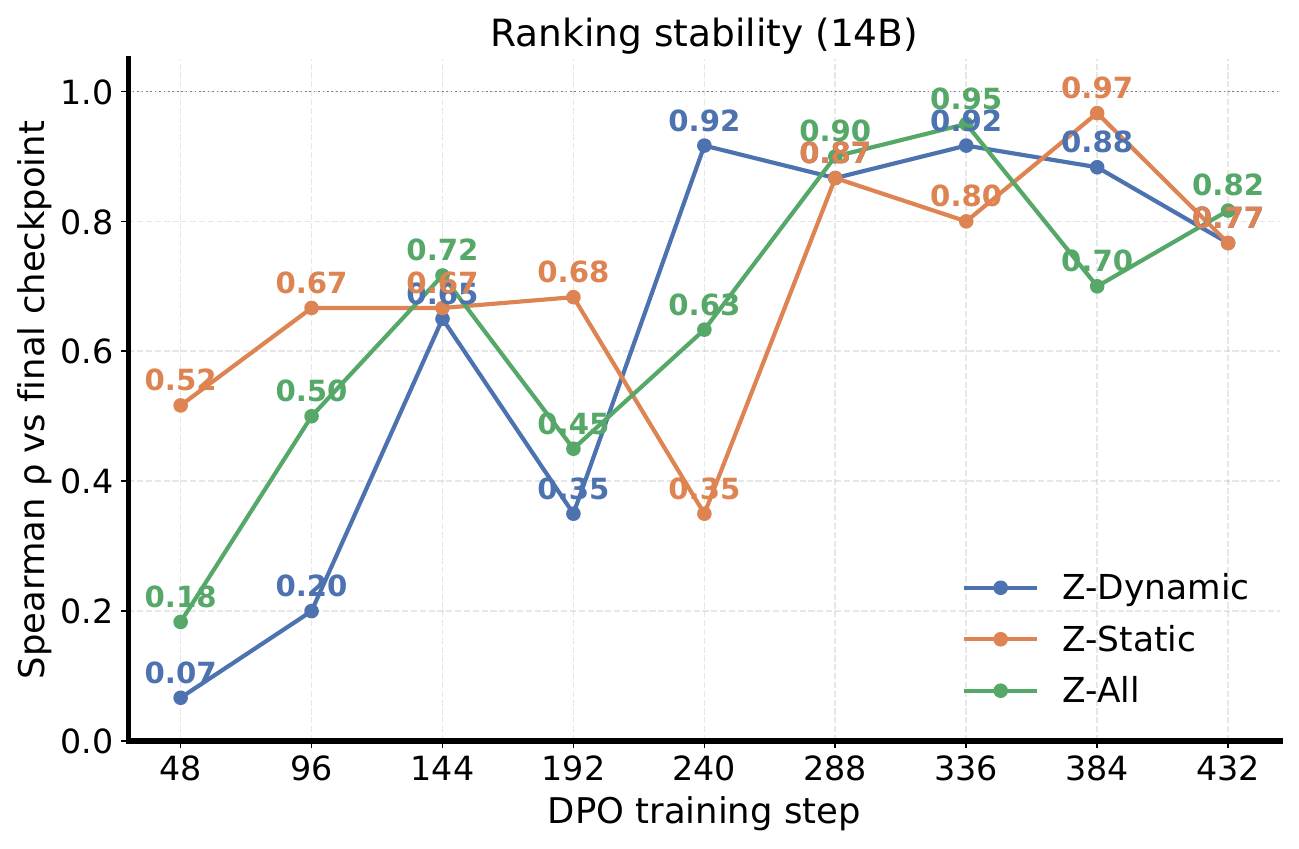}
\end{minipage}
\caption{\textbf{Intermediate-ranking stability across UltraFeedback model sizes.}
Spearman correlation between each intermediate configuration ranking and the final ranking on the shared nine-configuration grid. Dynamic rankings usually become informative by roughly $30\%$ of training, whereas static rankings fluctuate more across checkpoints and sizes.}
\label{fig:ultrafeedback-ranking-stability}
\end{figure}

Figure~\ref{fig:ultrafeedback-ranking-stability} compares intermediate and final rankings across the five UltraFeedback sizes. Dynamic-evaluation rankings are generally informative by 144 steps, roughly $30\%$ of training: correlations are at least $0.65$ for all sizes. The 1.7B dynamic ranking is a useful counterexample to a fixed earlier cutoff, falling to $0.08$ at step 96 before recovering at step 144. These results support using early target-model evaluations to prioritize candidates, while reinforcing that a fixed early-stopping point is not reliable across objectives or model sizes.

\paragraph{Benchmark agreement under UltraFeedback.}
The qualitative difference between dynamic and static benchmark agreement also holds under UltraFeedback. Across all tested sizes, dynamic benchmarks produce higher mean pairwise configuration-ranking correlations than static benchmarks (Table~\ref{tab:ultrafeedback-benchmark-agreement}). Thus, the greater consistency of chat-based evaluations is not specific to the Dolci-Instruct-DPO preference data.

\begin{table}
\centering
\small
\caption{\textbf{Benchmark-ranking agreement under UltraFeedback.} Mean pairwise Spearman correlation and population standard deviation across within-group benchmark pairs, computed separately at each size over the shared nine configurations.}
\label{tab:ultrafeedback-benchmark-agreement}
\begin{tabular}{lcc}
\toprule
Qwen3 size & Dynamic ($\bar{\rho}\pm\sigma$) & Static ($\bar{\rho}\pm\sigma$) \\
\midrule
0.6B & $0.49\pm0.21$ & $0.09\pm0.35$ \\
1.7B & $0.62\pm0.14$ & $0.40\pm0.34$ \\
4B   & $0.39\pm0.18$ & $0.16\pm0.48$ \\
8B   & $0.57\pm0.12$ & $0.53\pm0.31$ \\
14B   & $0.57\pm0.19$ & $-0.01\pm0.57$ \\
\bottomrule
\end{tabular}
\end{table}

\subsection{Llama Experiments}
\label{app:llama-experiments}

We repeat the principal analyses on Llama 3.2 at 1B and 3B parameters and on Llama 3.1 at 8B, using the same Dolci-Instruct SFT and preference data, training objective, nine-configuration core, and evaluation suite as in the Qwen3 experiments. Because the 8B model belongs to a different Llama release, we use it as a complementary transfer test rather than treating the three sizes as observations from one scaling series.

\subsubsection{Final aggregate optima}
Table~\ref{tab:llama-objective-optima} reports the best observed final configuration under each aggregate. Under \zall{}, the selected learning rate rises from $2\times10^{-6}$ at Llama~3.2-1B to $4\times10^{-6}$ at Llama~3.2-3B, matching the small-scale rise observed for Qwen3. We do not fit a Llama learning-rate scaling law, since the 8B experiment changes both model size and model version. The differences across columns also reinforce that an ``optimal'' DPO configuration is conditional on the evaluation objective: \zdynamic{} and \zstatic{} select different learning rates at every Llama size.

\begin{table}
\centering
\small
\caption{\textbf{Best observed final Llama configuration under each aggregate.} Each entry is $(\mathrm{learning\ rate},\beta)$, selected from the nine-configuration grid. $\downarrow$ and $\uparrow$ mark the lowest
and highest tested learning rates, respectively; $\beta=0.01$ is
the lower tested bound wherever selected. Boundary selections do
not locate unconstrained optima.}
\label{tab:llama-objective-optima}
\begin{tabular}{lccc}
\toprule
Model & \zall{} & \zdynamic{} & \zstatic{} \\
\midrule
Llama 3.2-1B & $(2\times10^{-6},0.01)$ & $(4\times10^{-6},0.01)^{\uparrow}$ & $(2\times10^{-6},0.01)$ \\
Llama 3.2-3B & $(4\times10^{-6},0.01)^{\uparrow}$ & $(2\times10^{-6},0.01)$ & $(4\times10^{-6},0.01)^{\uparrow}$ \\
Llama 3.1-8B & $(4\times10^{-6},0.02)^{\uparrow}$ & $(1\times10^{-6},0.02)^{\downarrow}$ & $(4\times10^{-6},0.02)^{\uparrow}$ \\
\bottomrule
\end{tabular}
\end{table}

\subsubsection{Benchmark agreement and sensitivity to AlpacaEval}
Table~\ref{tab:llama-benchmark-agreement} gives the mean and population standard deviation of pairwise benchmark-ranking correlations. Using all four chat benchmarks, chat agreement exceeds static agreement at 1B and 3B. At 8B, however, AlpacaEval ranks the configurations differently from the other three chat evaluations: the mean chat correlation is $0.32$ with a standard deviation of $0.61$. Excluding AlpacaEval in a sensitivity analysis raises the 8B chat mean to $0.92$ and reduces its standard deviation to $0.04$. The same exclusion has much smaller effects at 1B and 3B. This does not determine which evaluation is correct, but it shows that an aggregate can conceal disagreement specific to one benchmark and model setting. The selected 8B \zall{} and \zdynamic{} configurations are unchanged by this exclusion, even though the agreement statistics change substantially. One plausible contributor is the evaluation mechanism: AlpacaEval is the only chat evaluation in our study implemented as a one-token pairwise classifier, with its preference recovered from the relative probabilities of two label tokens and no generated rationale. This protocol difference may make its rankings more sensitive to prompt or calibration effects. However, our comparison does not isolate this mechanism from other differences in prompts, reference responses, or aggregation, so we treat it solely as our hypothesis.

\begin{table}
\centering
\small
\caption{\textbf{Agreement between Llama configuration rankings.} Entries are the mean pairwise Spearman correlation $\rho$ and population standard deviation $\sigma$ across benchmark pairs. ``Chat without AE'' excludes AlpacaEval 2.0 as a sensitivity analysis.}
\label{tab:llama-benchmark-agreement}
\begin{tabular}{lccc}
\toprule
Model & Chat & Chat without AE & Static \\
\midrule
Llama 3.2-1B & $0.65\pm0.18$ & $0.54\pm0.13$ & $0.05\pm0.62$ \\
Llama 3.2-3B & $0.84\pm0.09$ & $0.87\pm0.07$ & $0.37\pm0.57$ \\
Llama 3.1-8B & $0.32\pm0.61$ & $0.92\pm0.04$ & $0.01\pm0.66$ \\
\bottomrule
\end{tabular}
\end{table}

Table~\ref{tab:llama-chat-pairs} reports the individual chat-benchmark pairs behind these summaries. The three comparisons that do not involve AlpacaEval remain strongly positive at 8B, while all three comparisons involving AlpacaEval are negative. At 1B and 3B, every chat pair is positively correlated.

\begin{table}
\centering
\small
\caption{\textbf{Pairwise Spearman correlations between Llama chat-benchmark rankings.} Each correlation is computed over the nine DPO configurations at that model size. The upper block contains the three pairs retained by the analysis without AlpacaEval 2.0.}
\label{tab:llama-chat-pairs}
\begin{tabular}{lccc}
\toprule
Benchmark pair & Llama 3.2-1B & Llama 3.2-3B & Llama 3.1-8B \\
\midrule
Arena-Hard--MT-Bench & $0.63$ & $0.85$ & $0.95$ \\
Arena-Hard--Elo & $0.63$ & $0.97$ & $0.95$ \\
MT-Bench--Elo & $0.35$ & $0.80$ & $0.87$ \\
\midrule
Arena-Hard--AlpacaEval & $0.95$ & $0.88$ & $-0.28$ \\
MT-Bench--AlpacaEval & $0.60$ & $0.68$ & $-0.42$ \\
AlpacaEval--Elo & $0.73$ & $0.88$ & $-0.17$ \\
\bottomrule
\end{tabular}
\end{table}

\subsubsection{Ranking stability and early selection}
Figure~\ref{fig:llama-ranking-stability} applies the fixed-final normalization described in Appendix~\ref{app:full-qwen3-results}: each benchmark's final-grid mean and population standard deviation are used to transform every checkpoint before comparing intermediate and final rankings. Llama~3.2-1B and 3B develop high ranking correlations early in training. The Llama~3.1-8B \zall{} trajectory is less stable and changes markedly when AlpacaEval is excluded, consistent with the pairwise disagreement above. This sensitivity concerns the ordering of the full grid; the final \zall{}-selected configuration remains $(4\times10^{-6},0.02)$ in both variants.

\begin{figure}
\centering
\begin{minipage}{0.49\textwidth}
    \centering
    \includegraphics[width=\linewidth]{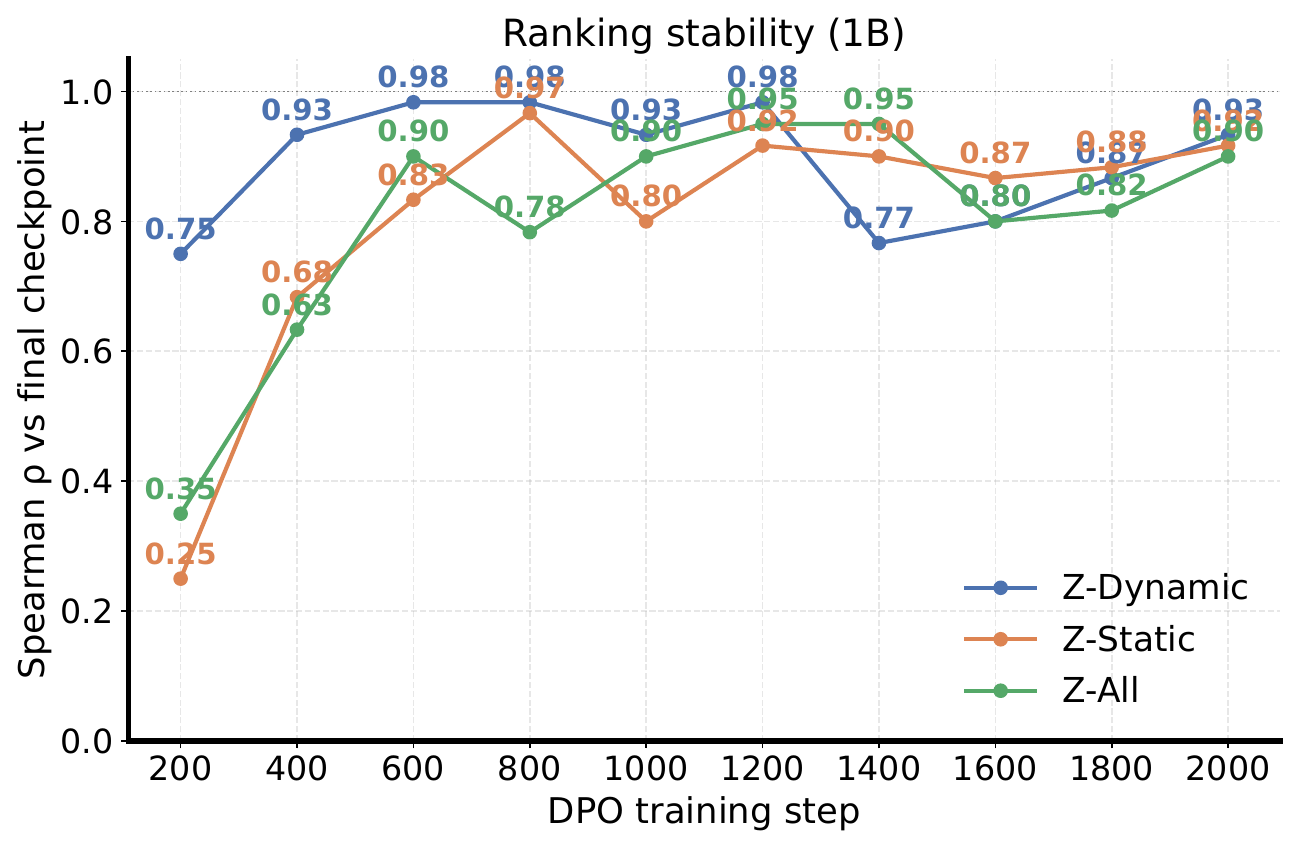}
\end{minipage}
\hfill
\begin{minipage}{0.49\textwidth}
    \centering
    \includegraphics[width=\linewidth]{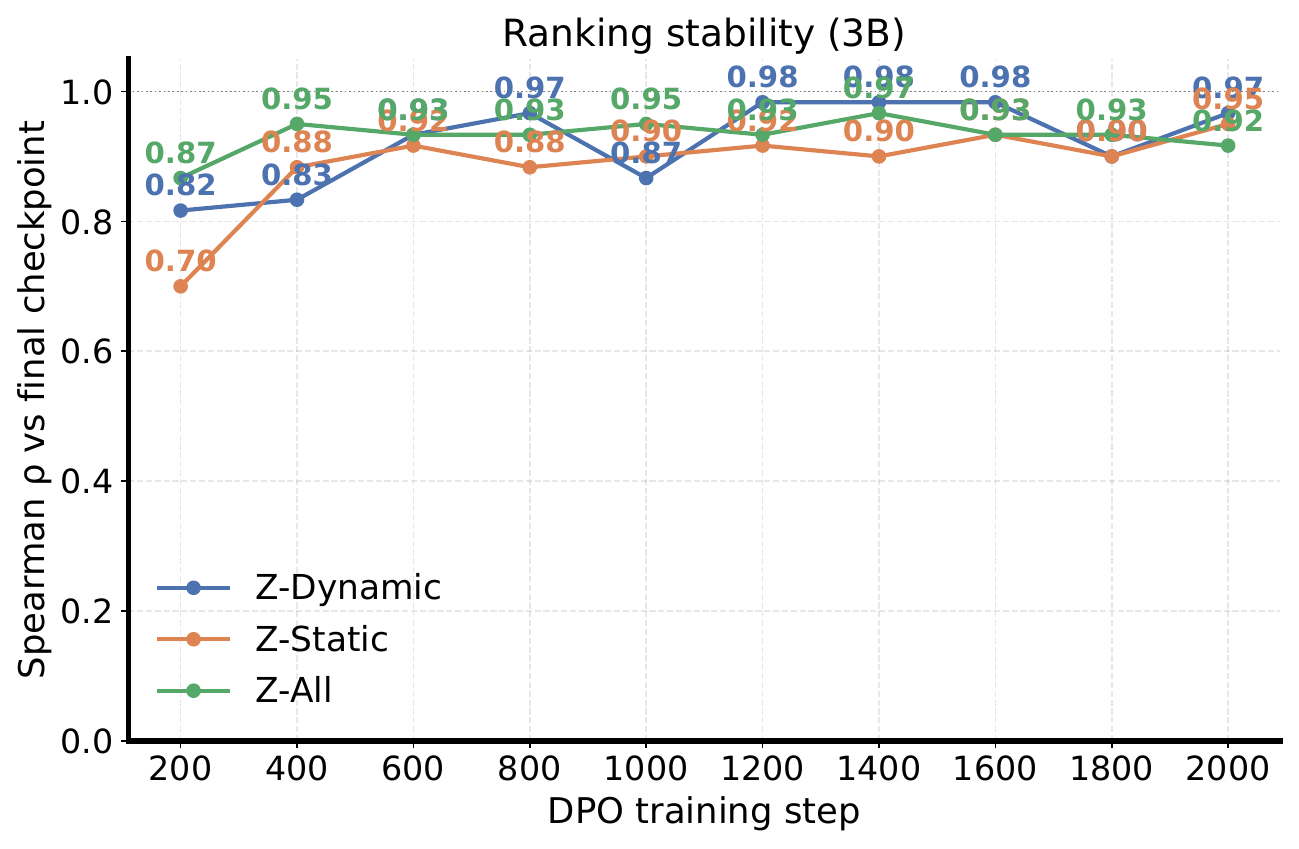}
\end{minipage}

\begin{minipage}{0.49\textwidth}
    \centering
    \includegraphics[width=\linewidth]{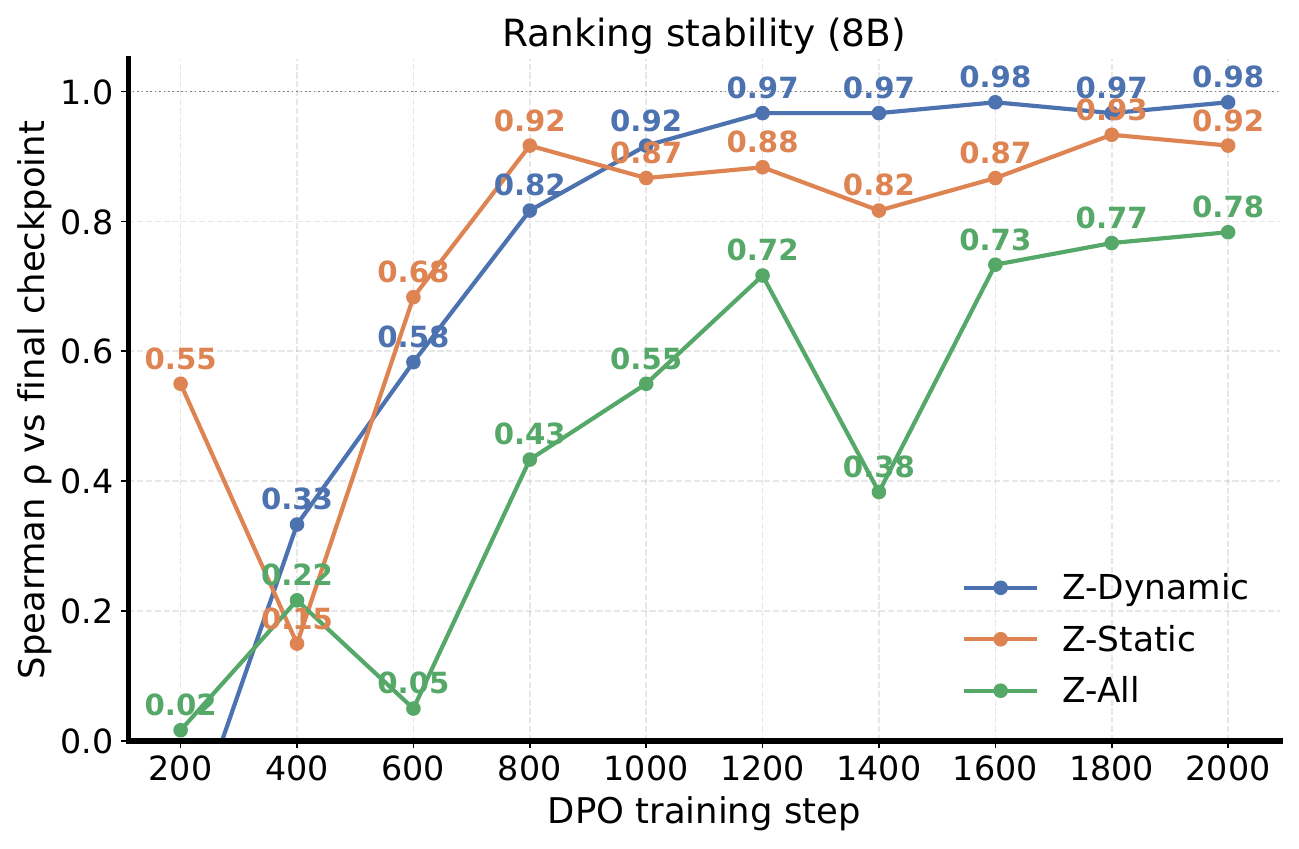}
\end{minipage}
\hfill
\begin{minipage}{0.49\textwidth}
    \centering
    \includegraphics[width=\linewidth]{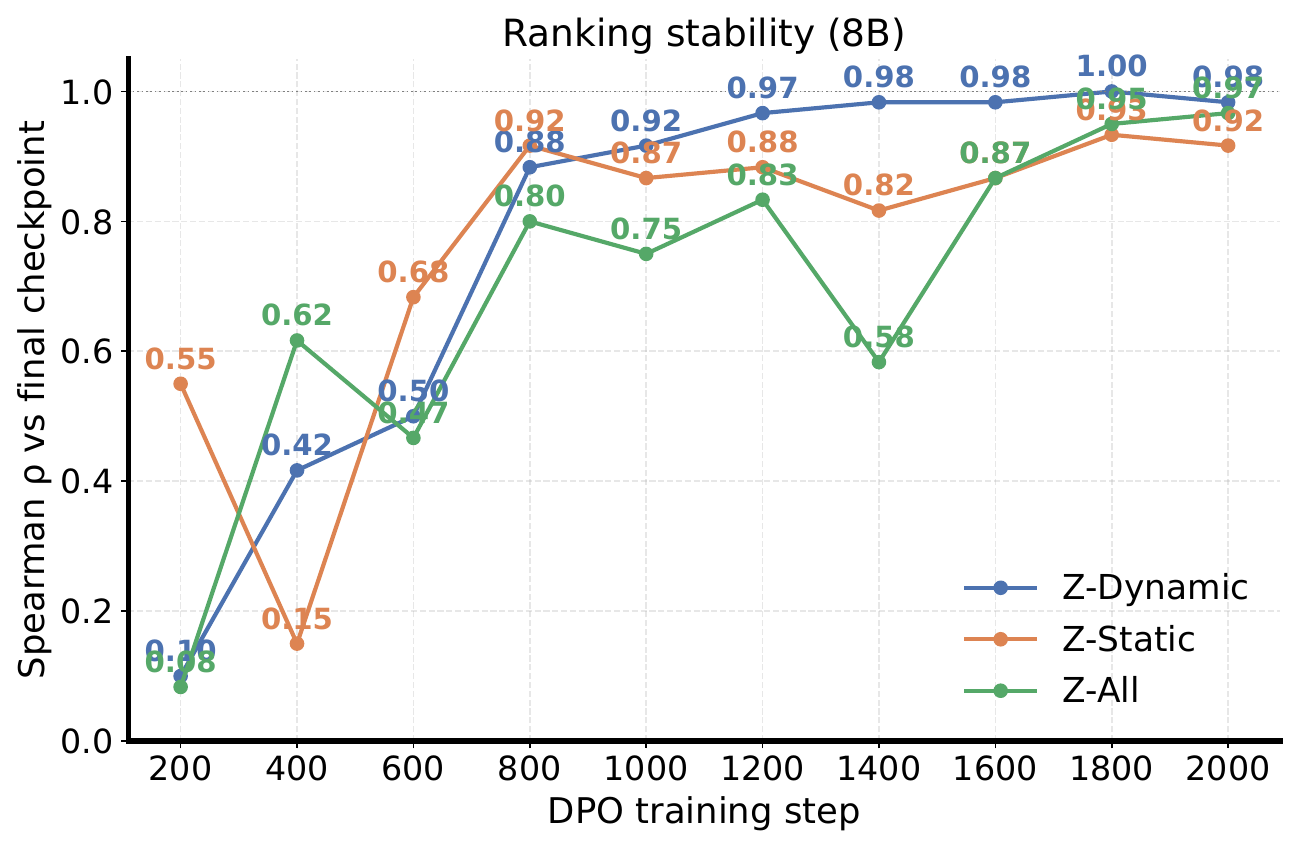}
\end{minipage}
\caption{\textbf{Intermediate-to-final ranking stability for the Llama grids.} The top row reports Llama~3.2-1B and 3B. The bottom row reports Llama~3.1-8B using the complete evaluation suite (left) and with AlpacaEval 2.0 excluded from \zall{} and \zdynamic{} (right). Lines give the Spearman correlation between intermediate and final configuration rankings.}
\label{fig:llama-ranking-stability}
\end{figure}

\begin{figure}[h]
\centering
\includegraphics[width=0.72\linewidth]{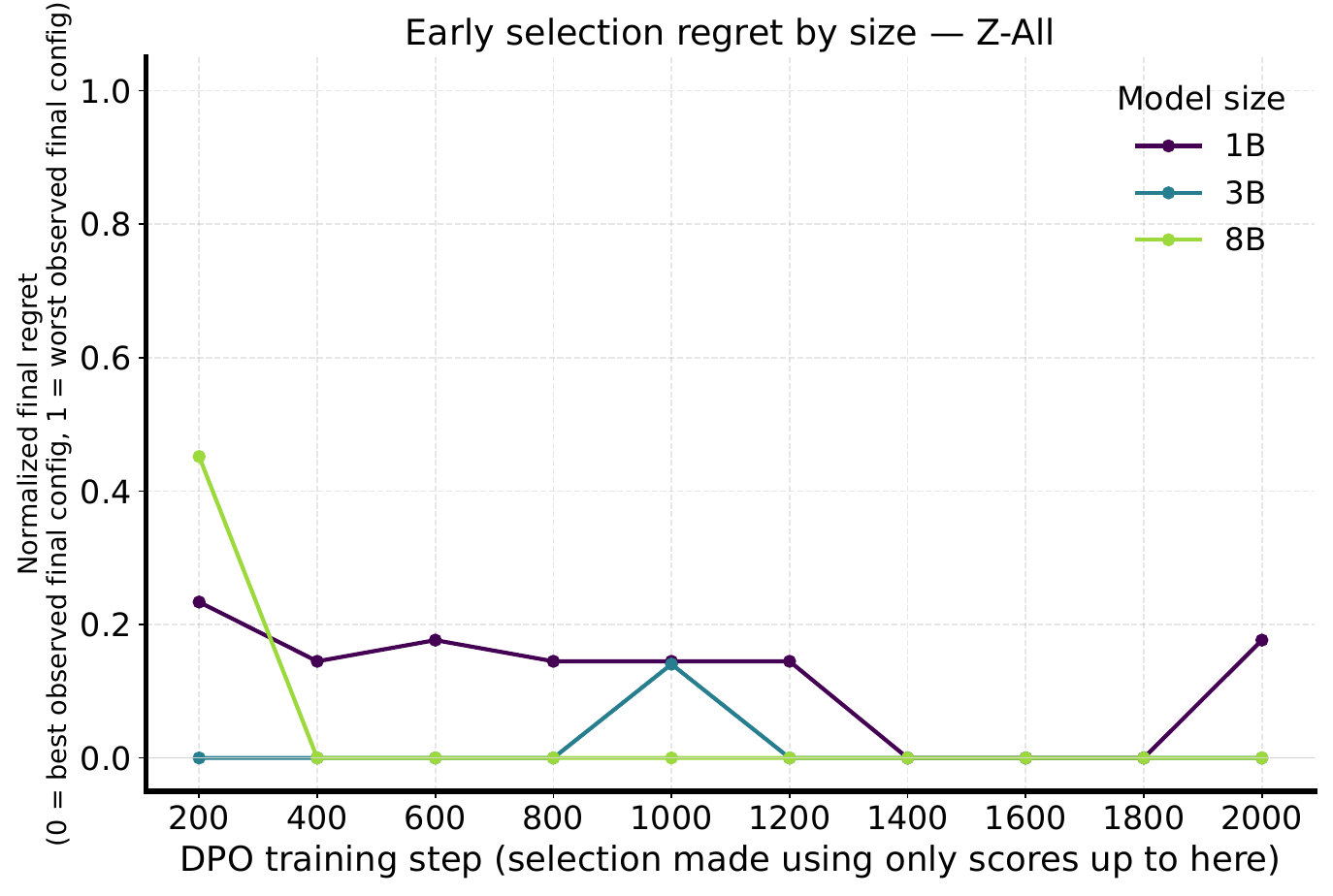}
\caption{\textbf{Early target-model selection on the Llama grids.} Final normalised \zall{} regret of the configuration selected at each intermediate checkpoint. Regret is zero for the best final configuration and one for the worst.}
\label{fig:llama-early-selection}
\end{figure}

Figure~\ref{fig:llama-early-selection} measures the top-of-ranking decision directly, using the per-checkpoint standardization defined for Figure~\ref{fig:qwen3-early-selection}. At step 400, approximately $20\%$ of a full run, early \zall{} selection identifies the eventual best configuration at 3B and 8B and incurs normalised regret of $0.145$ at 1B. The 1B optimum is first selected at step 1,400, and its selected configuration fluctuates again at the final intermediate checkpoint. Thus, the Llama results confirm that partial target-model runs can sharply reduce tuning cost, while also showing why a fixed stopping point cannot guarantee the best configuration at every scale.

\subsection{Judge Agreement Across Chat Evaluations}
\label{app:judge-agreement}

We compare configuration rankings rather than raw scores because the two judges need not use the same numerical scale. For each model size and chat benchmark, we restrict the comparison to configurations evaluated by both Qwen3-Next-80B-A3B-Instruct and Gemma 4 31B-it and compute their Spearman rank correlation. To compare the aggregate rankings, we standardize each judge's scores separately within the common configuration pool, average the four resulting benchmark z-scores into \zdynamic{}, and correlate the two \zdynamic{} rankings.

Table~\ref{tab:judge-agreement-full} reports the complete results. Aggregate agreement is high from 1.7B onward, with $\rho$ between $0.90$ and $0.98$, and is $0.79$ when all 52 jointly evaluated configurations are pooled. Agreement is not uniformly high at the benchmark level: for example, MT-Bench has $\rho=0.45$ at 14B and Elo has $\rho=0.57$ at 8B. Nevertheless, their aggregate rankings remain strongly correlated, suggesting that averaging several chat evaluations reduces sensitivity to judge-specific differences on an individual benchmark.

The 0.6B grid is the clear exception. Its \zdynamic{} rankings are negatively correlated ($\rho=-0.27$), driven particularly by Elo ($\rho=-0.73$) and, more weakly, MT-Bench ($\rho=-0.17$). The practical effect is also visible near the top of the ranking: among the five configurations preferred by each judge, the overlap is only one for Elo and two for MT-Bench, compared with three for Arena-Hard and AlpacaEval. At 4B, by contrast, the judges share four of their top five Arena-Hard configurations and all five for each of the other chat benchmarks. These comparisons support using a second judge for consequential decisions at the smallest scale, while also showing that high aggregate agreement does not imply identical judgments on every constituent benchmark.

\begin{table}
\centering
\small
\caption{\textbf{Agreement between configuration rankings produced by the two LLM judges on the shared Qwen3 core.} Entries are Spearman correlations over the nine learning-rate--$\beta$ configurations shared by every Qwen3 size. \zdynamic{} averages separately standardized Arena-Hard, MT-Bench, AlpacaEval 2.0, and Elo scores.}
\label{tab:judge-agreement-full}
\begin{tabular}{lcccccc}
\toprule
Qwen3 size & \zdynamic{} & Arena-Hard & MT-Bench & AlpacaEval & Elo \\
\midrule
0.6B & $-0.27$ & $0.53$ & $-0.17$ & $0.27$ & $-0.73$ \\
1.7B & $ 0.90$ & $0.98$ & $ 0.87$ & $0.95$ & $ 0.73$ \\
4B & $0.98$ & $0.85$ & $0.93$ & $0.97$ & $0.93$ \\
8B   & $ 0.90$ & $0.78$ & $ 0.52$ & $0.83$ & $ 0.57$ \\
14B  & $ 0.95$ & $0.83$ & $ 0.45$ & $0.88$ & $ 0.77$ \\
\bottomrule
\end{tabular}
\end{table}

\FloatBarrier
\subsection{Continuous-Surrogate HPO Results}
\label{app:cross-metric-hpo}

Table~\ref{tab:continuous-hpo-gains} reports the end-of-budget comparison. Every method searches \zdynamic{}, recommendations are selected using the common-fidelity rule in Appendix~\ref{app:hpo-settings}, and each recommendation is scored at final fidelity. Under the $8\times$ wall-clock budget, CQR is the only method with a paired 95\% interval excluding zero for every surrogate. BOHB has the largest mean improvement under the Gaussian-process surrogate ($0.098$), whereas CQR has the largest improvement under the 1-NN, 3-NN, and 5-NN environments ($0.09$, $0.11$, and $0.16$). TPE improves conclusively under the GP, 3-NN, and 5-NN surrogates; ASHA does so under the GP and 3-NN environments.

\begin{table*}[t]
\centering
\scriptsize
\setlength{\tabcolsep}{3.5pt}
\caption{\textbf{Final \zdynamic{} scores and paired improvements under the $8\times$ budget.} The random-search column gives its mean score and 95\% $t$-interval. Other columns give the paired difference from random search, after averaging the five sizes within each seed. Intervals are over 25 seed-level differences. Higher values are better; bold marks the largest mean improvement in each row. Comparisons use unrounded interval endpoints.}
\label{tab:continuous-hpo-gains}
\begin{tabular}{lccccc}
\toprule
Surrogate & RS score & TPE $\Delta$ & CQR $\Delta$ & ASHA $\Delta$ & BOHB $\Delta$ \\
\midrule
GP & \makecell{$1.07$\\$[1.04,1.10]$} & \makecell{$+0.052$\\$[0.005,0.100]$} & \makecell{$+0.077$\\$[0.034,0.120]$} & \makecell{$+0.083$\\$[0.050,0.116]$} & \makecell{$\mathbf{+0.098}$\\$[0.058,0.138]$} \\
$1$-NN & \makecell{$1.22$\\$[1.18,1.27]$} & \makecell{$+0.04$\\$[-0.02,0.11]$} & \makecell{$\mathbf{+0.09}$\\$[0.04,0.15]$} & \makecell{$+0.01$\\$[-0.05,0.07]$} & \makecell{$+0.03$\\$[-0.02,0.09]$} \\
$3$-NN & \makecell{$1.07$\\$[1.04,1.11]$} & \makecell{$+0.06$\\$[0.01,0.12]$} & \makecell{$\mathbf{+0.11}$\\$[0.07,0.16]$} & \makecell{$+0.04$\\$[0.01,0.08]$} & \makecell{$+0.05$\\$[0.02,0.09]$} \\
$5$-NN & \makecell{$0.95$\\$[0.93,0.98]$} & \makecell{$+0.09$\\$[0.03,0.14]$} & \makecell{$\mathbf{+0.16}$\\$[0.11,0.20]$} & \makecell{$+0.02$\\$[-0.01,0.05]$} & \makecell{$+0.06$\\$[0.03,0.10]$} \\
\bottomrule
\end{tabular}
\end{table*}

The pooled GP result is heterogeneous across sizes. Table~\ref{tab:gp-hpo-by-size} shows that CQR improves conclusively at 0.6B, 4B, and 8B, with its largest gain at 8B ($0.20$). At 14B, ASHA and BOHB have the largest conclusive gains ($0.29$ and $0.24$), whereas CQR is inconclusive. At 1.7B, none of the informed methods has a conclusive advantage. These differences reinforce the need to validate the search strategy on the target setting rather than infer its value from a single scale.

\begin{table}[t]
\centering
\small
\setlength{\tabcolsep}{5pt}
\caption{\textbf{Gaussian-process HPO improvement over random search by model size.} Entries are paired mean differences in final \zdynamic{} over 25 seeds. Bold entries have paired 95\% intervals excluding zero.}
\label{tab:gp-hpo-by-size}
\begin{tabular}{lrrrr}
\toprule
Qwen3 size & TPE & CQR & ASHA & BOHB \\
\midrule
0.6B & $\mathbf{+0.06}$ & $\mathbf{+0.05}$ & $\mathbf{+0.06}$ & $\mathbf{+0.08}$ \\
1.7B & $0.00$ & $+0.01$ & $+0.02$ & $+0.02$ \\
4B   & $+0.02$ & $\mathbf{+0.08}$ & $+0.04$ & $+0.05$ \\
8B   & $+0.10$ & $\mathbf{+0.20}$ & $0.00$ & $+0.10$ \\
14B  & $+0.08$ & $+0.06$ & $\mathbf{+0.29}$ & $\mathbf{+0.24}$ \\
\bottomrule
\end{tabular}
\end{table}

Under the restrictive one-unit budget, all five methods have statistically indistinguishable pooled \zdynamic{} scores for every surrogate. This budget permits only approximately two full-run equivalents across both workers and generally expires before the methods can move beyond their common initialization. We therefore treat it as evidence that adaptive search requires an initialization budget, rather than as a comparison of the optimisers themselves.

\begin{figure*}[t]
\centering
\includegraphics[width=0.92\textwidth]{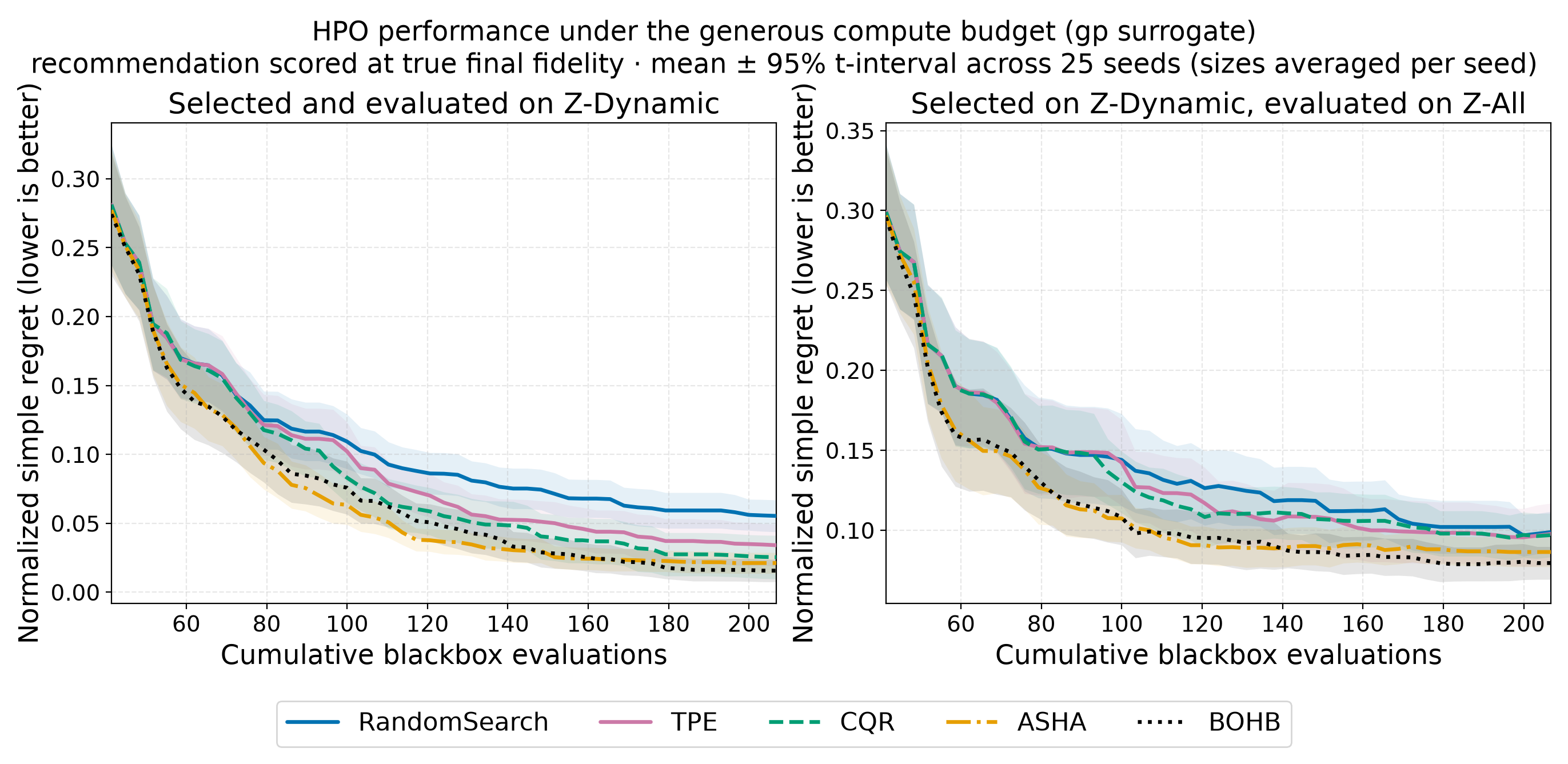}
\caption{\textbf{Recommendation regret under the $8\times$ budget with the Gaussian-process surrogate.} All methods select on \zdynamic{}. The left panel evaluates the recommendation on \zdynamic{}; the right evaluates the same recommendation on \zall{}. Curves average the five Qwen3 sizes within each seed and show the mean and 95\% $t$-interval across 25 seeds.}
\label{fig:hpo-gp-regret}
\end{figure*}

\begin{figure*}[t]
\centering
\includegraphics[width=0.92\textwidth]{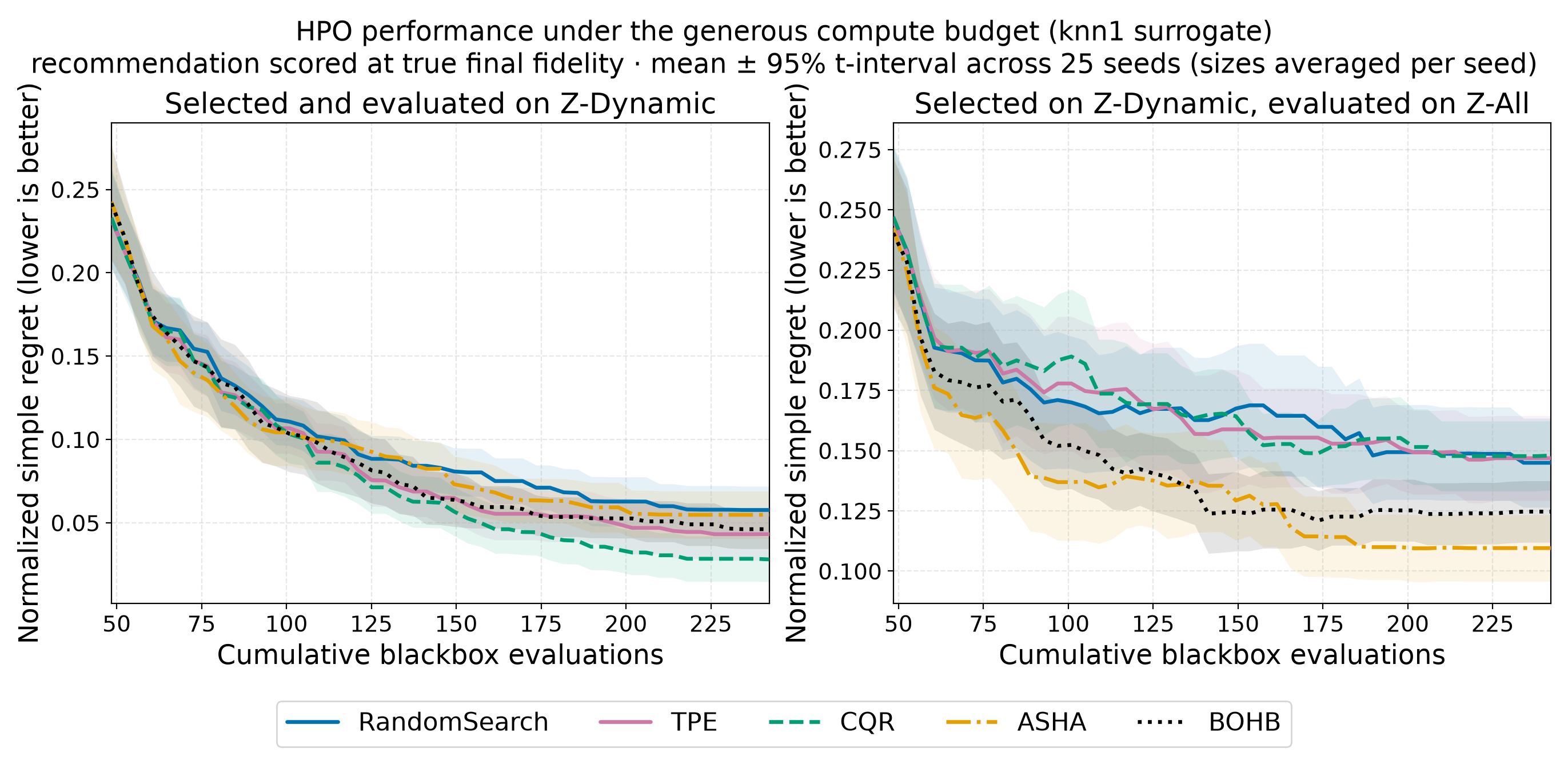}
\caption{\textbf{Recommendation regret under the $8\times$ budget with the one-nearest-neighbour surrogate.} All methods select on \zdynamic{}. The left panel evaluates the recommendation on \zdynamic{}; the right evaluates the same recommendation on \zall{}. Curves average the five Qwen3 sizes within each seed and show the mean and 95\% $t$-interval across 25 seeds. The one-nearest-neighbour surface is piecewise constant between measured configurations.}
\label{fig:hpo-knn1-regret}
\end{figure*}

\begin{figure*}[t]
\centering
\includegraphics[width=0.92\textwidth]{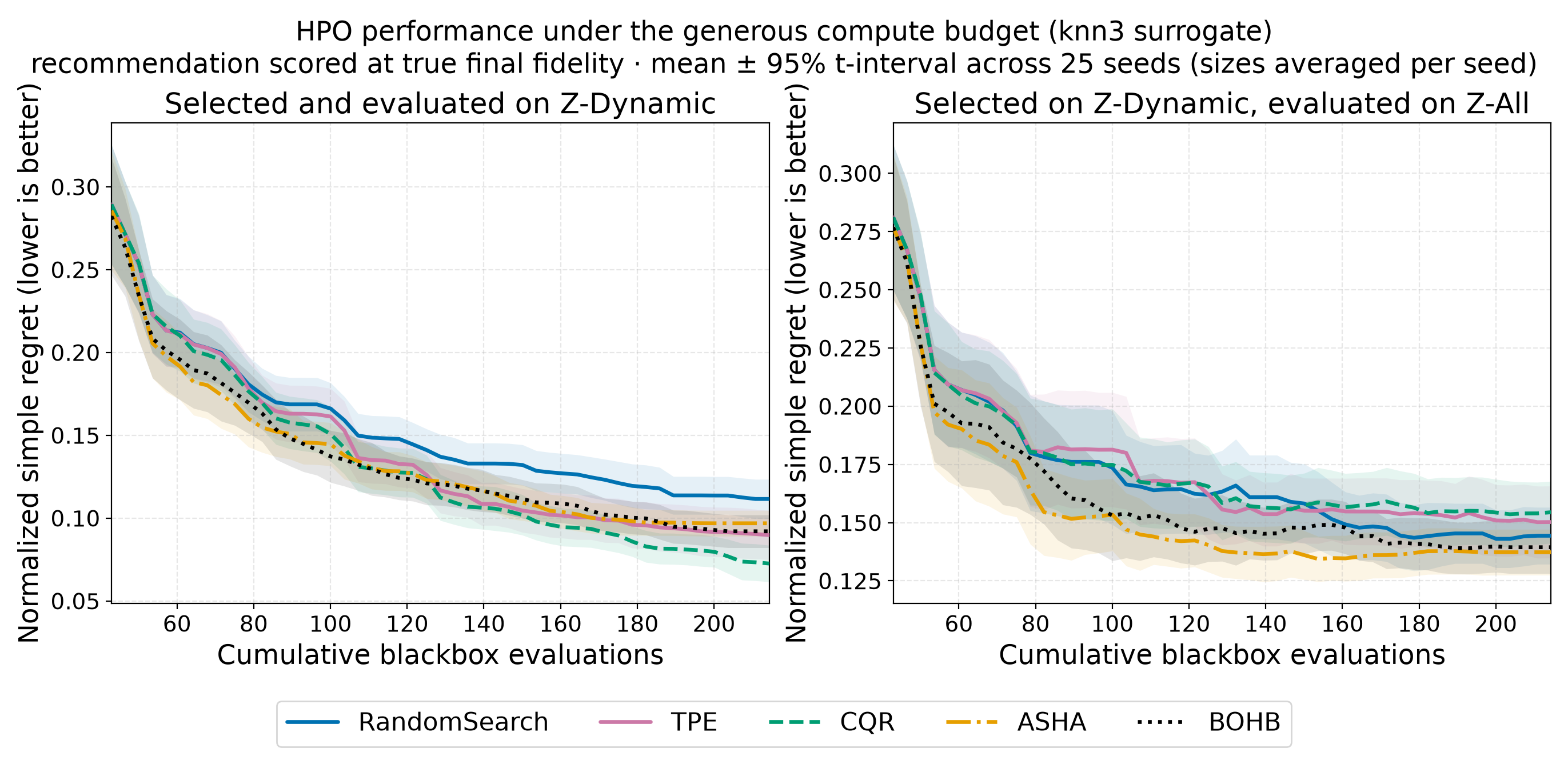}
\caption{\textbf{Recommendation regret under the $8\times$ budget with distance-weighted three-nearest-neighbour interpolation.} All methods select on \zdynamic{}. The left panel evaluates the recommendation on \zdynamic{}; the right evaluates the same recommendation on \zall{}. Curves average the five Qwen3 sizes within each seed and show the mean and 95\% $t$-interval across 25 seeds.}
\label{fig:hpo-knn3-regret}
\end{figure*}

\begin{figure*}[t]
\centering
\includegraphics[width=0.92\textwidth]{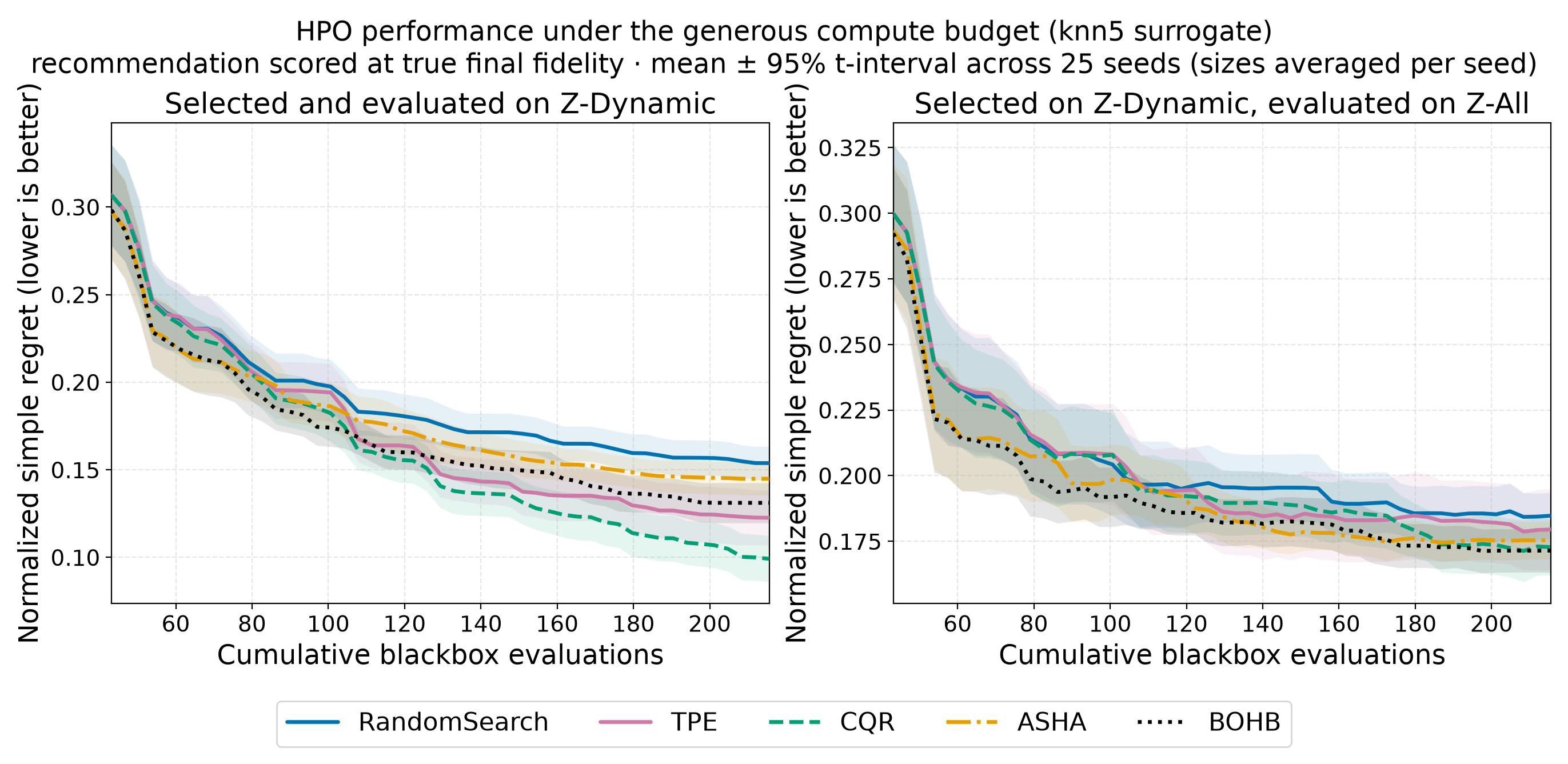}
\caption{\textbf{Recommendation regret under the $8\times$ budget with distance-weighted five-nearest-neighbour interpolation.} All methods select on \zdynamic{}. The left panel evaluates the recommendation on \zdynamic{}; the right evaluates the same recommendation on \zall{}. Curves average the five Qwen3 sizes within each seed and show the mean and 95\% $t$-interval across 25 seeds.}
\label{fig:hpo-knn5-regret}
\end{figure*}

Figures~\ref{fig:hpo-gp-regret}--\ref{fig:hpo-knn5-regret} show recommendation quality throughout the larger-budget runs. The larger budget is labelled ``generous'' inside these figures. These curves use only the configuration recommended at each cutoff and score it at final fidelity; they do not use the maximum checkpoint score observed so far. The curves show that relative ordering can differ by surrogate and can change when the \zdynamic{}-selected recommendation is evaluated on \zall{}. Per-size recommendation curves for every surrogate are included in the released artifact; Table~\ref{tab:gp-hpo-by-size} provides the corresponding compact GP summary.

These simulations interpolate the logged runs rather than adding new DPO observations. Agreement across the four independently fitted response surfaces makes the optimiser comparison less dependent on one interpolation scheme, but it does not remove uncertainty between measured configurations. We therefore interpret the results as evidence about search behaviour on the released offline benchmark, not as additional evidence about the exact location of the DPO optimum.

\subsection{Transferring Hyperparameters}

\subsubsection{Cross-scale ranking agreement by evaluation aggregate}
\label{app:cross-scale-ranking-agreement}

\begin{figure}
    \centering
    \includegraphics[width=0.8\textwidth]{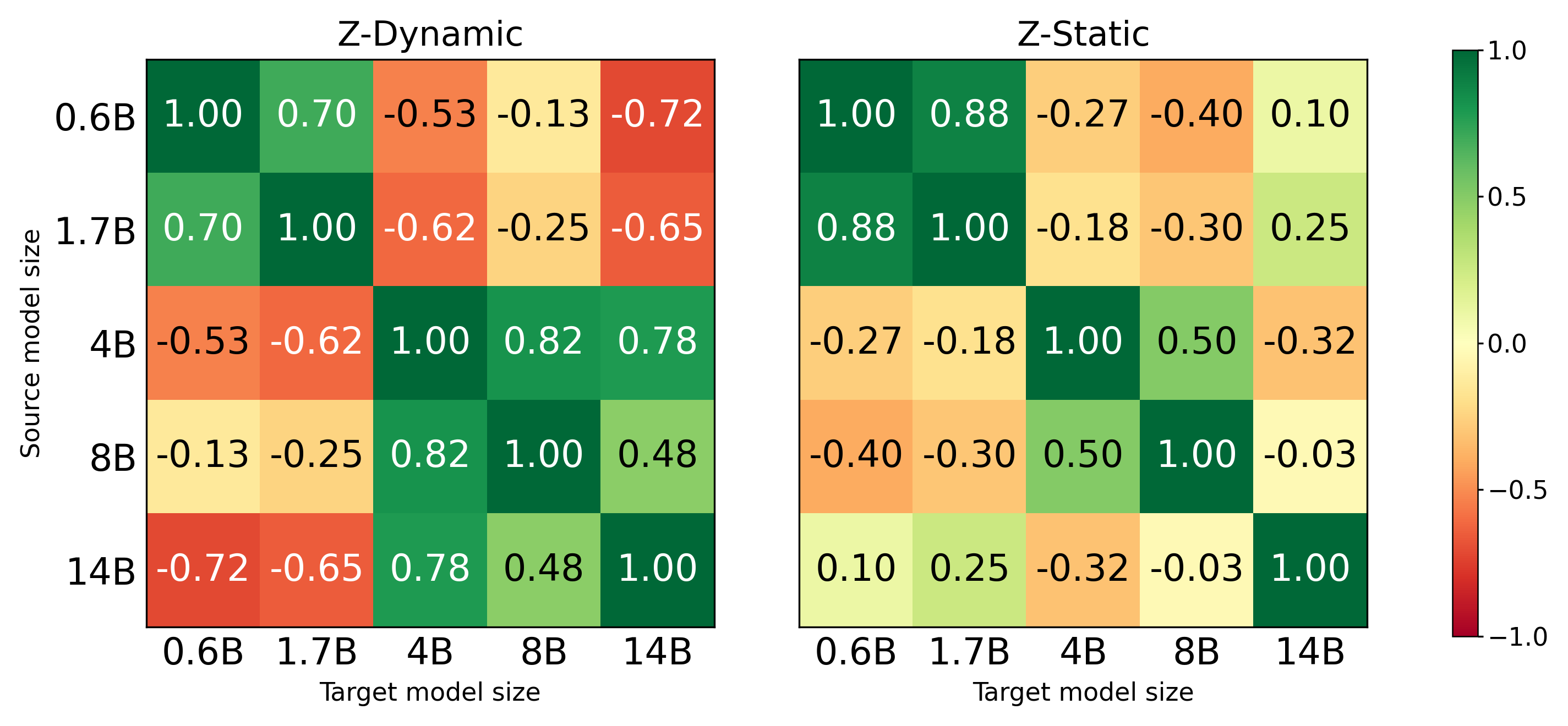}
    \caption{\textbf{Cross-scale configuration-ranking agreement by evaluation aggregate.}
    Each cell reports Spearman's $\rho$ between rankings of the nine configurations in the shared $3\times3$ learning-rate--$\beta$ grid. The left panel uses \zdynamic{} and the right panel uses \zstatic{}. Neither aggregate exhibits a consistent decline in agreement with model-size distance.}
    \label{fig:cross-scale-aggregate-agreement}
\end{figure}

The main paper reports cross-scale ranking agreement under \zdynamic{}. Figure~\ref{fig:cross-scale-aggregate-agreement} compares it with the static-benchmark \zstatic{} aggregates. The same lack of a simple size-based transfer pattern remains under both aggregates. Under \zdynamic{}, the two smallest models agree positively, as do the 4B, 8B, and 14B models, but rankings reverse between these groups. Under \zstatic{}, agreement is more heterogeneous and several adjacent model pairs have weak or negative correlation. Thus, the instability of configuration rankings across scale is not specific to the chat-benchmark aggregate.

\subsubsection{Directed cross-scale transfer}
\label{app:cross-scale-transfer}

We evaluate direct transfer on the nine configurations shared by all five Qwen3 grids. For each source size, we select its best final \zall{} configuration within this common set and evaluate that exact configuration at every target size. Normalised regret is computed relative to the target's best and worst configurations in the same shared set. Restricting both selection and evaluation to this common support avoids treating an unobserved target configuration as transferable. It also means that the 1.7B source uses its core-restricted optimum, $(4\times10^{-6},0.01)$, rather than its full-grid optimum at learning rate $8\times10^{-6}$.

\begin{figure*}[t]
    \centering
    \begin{minipage}{0.49\textwidth}
        \centering
        \includegraphics[width=\linewidth]{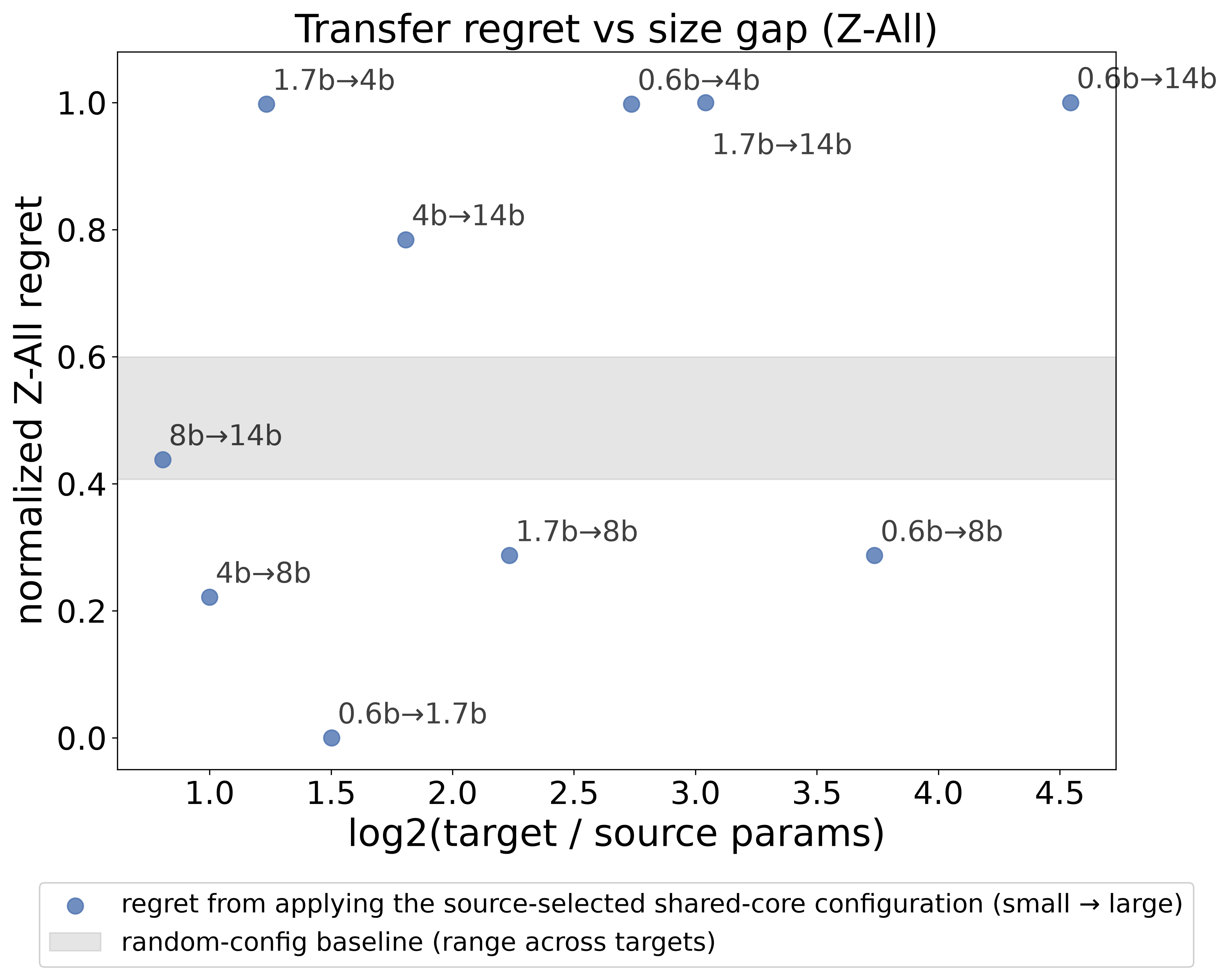}
    \end{minipage}
    \hfill
    \begin{minipage}{0.49\textwidth}
        \centering
        \includegraphics[width=\linewidth]{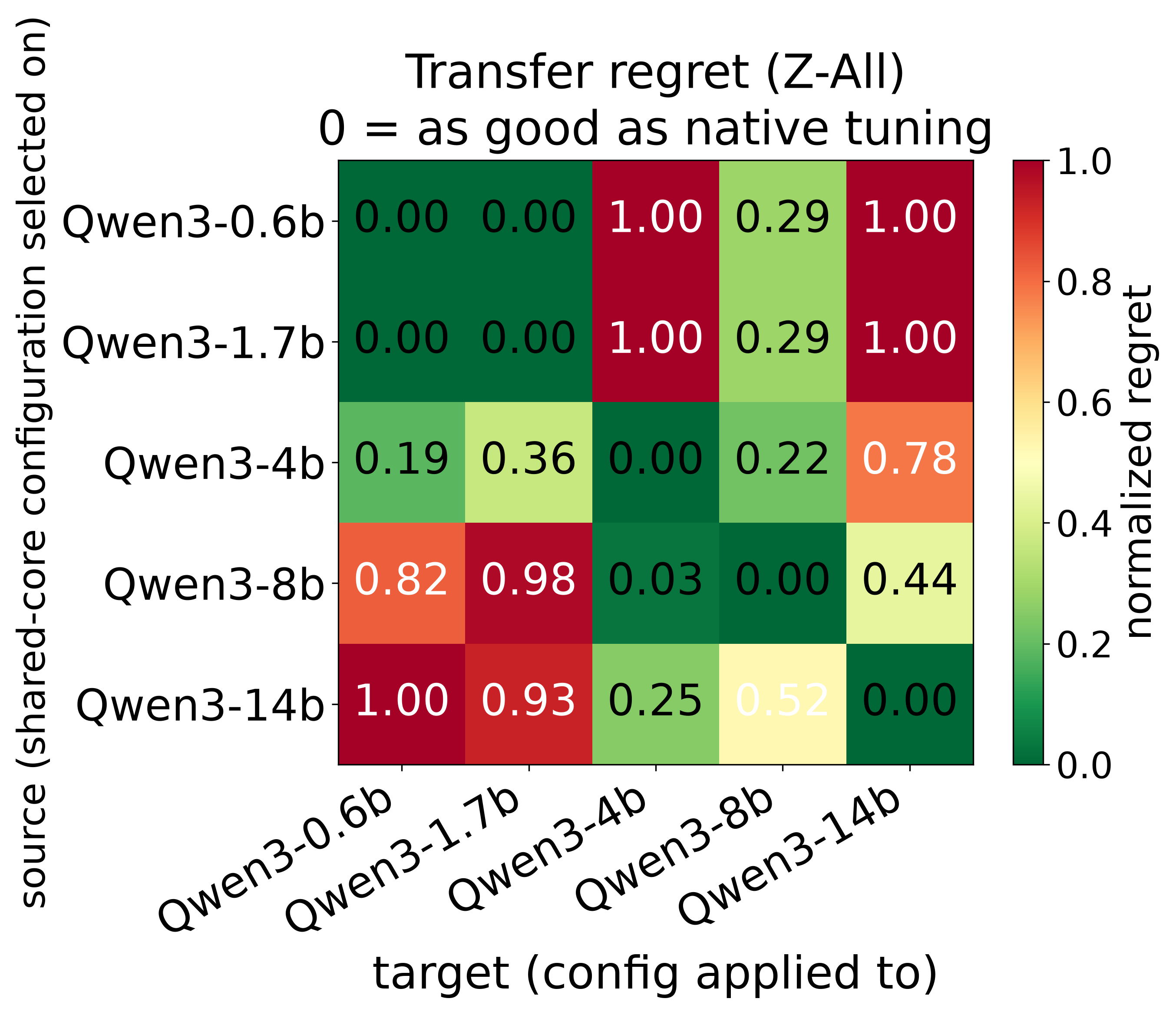}
    \end{minipage}
    \caption{\textbf{Cross-scale transfer regret is neither determined by model-size distance nor symmetric.}
    \textbf{Left:} Normalised \zall{} regret for smaller-to-larger transfers as a function of the log$_2$ parameter-count ratio. The shaded region gives the range of expected regret from choosing a shared-core configuration uniformly at random across target sizes.
    \textbf{Right:} The complete matrix of directed transfers. Each cell applies the source model's best \zall{} configuration from the shared $3\times3$ grid to the target model. Regret of zero denotes the target's best shared-core configuration and one its worst. Low and near-maximal regret occur at similar scale distances, while reversing the transfer direction can substantially change regret.}
    \label{fig:cross-scale-transfer-regret}
\end{figure*}

Figure~\ref{fig:cross-scale-transfer-regret} gives both the smaller-to-larger transfers and the complete directed matrix. Model-size distance is not predictive: 0.6B$\rightarrow$1.7B has zero regret, while 1.7B$\rightarrow$4B has regret $0.997$ despite being a smaller scale change; 0.6B$\rightarrow$8B has regret only $0.287$. Transfer is also asymmetric. The configuration selected at 8B incurs regret $0.028$ at 4B, whereas the 4B selection incurs regret $0.222$ at 8B. Similarly, 4B$\rightarrow$14B has regret $0.784$, compared with $0.251$ in the reverse direction. These results rule out parameter-count distance as a sufficient criterion for deciding whether a configuration should be reused.

\subsubsection{Cross-family}
\label{app:cross-family-transfer}

\begin{figure*}[t]
\centering
\includegraphics[width=\textwidth]{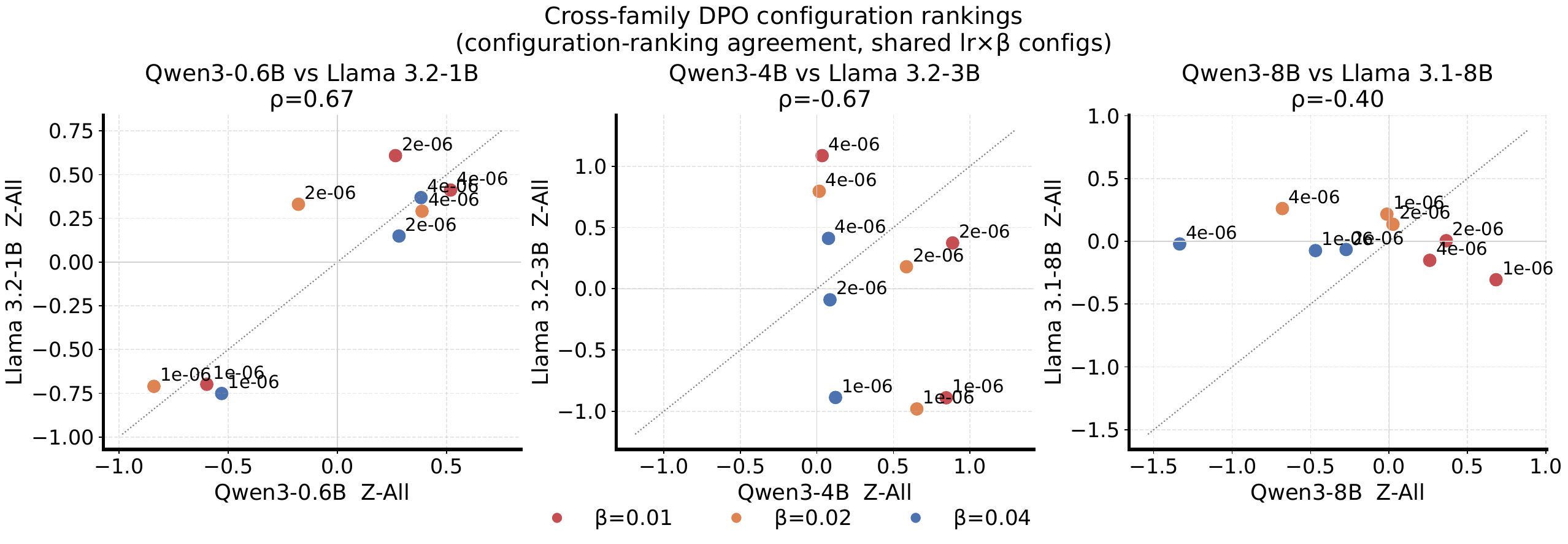}
\caption{\textbf{Configuration rankings do not transfer consistently across model families.} Each panel compares final \zall{} scores for the nine shared $(\mathrm{learning\ rate},\beta)$ configurations. Points are coloured by $\beta$ and annotated with learning rate; the dotted diagonal is a visual reference. Agreement is positive for the smallest pair but negative for the two larger comparisons.}
\label{fig:cross-family-transfer}
\end{figure*}

We next compare \zall{} rankings over the same nine configurations between the nearest available Qwen3 and Llama sizes. Scores are standardized separately within each model's grid, so the analysis compares configuration orderings rather than absolute score levels. Figure~\ref{fig:cross-family-transfer} shows positive agreement for Qwen3-0.6B and Llama~3.2-1B ($\rho=0.67$), but negative agreement for Qwen3-4B and Llama~3.2-3B ($\rho=-0.67$) and for Qwen3-8B and Llama~3.1-8B ($\rho=-0.40$). Similar parameter counts therefore do not ensure that the two families prefer the same DPO configuration.

This comparison is exploratory. It contains only three model pairs, two pairs are not exactly size-matched, and the 8B comparison uses Llama~3.1 whereas the smaller Llama models use Llama~3.2. The results establish that cross-family ranking reversal occurs in the tested settings, but they do not estimate how frequently it should occur across model families.

\section{Preliminary Validation Loss Directed BO Runs}
\label{app:validation-loss-bo}
Before constructing the primary grids, we ran a separate sequential search over DPO learning rate and $\beta$ at each Qwen3 size, using held-out DPO loss as the objective. These experiments train every proposed configuration and are distinct from the offline HPO simulations in Section~\ref{sec:rq3}: here, BO chooses continuous hyperparameters by minimizing validation loss, whereas the offline experiments compare search algorithms using downstream evaluation aggregates. The runs also start from SFT checkpoints trained with full-sequence loss, rather than the assistant-only-loss checkpoints used by the primary grids. We therefore report them as preliminary evidence and do not pool their configurations with the main analyses. 

\paragraph{Search protocol.}
We use Optuna's tree-structured Parzen estimator (TPE)~\citep{akiba2019optuna,bergstra2011algorithms}. TPE fits separate Parzen density models to configurations with relatively low and high observed loss and proposes points using its expected-improvement density ratio. Each model-size study begins with five startup trials, then continues sequentially to a fixed budget of 20 trials. No pruning or improvement-based stopping rule is used. The objective is the DPO loss on the same fixed 1,300-pair validation set used in Section~\ref{sec:rq1}; it is evaluated every 100 training steps. Table~\ref{tab:validation-loss-bo-settings} gives the remaining search settings.

\begin{table}[h]
\centering
\small
\setlength{\tabcolsep}{5pt}
\caption{\textbf{Settings for the preliminary validation loss directed searches.} The different $\beta$ domain at 0.6B reflects the bounds used in these preliminary runs.}
\label{tab:validation-loss-bo-settings}
\begin{tabular}{lll}
\toprule
Setting & Qwen3-0.6B & Qwen3-1.7B--14B \\
\midrule
Learning rate & \multicolumn{2}{c}{$[10^{-6},5\!\times\!10^{-5}]$, log-uniform} \\
$\beta$ & $[0.05,1]$, uniform & $[0.01,1]$, log-uniform \\
Sampler & \multicolumn{2}{c}{Optuna TPE; seed 8} \\
Objective & \multicolumn{2}{c}{minimize held-out DPO loss} \\
Startup / total trials & \multicolumn{2}{c}{5 / 20 per model size} \\
Stopping and pruning & \multicolumn{2}{c}{fixed trial budget; no pruner} \\
SFT loss & \multicolumn{2}{c}{full-sequence loss} \\
\bottomrule
\end{tabular}
\end{table}

\paragraph{Validation loss remains unreliable for chat-based selection.}
Table~\ref{tab:validation-loss-bo-selection} evaluates the completed trial with the lowest validation loss at each size. Of the 100 proposed trials, 93 have both validation loss and all three downstream aggregates in the export; trials missing either quantity are excluded. Regret is normalised within the complete preliminary trials for that model, with zero denoting the best downstream score and one the worst. The selected learning rate rises from 0.6B to 1.7B and then decreases from $4.69\times10^{-5}$ at 1.7B to $4.38\times10^{-6}$ at 14B, a $10.7\times$ reduction that resembles the non-monotonic shape observed in the primary grids. The selected $\beta$ has no monotonic size trend but remains in the lower part of its search domain: it is $0.011$--$0.064$ from 1.7B to 14B, while the 0.6B selection, $0.053$, is close to that run's lower bound of $0.05$. We therefore interpret the preliminary runs as additional evidence that larger models above 1.7B favour lower learning rates under this training recipe, while supporting only the weaker conclusion that the preferred $\beta$ remains low rather than constant across scale.

The downstream quality of the minimum-loss trial depends strongly on the evaluation objective. Mean regret across sizes is $0.21$ for \zall{} and $0.16$ for \zstatic{}, but $0.39$ for \zdynamic{}. At 1.7B, the minimum-loss trial ranks 16th of 18 under \zdynamic{} and incurs $0.89$ regret; at 4B, it ranks 16th of 19 with $0.48$ regret. Conversely, it is optimal under \zstatic{} at 4B and 14B. Thus, adaptive sampling does not make validation loss a uniformly reliable downstream objective: in these runs it aligns more closely with static performance than with chat-based performance.

\begin{table}
\centering
\small
\setlength{\tabcolsep}{4pt}
\caption{\textbf{Downstream quality of the minimum validation loss TPE trial.} $n$ is the number of trials with complete loss and downstream results. Aggregate columns report normalised regret followed by the selected trial's downstream rank in parentheses; lower values are better for both quantities.}
\label{tab:validation-loss-bo-selection}
\begin{tabular}{lrrrccc}
\toprule
Model & $n$ & Learning rate & $\beta$ & \zall{} & \zdynamic{} & \zstatic{} \\
\midrule
Qwen3-0.6B & 20 & $1.83\times10^{-5}$ & 0.0526 & 0.20 (4)  & 0.26 (11) & 0.12 (5) \\
Qwen3-1.7B & 18 & $4.69\times10^{-5}$ & 0.0109 & 0.63 (15) & 0.89 (16) & 0.36 (10) \\
Qwen3-4B   & 19 & $2.60\times10^{-5}$ & 0.0164 & 0.13 (5)  & 0.48 (16) & 0.00 (1) \\
Qwen3-8B   & 16 & $5.39\times10^{-6}$ & 0.0638 & 0.11 (8)  & 0.15 (3)  & 0.29 (9) \\
Qwen3-14B  & 20 & $4.38\times10^{-6}$ & 0.0165 & 0.00 (1)  & 0.17 (5)  & 0.00 (1) \\
\midrule
Mean regret & --- & --- & --- & 0.21 & 0.39 & 0.16 \\
\bottomrule
\end{tabular}
\end{table}

Rank correlations give the same qualification. We compute Spearman correlation between negative validation loss, so that larger values are better, and each downstream aggregate. Averaged across sizes, the correlations are $0.54$ with \zall{}, $0.16$ with \zdynamic{}, and $0.67$ with \zstatic{}. For \zdynamic{}, the correlation is negative at 1.7B ($-0.45$) and 4B ($-0.20$), showing that lower validation loss can favour the opposite configuration ordering from chat-based evaluation. Because these searches use a different SFT loss, search domain, and adaptively sampled set of configurations, their absolute regrets are not directly comparable to those of the primary grids in Section~\ref{sec:rq1}.

\paragraph{Benchmark agreement persists on the adaptively sampled configurations.}
Although TPE queries a different, continuous set of configurations at each size, the qualitative agreement pattern from Section~\ref{sec:rq1} remains visible. Table~\ref{tab:validation-loss-bo-agreement} reports the mean and population standard deviation of the pairwise Spearman correlations within each evaluation group. Mean chat-benchmark agreement is positive at every size and increases from $0.49$ at 0.6B to $0.85$ at 14B. Static agreement is lower throughout, ranging from $0.08$ to $0.36$, with substantially larger variation across benchmark pairs. Because IFEval was not available for enough preliminary trials to estimate its pairwise correlations, the static summary uses the other six static benchmarks. At 8B, the chat summary uses the three available pairs among Arena-Hard, MT-Bench, and Elo; the other sizes use all six chat-benchmark pairs. Pairwise correlations are computed over the configurations for which both constituent evaluations are available, yielding 16--20 configurations depending on model size and benchmark pair.

\begin{table}
\centering
\small
\setlength{\tabcolsep}{6pt}
\caption{\textbf{Benchmark-ranking agreement within the preliminary TPE runs.} Entries give the mean pairwise Spearman correlation $\rho$ and its population standard deviation across benchmark pairs. $P$ is the number of benchmark pairs contributing to each summary.}
\label{tab:validation-loss-bo-agreement}
\begin{tabular}{lrrrr}
\toprule
& \multicolumn{2}{c}{Chat benchmarks} & \multicolumn{2}{c}{Static benchmarks} \\
\cmidrule(lr){2-3}\cmidrule(lr){4-5}
Model & $\rho\pm\sigma$ & $P$ & $\rho\pm\sigma$ & $P$ \\
\midrule
Qwen3-0.6B & $0.49\pm0.19$ & 6 & $0.15\pm0.35$ & 15 \\
Qwen3-1.7B & $0.52\pm0.42$ & 6 & $0.17\pm0.31$ & 15 \\
Qwen3-4B   & $0.64\pm0.33$ & 6 & $0.30\pm0.53$ & 15 \\
Qwen3-8B   & $0.82\pm0.08$ & 3 & $0.36\pm0.34$ & 15 \\
Qwen3-14B  & $0.85\pm0.05$ & 6 & $0.08\pm0.45$ & 15 \\
\bottomrule
\end{tabular}
\end{table}

Together, these results show that the contrast between chat and static rankings is not confined to the rectangular grids. They also explain why a validation loss directed search can appear successful under some static evaluations while selecting a poor chat configuration: the two evaluation groups induce different configuration rankings, and validation loss aligns much more strongly with the static group in this preliminary training recipe.

\end{document}